%% file: arxiv.tex
\documentclass[runningheads]{llncs}

\usepackage{eccv}

\usepackage{eccvabbrv}

\usepackage{booktabs, graphicx, wrapfig}
\usepackage{tikz}
\usetikzlibrary{calc}
\usepackage{enumitem}
\usepackage{adjustbox}
\usepackage{siunitx,adjustbox}
\usepackage[table]{xcolor}
\usepackage{multirow}
\usepackage{comment}
\usepackage{tablefootnote}

\usepackage[accsupp]{axessibility}  

\usepackage{hyperref}

\usepackage{orcidlink}

\begin{document}

\title{\textsc{VLOD-TTA}: Test-Time Adaptation of Vision-Language Object Detectors} 


\author{Atif Belal\orcidlink{0009-0000-5167-1216} \and
Heitor R. Medeiros\orcidlink{0000-0002-1300-3337} \and
Marco Pedersoli\orcidlink{0000-0002-7601-8640} \and 
Eric Granger\orcidlink{0000-0001-6116-7945}}

\authorrunning{A. Belal et al.}

\institute{LIVIA, ILLS, Dept.\ of Systems Engineering, ETS Montr\'eal, Canada\\
\email{\{atif.belal.1, heitor.rapela-medeiros.1\}@ens.etsmtl.ca, \\ \{marco.pedersoli, eric.granger\}@etsmtl.ca}}

\maketitle

\input{0-abstract}
\input{1-introduction2}
\input{2-literature}
\input{3-proposed_method}
\input{4-experiments2}
\input{5-conclusion}

\input{6-supp}

\clearpage



%
%
\bibliographystyle{splncs04}
\bibliography{main}
\end{document}

%% file: 0-abstract.tex
\begin{abstract}
Vision-language object detectors (VLODs) such as YOLO-World and Grounding DINO exhibit strong zero-shot generalization, but their performance degrades under distribution shift. Test-time adaptation (TTA) offers a practical way to adapt models during inference using only unlabeled target (test) data. However, while TTA has made substantial progress in vision-language classification, its application to VLODs remains largely unexplored. The only prior method relies on a mean-teacher framework that introduces significant latency and memory overhead. 
To this end, we introduce \textsc{VLOD-TTA}, a TTA method that leverages dense proposal overlap and image-conditioned prompts to adapt VLODs with low additional overhead. \textsc{VLOD-TTA} combines (i) an IoU-weighted entropy objective that emphasizes spatially coherent proposal clusters and mitigates confirmation bias from isolated boxes, and (ii) image-conditioned prompt selection that ranks prompts by image-level compatibility and aggregates the most informative prompt scores for detection.
Our experiments across diverse distribution shifts, including artistic domains, adverse driving conditions, low-light imagery, and common corruptions, indicate that \textsc{VLOD-TTA} consistently outperforms standard TTA baselines and the prior state-of-the-art method using YOLO-World and Grounding DINO.\\
Our code: \url{https://github.com/imatif17/VLOD-TTA}
\keywords{Test-Time Adaptation \and Vision-Language Object Detection}
\end{abstract}

%% file: 1-introduction2.tex
\section{Introduction}

Object detectors (ODs) localize and classify objects in images~\cite{zou2023object}, with applications in surveillance~\cite{surveillance}, autonomous driving~\cite{self-driving}, and medical imaging~\cite{medical-imaging}. Recently, vision-language ODs (VLODs) such as YOLO-World~\cite{yoloworld} and Grounding DINO~\cite{groundingdino} have demonstrated strong zero-shot (ZS) generalization by aligning region-level visual features with textual representations through large-scale image-text pretraining~\cite{object365,gqa}.

Despite their strong ZS capability, VLODs remain sensitive to distribution shift at test time~\cite{tpt}. Although source-free domain adaptation~\cite{sfda1,sfda2} can mitigate this issue, it typically assumes access to pre-collected target-domain data and offline adaptation. These assumptions are often impractical in deployment settings, where the test environment changes over time and adaptation must be performed at inference. This motivates test-time adaptation (TTA) for VLODs, which adapts the model online during deployment using only unlabeled test data. 

\begin{figure}[t]
  \centering \includegraphics[width=.95\linewidth,height=0.3\textheight,keepaspectratio]{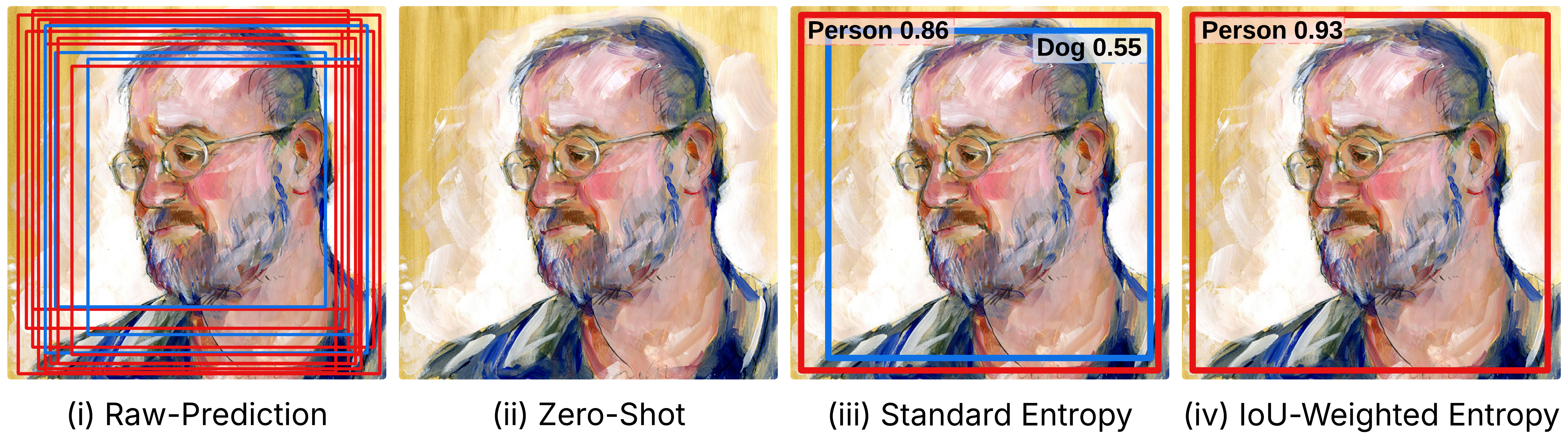}
  \vspace{-.18cm}
  \caption{\textbf{Motivation for IWE. Left→right:} (i) Raw predictions before thresholding for two classes — a true positive \textbf{Person (red)} (bbox cluster size = 167, max score = 0.14) and a false positive \textbf{Dog (blue)} (bbox cluster size = 45, max score = 0.15); (ii) In ZS, all scores remain below the detection threshold, causing a missed detection; (iii) Standard entropy minimization over-confidently sharpens scores, resulting in a dog false positive; and (iv) \textbf{IWE} focuses updates on dense bbox clusters and suppresses isolated boxes, producing only the correct person detection.}
  \label{fig:intro_ent}
    \vspace{-.2cm}
\end{figure}
\begin{figure}[t]
  \centering
  \includegraphics[width=.95\linewidth,height=0.3\textheight,keepaspectratio]{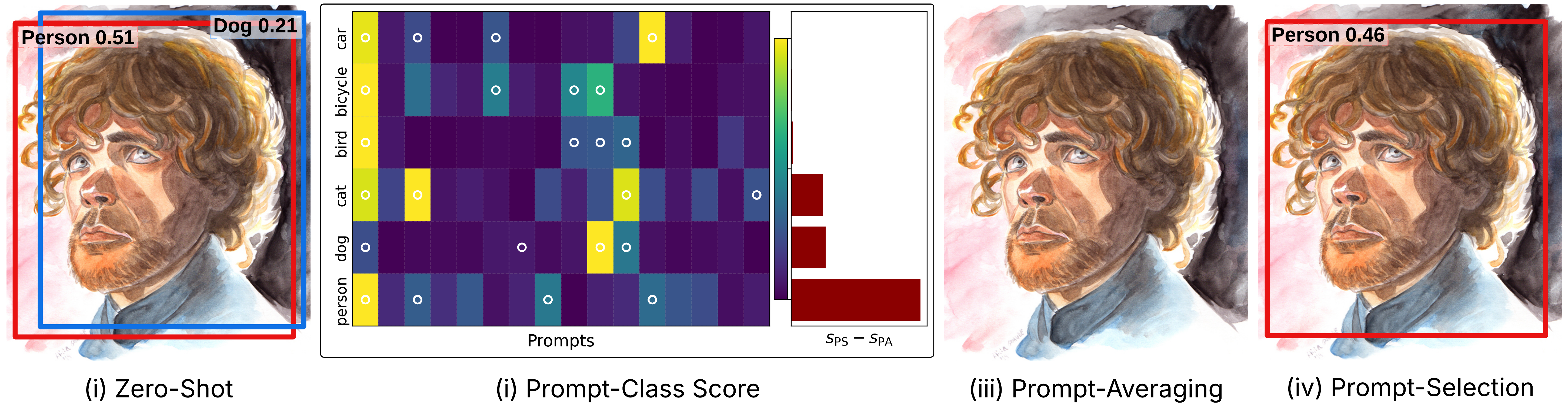}
  \vspace{-.18cm}
  \caption{\textbf{Motivation for IPS. Left→right:} (i) ZS produces a correct \textbf{Person (red)} detection but also a \textbf{Dog (blue)} false positive; (ii) prompt–class score heatmap, where circles denote prompts selected by \textbf{IPS} and the right-margin bars show $S_{\text{PS}}-S_{\text{PA}}$, indicating that prompt selection increases the correct person-class score; (iii) prompt averaging (PA) reduces detection confidence, producing no detections; and (iv) prompt selection (PS) suppresses the dog false positive while preserving the person detection.}
  \label{fig:intro_prompt}
  \vspace{-.5cm}
\end{figure}

The only prior TTA method specialized for VLODs, TTAOD-F~\cite{ttaod-f}, adopts a multimodal, prompt-based mean-teacher framework with joint text- and visual-prompt tuning, along with memory-enhanced pseudo-labeling. 
While effective, its teacher-based design introduces substantial computational overhead at test time due to extra model components and auxiliary feature processing. Moreover, it is tailored to transformer-based detectors and does not readily extend to CNN-based VLODs such as YOLO-World. This highlights the need for an efficient TTA framework for VLODs that avoids teacher-based adaptation and extends beyond transformer-based detectors.

In parallel, TTA methods for vision-language classification often rely on entropy minimization over augmented views~\cite{tpt,swapprompt}. While entropy minimization offers a lower-overhead alternative to teacher-based approaches, it has two key limitations for ODs. First, it amplifies confirmation bias by sharpening the highest class score, which can make mislocalized proposals overly confident~\cite{frustrating_easy}. Second, it ignores proposal structure and assigns the same weight to isolated or cross-instance boxes as to spatially consistent overlapping proposal clusters. As illustrated in \Cref{fig:intro_ent}, standard entropy minimization can increase the scores of both \emph{person} and \emph{dog} predictions without considering spatial coherence, leading to a false-positive dog detection.

Beyond model adaptation, prompt ensembling is a common strategy for improving the robustness of vision-language models (VLMs), typically by averaging multiple prompt templates per class~\cite{clip}. However, for VLODs, prompt averaging provides only marginal gains and can even degrade performance, as shown in \Cref{sec:ablation}. As shown in \Cref{fig:intro_prompt}, prompt averaging reduces the \emph{person} score below the detection threshold, resulting in a missed detection.

To address these limitations, we propose \textbf{\textsc{VLOD-TTA}}, a TTA framework for VLODs with two key components: \textbf{IoU-weighted entropy minimization (IWE)} and \textbf{image-conditioned prompt selection (IPS)}. Modern ODs generate dense, overlapping proposals that provide partially redundant views of the same instance. IWE leverages this spatial redundancy by assigning each proposal a weight based on its local IoU affinity with overlapping, class-consistent proposals. This biases adaptation toward spatially coherent proposals and reduces confirmation bias from isolated or mislocalized ones. In \Cref{fig:intro_ent}, IWE increases scores within the dominant \emph{person} cluster instead of uniformly amplifying all proposals, thereby suppressing the false-positive \emph{dog} prediction produced by standard entropy minimization.

We further introduce IPS, an OD-specific prompt selection mechanism. Rather than averaging all prompts, IPS computes an image-conditioned selection score for each prompt and retains only the top-$\rho$ fraction per class. The retained prompt logits are then fused with the base OD logits. By conditioning on test-image features, IPS retains the most informative, context-relevant prompts while suppressing irrelevant ones.
In \Cref{fig:intro_prompt}, IPS selects prompts aligned with the input image and raises the \emph{person} score relative to prompt averaging, resulting in a correct detection. Together, IWE and IPS provide an efficient TTA framework that improves VLOD robustness under distribution shift with low overhead.

\noindent\textbf{Our main contributions are summarized as follows.}
\textbf{(1)} We introduce IWE, a detection-specific entropy minimization objective for VLOD TTA that exploits local proposal overlap and mitigates confirmation bias.
\textbf{(2)} An IPS mechanism is introduced that replaces prompt averaging with image-relevant prompt fusion, improving VLOD robustness under distribution shift.
\textbf{(3)} Comprehensive benchmarking on six standard detection datasets and 15 common corruptions, covering 96 distinct test scenarios, shows that \textsc{VLOD-TTA} consistently outperforms TTA baselines on state-of-the-art CNN- and transformer-based VLODs.



%% file: 2-literature.tex
\section{Related Work}

\noindent \textbf{Test-Time Adaptation.  } TTA mitigates domain shift by updating a subset of model parameters using only unlabeled test data. Tent~\cite{tent} minimizes predictive entropy by updating batch normalization parameters over batches of images. Memo~\cite{memo} avoids batches by minimizing marginal entropy across augmented views of a single test image.
For VLMs, TTA methods mainly fall into two families: prompt-tuning~\cite{tpt,difftpt} and cache-based methods~\cite{tda,dpe}. Prompt-tuning methods optimize continuous prompt tokens in the text encoder for each test image.
Cache-based methods maintain a memory of high-confidence target features and pseudo-labels to calibrate predictions online.
These methods operate at the image level for classification and do not involve region proposals, leaving localization unaddressed. In contrast, \textsc{VLOD-TTA} performs proposal-level adaptation for open-vocabulary detection, jointly addressing localization and classification.\\

\noindent \textbf{Vision-Language Object Detectors.  } VLODs localize and classify categories specified by text, thereby relaxing the closed-set constraint of conventional ODs. Early methods leverage VLMs~\cite{clip} to transfer language-aligned semantics into detector classifier heads~\cite{vild,regionclip,ovrcnn}. Vocabulary scaling further decouples localization and classification by training large-vocabulary classifiers with image-level labels while keeping proposals class-agnostic~\cite{detic}. 
Grounded pretraining unifies detection and phrase grounding to learn language-aware object representations~\cite{glip}. Grounding DINO~\cite{groundingdino} fuses language into a transformer detector via language-conditioned queries and cross-modal decoding. For efficiency, YOLO-World~\cite{yoloworld} introduces reparameterizable vision–language fusion for real-time open-vocabulary detection. Despite strong ZS performance, VLODs remain sensitive to domain shift, motivating the study of TTA for open-vocabulary ODs.\\ 

\noindent \textbf{Test-Time Adaptation for Object Detectors.  }For ODs, TTA aims to improve robustness under domain shift without requiring source data during deployment. Prior TTA methods for ODs mainly rely on self-training and pseudo-label refinement. These include mean-teacher adaptation with feature alignment~\cite{stfar}, stability-aware adapter updates~\cite{what_how_when}, object-level contrastive alignment with selective restoration~\cite{armod}, and single-image adaptation with IoU-guided pseudo-label filtering~\cite{fullytesttimeOD}. Although effective, many TTA methods for ODs rely on heavy augmentations and multi-step updates, which can limit practical deployment. Moreover, they are designed for conventional closed-set ODs with a fixed vision-only label space and assume source-pretrained detectors. In comparison, \textsc{VLOD-TTA} adapts VLODs in an open-vocabulary setting by leveraging both visual and textual cues. TTAOD-F~\cite{ttaod-f} is the only prior study on TTA for VLODs. It uses a mean-teacher framework with text and visual prompt tuning in a transformer architecture. This design incurs substantial adaptation-time overhead, as it maintains a second detector as teacher, requires an additional forward pass for pseudo-label generation, and uses DINOv2~\cite{dinov2} features for memory-based prediction refinement. %
In contrast, \textsc{VLOD-TTA} uses an efficient entropy-based objective that does not require a teacher forward pass and generalizes to both CNN- and transformer-based VLODs.






%% file: 3-proposed_method.tex
\section{Proposed VLOD-TTA Method} 
\label{sec:method}

\begin{figure}[!t]
\centering	\includegraphics[width=0.99\linewidth]{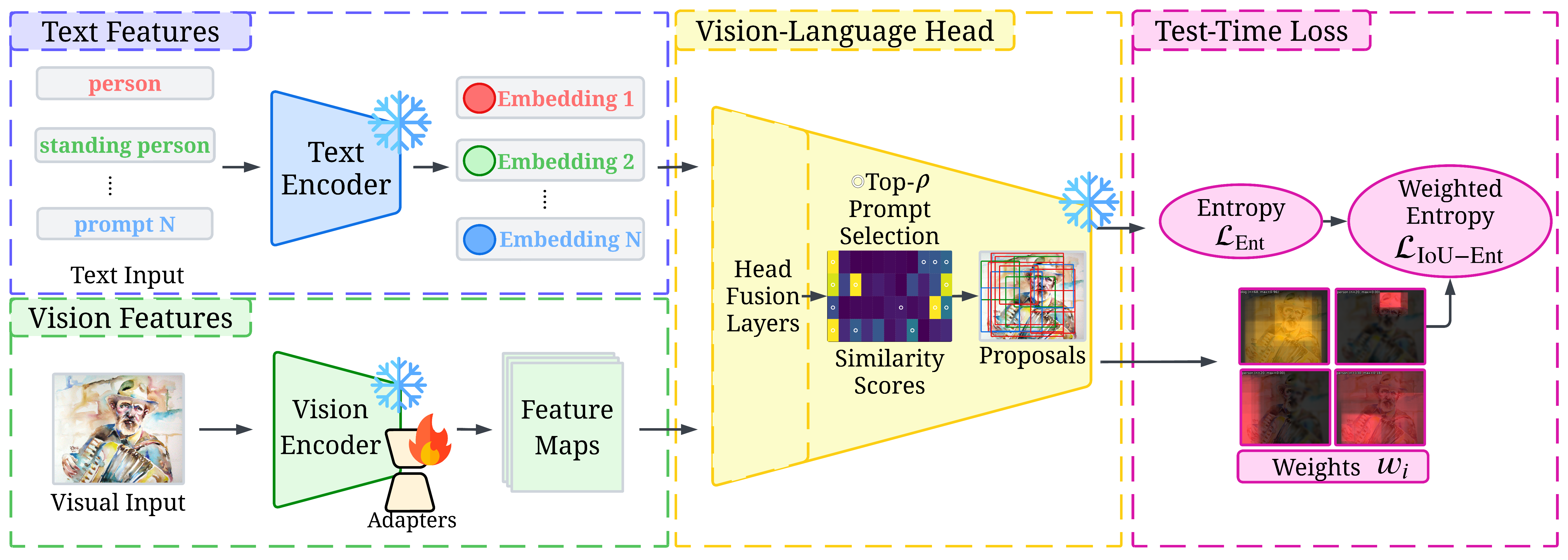}
\vspace{-.1cm}
    \caption{\textbf{Overview of \textsc{VLOD-TTA}.} 
Given an input image and a pool of class prompts, the text encoder produces embeddings that interact with region proposals through the vision–language head to compute similarity scores. 
IPS selects the top-$\rho$ prompts and averages their scores to obtain per-proposal class scores. IWE then combines per-proposal entropy with IoU-based weights to form the adaptation objective.}
\label{fig:proposed_method}
\vspace{-.4cm}
\end{figure}

An overview of \textsc{VLOD-TTA} is shown in \Cref{fig:proposed_method}. Our framework consists of two components: IoU-weighted entropy minimization (IWE) and image-conditioned prompt selection (IPS). IWE reweights proposal entropies using local IoU consistency to emphasize reliable regions during adaptation. IPS computes image-conditioned prompt relevance and retains only the most informative prompts. We first introduce the preliminaries and then describe each component in detail.

\subsection{Preliminary Definitions}

\noindent \textbf{Vision–Language Object Detection. }
A VLOD couples a visual detector with a text encoder in a shared embedding space.
Given an image $X \in \mathbb{R}^{C \times H \times W}$, the visual detector outputs $N$ candidate boxes $B=\{b_i\}_{i=1}^{N}$ and corresponding region features $\{\mathbf{v}_i\}_{i=1}^N$, where $\mathbf{v}_i \in \mathbb{R}^d$. For each category name $y_k$ in a label set $Y$ with $|Y|=K$, the text encoder produces an embedding $\mathbf{t}_k \in \mathbb{R}^d$. The similarity score between proposal $i$ and class $k$ is
\begin{equation}
\label{eq:detector_score}
s_{i,k}= \hat{\mathbf{v}}_i^{\top}\hat{\mathbf{t}}_k,\qquad
\hat{\mathbf{v}}_i=\frac{\mathbf{v}_i}{\lVert \mathbf{v}_i\rVert_2},\quad
\hat{\mathbf{t}}_k=\frac{\mathbf{t}_k}{\lVert \mathbf{t}_k\rVert_2}.
\end{equation}
Final detections are obtained using detector-specific post-processing, such as score thresholding and non-maximum suppression.\\

\noindent \textbf{Entropy Minimization for ODs. } Given class scores $s_{i,k}$ for proposal $i$ and class $k$, the categorical posterior for proposal $b_i$ is $p_{i,k}=\big[\mathrm{softmax}(s_{i,1},\ldots,s_{i,K})\big]_k$. The Shannon entropy~\cite{shannon1948} of proposal $i$ is 
$\mathcal{H}(\mathbf{p}_i)=-\sum_{k=1}^K p_{i,k}\,\log p_{i,k}.$
A standard TTA objective minimizes the average entropy over all proposals:
\begin{equation}
\mathcal{L}_{\mathrm{Ent}}=\frac{1}{N}\sum_{i=1}^N \mathcal{H}(\mathbf{p}_i).
\end{equation}
This objective sharpens proposal-level class posteriors and reduces predictive uncertainty.

\subsection{IoU-weighted Entropy Minimization (IWE)}
VLODs produce a large number of candidate boxes per image. In standard configurations, YOLO-World (YW) produces approximately $8{,}400$ proposals, and Grounding DINO (GD) produces approximately $900$. Although post-processing removes most proposals, their spatial structure remains informative. Regions containing many mutually overlapping proposals with the same predicted class are more likely to correspond to true objects, whereas sparse or dispersed proposals are typically less reliable. Standard entropy minimization, such as Tent~\cite{tent}, treats proposals independently and ignores their overlap structure, which can sharpen predictions in unreliable regions and amplify confirmation bias.

To address this limitation, \textsc{VLOD-TTA} introduces IWE, which assigns larger weights to groups of mutually overlapping proposals and smaller weights to isolated proposals. Let $\hat c_i=\arg\max_k p_{i,k}$ denote the predicted class of proposal $b_i$. For each class $c$, we construct a class-specific IoU graph $G_c=(V_c, E_c)$, whose vertices are proposals with predicted label $c$. Two vertices $u$ and $v$ are connected by an edge if $\operatorname{IoU}(b_u,b_v)\ge\theta$, where $\theta\in[0,1]$ is a fixed threshold.
Let $\mathcal{C}(i)$ denote the connected component of $G_c$ containing proposal $i$. We define the weight as $w_i = |\mathcal{C}(i)|^{\gamma}$, where $|\mathcal{C}(i)|$ is the size of the corresponding connected component and $\gamma\ge 0$ controls the influence of component size. The IWE objective is 
\begin{equation}
\label{eq:iou_ent}
 \mathcal{L}_{\mathrm{IoU-Ent}}
=\frac{\sum_{i=1}^N w_i\,\mathcal{H}(\mathbf{p}_i)}{\sum_{i=1}^N w_i}.
\end{equation}
The weights depend only on the IoU graph and are treated as constants during backpropagation, so gradients flow only through $\mathcal{H}(\mathbf{p}_i)$. This objective emphasizes spatially coherent proposal groups while reducing the influence of isolated or inconsistent proposals.

\subsection{Image-Conditioned Prompt Selection (IPS)}

VLM performance is sensitive to prompt wording. A common strategy is to use multiple prompt templates per class and average their text embeddings. Although CLIP reports improved zero-shot accuracy with this approach~\cite{clip}, we find that uniform prompt averaging is often ineffective for VLODs and can even degrade performance. Instead of treating all prompts equally, IPS selects informative prompts for each image and adapts the text representation at test time.

For each class $k\in\{1,\dots,K\}$, let $\{t_{k,1},\dots,t_{k,T}\}$ denote a pool of $T$ prompts with text embeddings $\{\mathbf{e}_{k,t}\in\mathbb{R}^d\}$. Given $N$ region proposals with features $\{\mathbf{v}_i\}_{i=1}^{N}$, we compute prompt-conditioned similarities as $z_{i,k,t} = \hat{\mathbf{v}}_i^\top \hat{\mathbf{e}}_{k,t}$, where $\hat{\mathbf{v}}_i$ and $\hat{\mathbf{e}}_{k,t}$ are $\ell_2$-normalized. For each class $k$ and prompt $t$, we define an image-conditioned compatibility score as $r_{k,t}=\frac{1}{N}\sum_{i=1}^{N} z_{i,k,t}$. This score measures the average compatibility of prompt $t$ with the proposal features for class $k$ in the current image (see Supp. for theoretical justification). 
To suppress irrelevant prompts, IPS selects, for each class $k$, the top-$\rho$ fraction of prompts with the highest compatibility scores $r_{k,t}$. Let $\mathcal{S}_k$ denote the selected prompt indices. The class score for proposal $i$ is then computed by averaging only over the selected prompts, $\tilde{z}_{i,k}=\frac{1}{|\mathcal{S}_k|}\sum_{t\in\mathcal{S}_k} z_{i,k,t}$.

To further align text and region embeddings, we introduce a lightweight residual vector in the text embedding space~\cite{dpe}. Let $\Delta\in\mathbb{R}^d$ be a learnable residual added to each prompt embedding: 
\begin{equation}
    \label{eq:residual}
\tilde{\mathbf{e}}_{k,t} \;=\; \frac{\mathbf{e}_{k,t}+\Delta}{\lVert \mathbf{e}_{k,t}+\Delta\rVert_2}.
\end{equation}
 
Let $s_{i,k}$ denote the base VLOD class score from \Cref{eq:detector_score}. The final fused score is $g_{i,k} = \lambda \tilde{z}_{i,k} + (1-\lambda) s_{i,k}$, where $\lambda \in (0,1)$.

\subsection{Model Update}
Although \textsc{VLOD-TTA} can in principle adapt any subset of parameters, we keep the base network fixed and update only a small set of adapter parameters~\cite{adapter, convadapter}. For YW, we insert adapters into the backbone and neck, whereas for GD, we insert adapters only into the text encoder. This distinction reflects their different architectures and pretraining paradigms (see \Cref{sec:ablation} for details). Let $\Theta$ denote all network parameters, with $\Theta=(\Theta_{\text{frozen}},\Phi,\Delta)$, where $\Phi$ represents the adapter parameters and $\Delta$ is the residual parameter defined in \Cref{eq:residual}. At test time, we zero-initialize $(\Phi,\Delta)$ as $(\Phi_0,\Delta_0)$. Given the final fused score $g_{i,k}$, we retain the top-$M$ proposals ranked by $\max_k g_{i,k}$ to accelerate IoU graph construction. We then optimize $(\Phi,\Delta)$ for a single adaptation step using the IWE objective. Since adaptation is performed on a single test image, the updated parameters are not expected to generalize across images. We therefore reset $(\Phi,\Delta)$ to $(\Phi_0,\Delta_0)$ after each prediction.

%% file: 4-experiments2.tex
\section{Results and Discussion}

\textsc{VLOD-TTA} is evaluated on diverse domain shift benchmarks. We first describe the benchmark setup and TTA baselines. Then, the main results are analyzed across different shift types, followed by ablations on the key design choices of the proposed method.

\subsection{Benchmarking VLOD-TTA}
We compare ZS inference, four TTA baselines adapted to VLODs, and our \textsc{VLOD-TTA}. The benchmark covers four types of domain shift: texture and style, weather, illumination, and common corruptions. It includes six domain-shift datasets and two corruption benchmarks. Results are reported for YOLO-World (YW) and Grounding DINO (GD) using mean average precision (mAP)~\cite{lin2014microsoft}.\\

\noindent\textbf{Datasets. } \textbf{Watercolor, ClipArt, Comic:} Watercolor, ClipArt, and Comic~\cite{inoue2018crossdomain} are stylized artistic datasets used to evaluate robustness to style shifts.
\textbf{Cityscapes:} Cityscapes contains urban street scenes from multiple European cities~\cite{cityscapes}. We use it to study domain shift in driving scenes across geography, weather, and time of day.
\textbf{BDD100K:} BDD100K is a large-scale driving dataset spanning day and night, multiple cities, and diverse weather conditions~\cite{bdd100k}. We use it to evaluate robustness under real-world distribution shifts.
\textbf{ExDark:} ExDark is a low-light OD dataset~\cite{exdark} used to measure robustness under poor illumination.
\textbf{PASCAL-C and COCO-C:} PASCAL-C and COCO-C~\cite{michaelis2019benchmarking} corrupt PASCAL-VOC~\cite{pascal_voc} and COCO~\cite{lin2014microsoft} using 15 common corruption types at five severity levels, following~\cite{hendrycksbenchmarking}, to evaluate OD robustness under common corruptions. \\

\noindent\textbf{Baselines.  } \textbf{Zero-shot:} Pretrained VLODs are used directly for inference without any adaptation~\cite{groundingdino,yoloworld}.
\textbf{Test-Time Prompt Tuning (TPT):} We adapt TPT~\cite{tpt} from classification to OD by optimizing only text prompt vectors at test time. Candidate boxes are selected based on entropy, and a marginal-entropy objective is minimized over those proposals.
\textbf{Visual Prompt Tuning (VPT):} Following the TPT pipeline, we optimize only visual prompts~\cite{vpt}. Visual prompting has been shown to be effective for modality adaptation~\cite{modprompt}.
\textbf{DPE:} We adapt DPE~\cite{dpe} to OD by maintaining per-class text and visual caches constructed from high-confidence proposals. At test time, only the residual cache parameters are updated using a marginal-entropy objective combined with a cache-contrastive loss.
\textbf{Adapter Tuning:} We adapt lightweight bottleneck adapters using the Tent objective~\cite{tent}. This removes dependence on model-specific prompt parameters and provides a fairer comparison with our method. \\

\noindent\textbf{Implementation Details.  } We report AP using the COCO API~\cite{lin2014microsoft}. Unless stated otherwise, we use YW-Small and GD-Tiny. Each experiment uses a batch size of 1 and a single adaptation step to target real-time deployment settings.
We set $\gamma=1.1$, $\rho=0.25$, $M=600$, and $\lambda=0.3$ for YW, and $\lambda=0.1$ for GD (see Supp. for hyperparameter sensitivity analysis).
Due to architectural differences, we use Conv-Adapters~\cite{convadapter} in YW with a reduction factor of 4 and a kernel size of 3, and an MLP Adapter~\cite{adapter} in GD with a reduction factor of $r=16$ (see Supp. for details).
We use $T=16$ GPT-generated prompts per class. We use ChatGPT-5 to generate textual prompts with the instruction: ``Generate 16 prompts for each object category: $<\text{\textit{category list}}>$'' (see Supp. for examples of prompts).


\subsection{Main Results}

\begin{table*}[t]
\centering
\caption{\textbf{Detection performance on benchmark datasets.}  We report mAP, AP$_{50}$, and AP$_{75}$ for both \textbf{YW} and \textbf{GD} ODs on six benchmark datasets -- Watercolor, ClipArt, Comic, Cityscapes, BDD100K, and ExDark. 
Best results are in bold.}
\scriptsize
\setlength{\tabcolsep}{2.4pt}
\renewcommand{\arraystretch}{0.95}

\begin{adjustbox}{max width=\textwidth}
\begin{tabular}{
  l
  *{18}{S[table-format=2.1]}
}
\toprule

\multicolumn{19}{c}{\textbf{YOLO-World}}\\
\addlinespace[0.25em]
\cmidrule(lr){1-19}  

& \multicolumn{3}{c}{\textbf{Watercolor}}
& \multicolumn{3}{c}{\textbf{ClipArt}}
& \multicolumn{3}{c}{\textbf{Comic}}
& \multicolumn{3}{c}{\textbf{Cityscapes}}
& \multicolumn{3}{c}{\textbf{BDD100K}}
& \multicolumn{3}{c}{\textbf{ExDark}}\\
\cmidrule(lr){2-4}\cmidrule(lr){5-7}\cmidrule(lr){8-10}
\cmidrule(lr){11-13}\cmidrule(lr){14-16}\cmidrule(lr){17-19}
\textbf{Method}
& \textbf{mAP} & \textbf{AP$_{50}$} & \textbf{AP$_{75}$}
& \textbf{mAP} & \textbf{AP$_{50}$} & \textbf{AP$_{75}$}
& \textbf{mAP} & \textbf{AP$_{50}$} & \textbf{AP$_{75}$}
& \textbf{mAP} & \textbf{AP$_{50}$} & \textbf{AP$_{75}$}
& \textbf{mAP} & \textbf{AP$_{50}$} & \textbf{AP$_{75}$}
& \textbf{mAP} & \textbf{AP$_{50}$} & \textbf{AP$_{75}$} \\
\midrule
ZS\cite{yoloworld}                         & 26.9 & 47.9 & 25.9 & 24.4 & 40.1 & 26.2 & 17.8 & 29.4 & 18.8 & 18.8 & 31.0 & 17.9 & 13.3 & 22.0 & 13.4 & 35.2 & 64.7 & 34.6 \\
TPT\cite{tpt}                        & 27.3 & 48.5 & 26.1 & 24.9 & 41.3 & 26.8 & 18.1 & 29.9 & 19.1 & 18.8 & 31.1 & 18.0 & 13.4 & 22.2 & 13.5 & 35.8 & 65.1 & 34.7 \\
VPT\cite{vpt}                       & 26.9 & 49.1 & 25.1 & 25.0 & 41.4 & 26.9 & 18.3 & 30.9 & 19.3 & 18.9 & 31.2 & 18.0 & 13.5 & 22.3 & 13.2 & 35.8 & 65.8 & 34.9 \\
DPE\cite{dpe}                        & 27.2 & 48.9 & 26.3 & 24.9 & 41.5 & 27.1 & 18.9 & 31.7 & 19.8 & 19.0 & 31.3 & 18.0 & 13.5 & 22.3 & 13.3 & 35.9 & 66.4 & 35.1 \\
Adapter\cite{tent}                   & 28.3 & 51.5 & 26.7 & 26.9 & 44.1 & 27.8 & 20.8 & 34.7 & 21.7 & 19.1 & 31.3 & 18.3 & 13.7 & 21.7 & 13.1 & 35.8 & 66.4 & 35.1 \\ 

\bottomrule
\rowcolor{gray!15}
\textbf{VLOD-TTA}               & \textbf{29.6} & \textbf{53.1} & \textbf{28.7} & \textbf{28.1} & \textbf{45.4} & \textbf{29.9} & \textbf{21.4} & \textbf{36.1} & \textbf{22.1} & \textbf{19.4} & \textbf{31.8} & \textbf{18.6} & \textbf{14.6} & \textbf{24.3} & \textbf{14.8} & \textbf{36.4} & \textbf{67.4} & \textbf{35.6} \\

\toprule
\multicolumn{19}{c}{\textbf{Grounding DINO}}\\
\addlinespace[0.25em]
\cmidrule(lr){1-19}

& \multicolumn{3}{c}{\textbf{Watercolor}}
& \multicolumn{3}{c}{\textbf{ClipArt}}
& \multicolumn{3}{c}{\textbf{Comic}}
& \multicolumn{3}{c}{\textbf{Cityscapes}}
& \multicolumn{3}{c}{\textbf{BDD100K}}
& \multicolumn{3}{c}{\textbf{ExDark}} \\
\cmidrule(lr){2-4}\cmidrule(lr){5-7}\cmidrule(lr){8-10}
\cmidrule(lr){11-13}\cmidrule(lr){14-16}\cmidrule(lr){17-19}
\textbf{Method}
& \textbf{mAP} & \textbf{AP$_{50}$} & \textbf{AP$_{75}$}
& \textbf{mAP} & \textbf{AP$_{50}$} & \textbf{AP$_{75}$}
& \textbf{mAP} & \textbf{AP$_{50}$} & \textbf{AP$_{75}$}
& \textbf{mAP} & \textbf{AP$_{50}$} & \textbf{AP$_{75}$}
& \textbf{mAP} & \textbf{AP$_{50}$} & \textbf{AP$_{75}$}
& \textbf{mAP} & \textbf{AP$_{50}$} & \textbf{AP$_{75}$} \\
\midrule
ZS\cite{groundingdino}       & 37.4 & 62.9 & 37.6 & 38.4 & 58.8 & 41.8 & 31.2 & 52.9 & 31.5 & 24.1 & 38.2 & 24.5 & 16.6 & 28.3 & 16.2 & 35.4 & 66.2 & 34.5 \\
TPT\cite{tpt}      & 37.4 & 63.1 & 37.6 & 38.6 & 59.1 & 42.3 & 31.5 & 53.6 & 31.8 & 24.6 & 38.6 & 24.8 & 16.6 & 28.4 & 16.2 & 35.6 & 66.5 & 34.9 \\
VPT\cite{vpt}      & 37.2 & 63.0 & 37.8 & 38.6 & 59.0 & 41.9 & 31.1 & 52.6 & 31.2 & 24.0 & 38.2 & 24.3 & 16.7 & 28.5 & 16.2 & 35.1 & 66.4 & 34.4 \\
DPE\cite{dpe}      & 37.6 & 63.2 & 38.1 & 38.2 & 59.3 & 42.0 & 31.8 & 53.3 & 31.7 & 24.4 & 38.3 & 24.4 & 16.7 & 28.6 & 16.3 & 35.2 & 66.6 & 34.3 \\
Adapter\cite{tent} & 38.4 & 63.6 & 39.0 & 38.6 & 58.9 & 41.8 & 31.7 & 54.1 & 32.0 & 24.6 & 39.1 & 24.6 & 16.8 & 28.7 & 16.6 & 35.7 & 66.8 & 34.5 \\
\rowcolor{gray!15}
\bottomrule
\textbf{VLOD-TTA} 
         & \textbf{38.9} & \textbf{64.7} & \textbf{39.5} 
         & \textbf{41.2} & \textbf{62.1} & \textbf{43.3} 
         & \textbf{34.2} & \textbf{57.8} & \textbf{35.3} 
         & \textbf{25.8} & \textbf{40.8} & \textbf{25.9} 
         & \textbf{18.1} & \textbf{31.1} & \textbf{18.5} 
         & \textbf{37.3} & \textbf{68.9} & \textbf{36.8} \\
\bottomrule

\end{tabular}
\end{adjustbox}

\vspace{-.5cm}

\label{tab:main_yolo_world}
\end{table*}

\textbf{Texture and style shifts (\textit{Watercolor, ClipArt, Comic}).  } Results in \Cref{tab:main_yolo_world} show that \textsc{VLOD-TTA} achieves the best performance on all three stylized datasets for both YW and GD. Among prompt-based baselines, VPT is slightly more effective than TPT on YW, with an average gain of +0.6 AP$_{50}$, whereas TPT is slightly more effective than VPT on GD by +0.4 AP$_{50}$ on average. This trend is consistent with the adapter placement analysis in \Cref{sec:ablation}. Adapter is the strongest prior baseline, improving AP$_{50}$ over ZS by +4.3 on YW averaged across the three datasets, but only by +0.7 on GD. In contrast, \textsc{VLOD-TTA} delivers larger and more consistent gains. Relative to ZS, \textsc{VLOD-TTA} improves YW by an average of +3.3 mAP, +5.8 AP$_{50}$, and +3.2 AP$_{75}$ across the three datasets. For GD, the corresponding gains are +2.4 mAP, +3.3 AP$_{50}$, and +2.4 AP$_{75}$. These results indicate that combining IWE and IPS is particularly effective under appearance shifts that alter texture and visual style while preserving object semantics.\\

\noindent \textbf{Autonomous driving under varying conditions (\textit{Cityscapes, BDD100K}).  } \Cref{tab:main_yolo_world} shows that adaptation on driving scenes is more challenging than on stylized domains, likely because these datasets contain many small objects with weak or sparse proposal overlap (see Supp. for a detailed analysis and ways to mitigate this issue). As a result, standard entropy-based baselines provide only marginal improvements over ZS. On YW, Adapter even underperforms ZS on BDD100K in both AP$_{50}$ and AP$_{75}$, suggesting that standard entropy can overfit unreliable proposals in crowded scenes. In contrast, \textsc{VLOD-TTA} achieves the best results on both datasets and both detectors. Averaged over Cityscapes and BDD100K, it improves YW over ZS by +1.0 mAP, +1.6 AP$_{50}$, and +1.1 AP$_{75}$. For GD, the corresponding average gains are +1.6 mAP, +2.7 AP$_{50}$, and +1.9 AP$_{75}$. Although the absolute gains are smaller than on stylized datasets, they remain consistent, showing that \textsc{VLOD-TTA} is effective even in more structured and challenging driving scenarios.\\

\noindent \textbf{Illumination shift (\textit{ExDark}).  } Under low-light conditions, \textsc{VLOD-TTA} again achieves the best performance on both YW and GD, as shown in \Cref{tab:main_yolo_world}. On YW, DPE and Adapter are the strongest prior baselines, both reaching 66.4 AP$_{50}$, which indicates that low-light adaptation benefits from strong prior information. \textsc{VLOD-TTA} further improves over ZS by +1.2 mAP, +2.7 AP$_{50}$, and +1.0 AP$_{75}$. On GD, Adapter is the strongest prior baseline, while the gains of the other baselines remain limited. In comparison, \textsc{VLOD-TTA} improves over ZS by +1.9 mAP, +2.7 AP$_{50}$, and +2.3 AP$_{75}$. These results indicate that \textsc{VLOD-TTA} improves both detection confidence and localization quality under severe low-light conditions.\\

\begin{table*}[t]
\centering
\caption{\textbf{Detection performance on PASCAL-C.}  AP$_{50}$ is reported for the \textbf{YW} detector on 15 different data corruptions. }
\vspace{-.1cm}

\scriptsize
\setlength{\tabcolsep}{2.4pt}
\renewcommand{\arraystretch}{0.95}

\begin{adjustbox}{max width=\textwidth}
\begin{tabular}{
  l
  *{15}{S[table-format=2.1]}
  | S[table-format=2.1]
}
\toprule

& \multicolumn{3}{c}{\textbf{Noise}}
& \multicolumn{4}{c}{\textbf{Blur}}
& \multicolumn{3}{c}{\textbf{Weather}}
& \multicolumn{5}{c}{\textbf{Digital}}
& \multicolumn{1}{c}{\textbf{}} \\
\cmidrule(lr){2-4}\cmidrule(lr){5-8}\cmidrule(lr){9-11}\cmidrule(lr){12-16}\cmidrule(lr){17-17}

\textbf{Method}
& \textbf{Gauss}
& \textbf{Shot}
& \textbf{Impul}
& \textbf{Defoc}
& \textbf{Glass}
& \textbf{Motn}
& \textbf{Zoom}
& \textbf{Snow}
& \textbf{Frost}
& \textbf{Fog}
& \textbf{Brit}
& \textbf{Contr}
& \textbf{Elast}
& \textbf{Pixel}
& \textbf{JPEG}
& \textbf{Avg}
\\
\midrule
ZS\cite{yoloworld}  & 16.9 & 17.2 & 16.2 & 32.4 & 11.3 & 26.9 & 30.1 & 46.4 & 50.6 & 70.4 & 73.6 & 41.3 & 42.0 & 8.9  & 26.1 & 34.0 \\
TPT\cite{tpt}      & 17.4 & 17.9 & 16.4 & 33.1 & 11.7 & 27.1 & 30.7 & 47.2 & 51.9 & 68.9 & 71.1 & 42.8 & 42.7 & 9.5  & 27.2 & 34.4 \\
VPT\cite{vpt}      & 17.8 & 18.2 & 16.7 & 33.0 & 12.3 & 27.7 & 30.5 & 47.7 & 51.3 & 69.9 & 72.5 & 42.5 & 43.5 & 10.9 & 28.4 & 34.9 \\
DPE\cite{dpe}      & 17.7 & 18.7 & 17.5 & 32.8 & 12.9 & 28.5 & 30.7 & 48.4 & 52.1 & 70.9 & 73.4 & 42.9 & 43.2 & 11.5 & 30.7 & 35.5 \\
Adapter\cite{tent} & 20.5 & 22.8 & 23.1 & 35.9 & 15.1 & 29.0 & 30.2 & 48.1 & 52.5 & 71.1 & 72.1 & 45.8 & 46.5 & 16.1 & 36.8 & 37.7 \\
\bottomrule
\textbf{VLOD-TTA} & \textbf{22.9} & \textbf{24.4} & \textbf{24.2} & \textbf{37.2} & \textbf{16.6} & \textbf{29.3} & \textbf{30.9} & \textbf{50.9} & \textbf{53.8} & \textbf{73.2} & \textbf{74.1} & \textbf{48.7} & \textbf{49.4} & \textbf{17.8} & \textbf{40.2} & \textbf{39.6}\\
\bottomrule
\vspace{-1cm}

\end{tabular}
\end{adjustbox}
\label{tab:pascal-c}

\end{table*} 

\noindent \textbf{Common corruptions.   } \Cref{tab:pascal-c} reports AP$_{50}$ on PASCAL-C across 15 corruption types using YW. \textsc{VLOD-TTA} achieves the best performance on every corruption and the highest overall average of 39.6 AP$_{50}$. It outperforms the strongest baseline, Adapter, by 1.9 AP$_{50}$ and improves over ZS by 5.6 AP$_{50}$. The largest gains over ZS are observed on JPEG Compression (+14.1), Pixelate (+8.9), Impulse Noise (+8.0), Contrast (+7.4), Elastic Transform (+7.4), and Shot Noise (+7.2). These improvements are particularly strong under noise and digital corruptions. Overall, the results show that \textsc{VLOD-TTA} provides broad and consistent robustness across common corruption types.

\subsection{Ablation Studies}
\label{sec:ablation}

\noindent\textbf{Contribution of individual components. } We ablate the two components of \textsc{VLOD-TTA} on YW across Watercolor, ClipArt, and Comic in \Cref{tab:component_ablation}. Adapter serves as the standard entropy-minimization baseline. IWE replaces standard entropy with IoU-weighted entropy, and \textsc{VLOD-TTA} combines IWE with IPS. Compared with Adapter, IWE consistently improves both mAP and AP$_{50}$ across all three datasets. Adding IPS on top of IWE yields further consistent gains. These results show that both components contribute positively, and their combination achieves the best performance.\\

\noindent\textbf{Adapters versus batch normalization parameters. } In the main experiments, we optimize adapter parameters. To show that \textsc{VLOD-TTA} is not restricted to a particular parameter subset, we compare adapter tuning with batch normalization updates under both standard entropy and \textsc{VLOD-TTA} in \Cref{tab:bn_vs_adapter}. Under standard entropy, the two parameter choices yield similar improvements over ZS on Watercolor, ClipArt, and Comic. Applying \textsc{VLOD-TTA} further improves both and consistently outperforms the corresponding entropy-only baseline. Batch norm with \textsc{VLOD-TTA} nearly matches the adapter-based version, showing that the proposed objective is effective beyond adapters. However, batch norm requires dataset-specific learning-rate tuning to achieve its best performance, for example $1e{-2}$ on Watercolor and $3e{-2}$ on ClipArt, whereas adapters perform well with a single learning rate of $5e{-3}$ across datasets. In the TTA setting, the target domain is unknown at test time, which makes per-dataset learning-rate tuning impractical for real-time detection. We therefore use adapters in the main experiments.

\begin{figure}[t]
\centering

\begin{minipage}[t]{0.48\textwidth}
\vspace{.02cm}
\centering
\captionof{table}{\textbf{Ablation study on the components of \textsc{VLOD-TTA}.} Detection performance on three style-shift datasets.}
\label{tab:component_ablation}

\scriptsize
\setlength{\tabcolsep}{2.4pt}
\renewcommand{\arraystretch}{0.95}

\begin{adjustbox}{max width=\linewidth}
\begin{tabular}{
  l
  *{6}{S[table-format=2.1]}
}
\toprule
& \multicolumn{2}{c}{\textbf{Watercolor}}
& \multicolumn{2}{c}{\textbf{ClipArt}}
& \multicolumn{2}{c}{\textbf{Comic}} \\
\cmidrule(lr){2-3}\cmidrule(lr){4-5}\cmidrule(lr){6-7}
\textbf{Method}
& \textbf{mAP} & \textbf{AP$_{50}$}
& \textbf{mAP} & \textbf{AP$_{50}$}
& \textbf{mAP} & \textbf{AP$_{50}$} \\
\midrule
ZS              & 26.9 & 47.9 & 24.4 & 40.1 & 17.8 & 29.4 \\
Adapter                & 28.3 & 51.5 & 26.9 & 44.1 & 20.8 & 34.7 \\
IWE                    & 29.3 & 52.6 & 27.5 & 44.7 & 21.2 & 35.6 \\
\midrule
\rowcolor{gray!15}
\textsc{\textbf{VLOD-TTA}}
                       & \textbf{29.6} & \textbf{53.1}
                       & \textbf{28.1} & \textbf{45.4}
                       & \textbf{21.4} & \textbf{36.1} \\
\bottomrule
\end{tabular}
\end{adjustbox}
\end{minipage}
\hfill
\begin{minipage}[t]{0.48\textwidth}
\vspace{-.04cm}
\centering
\captionof{table}{\textbf{Adapters vs.\ batch norm. as adaptation parameters.} Detection performance on three style-shift datasets.}
\label{tab:bn_vs_adapter}
\scriptsize
\setlength{\tabcolsep}{2.4pt}
\renewcommand{\arraystretch}{0.95}

\begin{adjustbox}{max width=\linewidth}
\begin{tabular}{
  l
  *{6}{S[table-format=2.1]}
}
\toprule
& \multicolumn{2}{c}{\textbf{Watercolor}}
& \multicolumn{2}{c}{\textbf{ClipArt}}
& \multicolumn{2}{c}{\textbf{Comic}} \\
\cmidrule(lr){2-3}\cmidrule(lr){4-5}\cmidrule(lr){6-7}
\textbf{Method}
& \textbf{mAP} & \textbf{AP$_{50}$}
& \textbf{mAP} & \textbf{AP$_{50}$}
& \textbf{mAP} & \textbf{AP$_{50}$} \\
\midrule
ZS           & 26.9 & 47.9 & 24.4 & 40.1 & 17.8 & 29.4 \\
BN              & 28.4 & 51.3 & 26.7 & 44.1 & 20.6 & 34.5 \\
Adapters            & 28.3 & 51.5 & 26.9 & 44.1 & 20.8 & 34.7 \\
\midrule
\rowcolor{gray!12}
\textbf{VLOD-TTA BN}
                    & 29.4 & 52.9
                    & \textbf{28.3} & 45.3
                    & 21.3 & \textbf{36.1} \\
\rowcolor{gray!12}
\textbf{VLOD-TTA}   & \textbf{29.6} & \textbf{53.1}
                    & 28.1 & \textbf{45.4}
                    & \textbf{21.4} & \textbf{36.1} \\
\bottomrule
\end{tabular}
\end{adjustbox}
\end{minipage}
\vspace{-.5cm}
\end{figure}

\vspace{-.5cm}
\begin{table}[h]
\centering
\caption{\textbf{Performance and inference cost comparison on COCO-C.} We report corruption-group mAP along with total and tuned parameter counts, GPU memory usage, and latency. All experiments are run on an RTX3090 GPU. \textsc{VLOD-TTA}$^*$ denotes our method with TTAOD-F-based initialization.}
\scriptsize
\setlength{\tabcolsep}{2.4pt}
\renewcommand{\arraystretch}{0.95}

\begin{adjustbox}{max width=\textwidth}
\begin{tabular}{
  l
  *{4}{S[table-format=2.1]}
  | S[table-format=2.1]
  | S[table-format=3.1]
  S[table-format=1.2]
  S[table-format=2.2]
  S[table-format=3.1]
}
\toprule

& \multicolumn{4}{c|}{\textbf{Corruption Type Avg. mAP} $\uparrow$}
& \multicolumn{1}{c|}{\textbf{Overall}}
& \multicolumn{4}{c}{\textbf{Inference Cost} $\downarrow$} \\
\cmidrule(lr){2-5}\cmidrule(lr){6-6}\cmidrule(lr){7-10}

\textbf{Method}
& \textbf{Noise}
& \textbf{Blur}
& \textbf{Weather}
& \textbf{Digital}
& \textbf{Avg}
& \textbf{\shortstack[c]{Total Params\\(Mil)}}
& \textbf{\shortstack[c]{Tuned Params\\(Mil)}}
& \textbf{\shortstack[c]{Memory\\(GB)}}
& \textbf{\shortstack[c]{Latency\\(ms/img)}}

\\
\midrule
ZS\cite{groundingdino}
& 14.9 & 11.2 & 34.7 & 23.0 & 20.6
& 172.9 & 0.00 & 1.36 & 210.6\\

TTAOD-F\cite{ttaod-f}\tablefootnote{\scriptsize TTAOD-F uses batch size 4 (original setting), whereas \textsc{VLOD-TTA} uses batch size 1. Although batch size 1 reduces TTAOD-F memory to 4.43 GB, we observe a deterioration in its performance under this setting.}
& 21.2 & 14.3 & 37.0 & 31.7 & 26.0
& 629.9 & \textbf{0.08} & 11.37 & 701.9 \\

\midrule
\rowcolor{gray!12}
\textbf{VLOD-TTA}
& 21.1 & 15.6 & 38.5 & 30.4 & 26.2
& \textbf{173.9} & 0.89 & 3.76 & \textbf{531.6} \\

\rowcolor{gray!12}
\textbf{VLOD-TTA$^*$}
& \textbf{21.5} & \textbf{15.9} & \textbf{38.6} & \textbf{32.8} & \textbf{27.3}
& \textbf{173.9} & 0.95 & 3.92 & 567.3 \\
\bottomrule
\end{tabular}
\end{adjustbox}
\label{tab:sota_comaprison}
\vspace{-0.5cm}
\end{table}

\noindent\textbf{Comparison with TTAOD-F. } TTAOD-F~\cite{ttaod-f} is the only prior TTA method designed specifically for VLODs. In \Cref{tab:sota_comaprison}, we compare it with \textsc{VLOD-TTA} on COCO-C using GD, reporting corruption-group mAP together with parameter count and latency (see Supp. for full results). \textsc{VLOD-TTA} improves over ZS on all corruption groups and achieves slightly higher average mAP than TTAOD-F (26.2 vs.\ 26.0), while using far fewer total parameters (173.9M vs.\ 629.9M) and lower latency (531.6 vs.\ 701.9 ms/img), with substantially lower GPU memory usage (3.76 vs.\ 11.37 GB). TTAOD-F also uses a test-time warm-start strategy that initializes visual prompts by average pooling image tokens from the first test sample. For fair comparison, we additionally evaluate a warm-start variant, denoted \textsc{VLOD-TTA}$^*$, using the same initialization strategy as TTAOD-F. This variant further improves the average mAP to 27.3 and outperforms both TTAOD-F and the default \textsc{VLOD-TTA} across all corruption groups. In addition, \textsc{VLOD-TTA} keeps the tuned parameter budget small (0.89M, or 0.95M for \textsc{VLOD-TTA}$^*$). These results show that \textsc{VLOD-TTA} offers a substantially better accuracy--efficiency trade-off at test time. The latency gap is consistent with the higher adaptation-time cost of TTAOD-F, which requires multiple forward passes and one backward pass, whereas \textsc{VLOD-TTA} uses a single adaptation step with one forward pass and one backward pass.\\

\begin{figure}[t]
    \centering
    \includegraphics[width=0.7\linewidth]{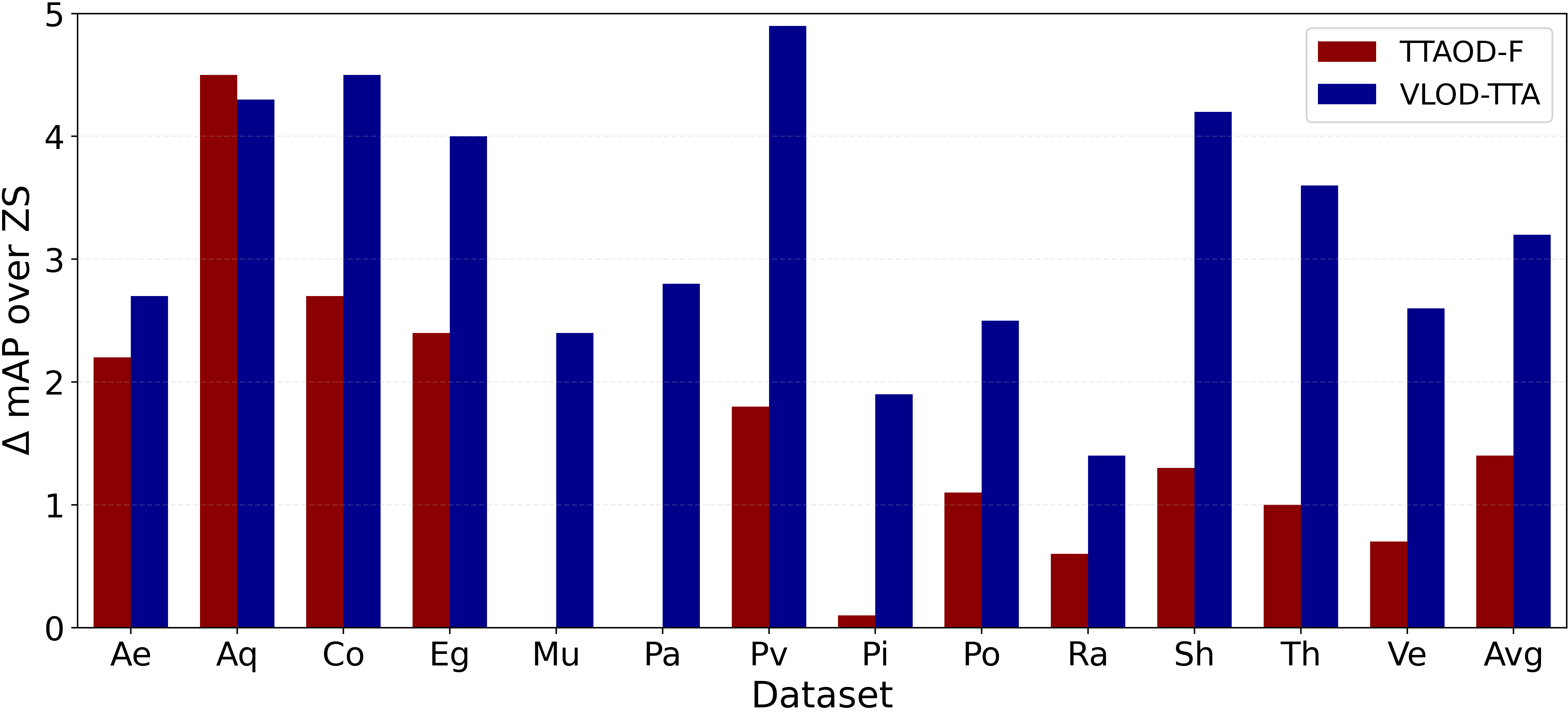}
    \vspace{-0.1cm}
    \caption{\textbf{Improvement over ZS on ODinW-13 using Grounding DINO.} \textsc{VLOD-TTA} achieves larger mAP gains than \textsc{TTAOD-F} on 12 of the 13 datasets.}

    \vspace{-.4cm}
    \label{fig:odinw}
\end{figure}

\noindent\textbf{Performance on ODinW-13. }
ODinW-13 consists of 13 diverse object detection datasets spanning heterogeneous visual domains. As shown in~\Cref{fig:odinw}, \textsc{VLOD-TTA} improves the ZS baseline from 52.8 to 56.0 mAP, corresponding to a gain of $+3.2$ mAP, compared with $+1.4$ mAP for \textsc{TTAOD-F}. Moreover, \textsc{VLOD-TTA} outperforms \textsc{TTAOD-F} on 12 of the 13 datasets. These results demonstrate that \textsc{VLOD-TTA} generalizes effectively across diverse object detection domains.\\

\noindent\textbf{Scalability to larger VLODs. }
We further evaluate VLOD-TTA on the larger YOLO-World-Large and Grounding DINO-Big variants. As shown in~\Cref{tab:large_backbones}, VLOD-TTA consistently improves AP$_{50}$ across all three style-shift datasets. For YOLO-World-Large, the gains range from $+3.0$ to $+4.9$ AP$_{50}$, while Grounding DINO-Big improves by $+2.3$ to $+3.3$ AP$_{50}$. These results indicate that VLOD-TTA remains effective as model capacity increases and is not limited to the smaller detector variants used in our main experiments.

\begin{table}[h]
\centering
\vspace{-0.5cm}
\caption{\textbf{Scalability to larger VLODs.} We report AP$_{50}$ on three style-shift datasets using YOLO-World-Large and Grounding DINO-Big.}
\scriptsize
\setlength{\tabcolsep}{6.0pt}
\renewcommand{\arraystretch}{0.95}

\begin{adjustbox}{max width=\textwidth}
\begin{tabular}{
  l
  *{3}{S[table-format=2.1]}
}
\toprule
\multicolumn{4}{c}{\textbf{YOLO-World-Large}} \\
\cmidrule(lr){1-4}
\textbf{Method} & \textbf{Watercolor} & \textbf{ClipArt} & \textbf{Comic} \\
\midrule
ZS\cite{yoloworld} & 55.3 & 50.6 & 37.9 \\
\rowcolor{gray!12}
\textbf{\textsc{VLOD-TTA}} & \bfseries 58.3 & \bfseries 53.9 & \bfseries 42.8 \\
\midrule
\multicolumn{4}{c}{\textbf{Grounding DINO-Big}} \\
\cmidrule(lr){1-4}
\textbf{Method} & \textbf{Watercolor} & \textbf{ClipArt} & \textbf{Comic} \\
\midrule
ZS\cite{groundingdino} & 70.5 & 77.9 & 64.7 \\
\rowcolor{gray!12}
\textbf{\textsc{VLOD-TTA}} & \bfseries 72.8 & \bfseries 81.2 & \bfseries 67.1 \\
\bottomrule
\end{tabular}
\end{adjustbox}
\label{tab:large_backbones}
\vspace{-0.5cm}
\end{table}

\begin{figure*}[t]
\centering

\begin{minipage}[t]{0.32\textwidth}
    \centering
    \vspace{0pt}
    \includegraphics[width=\linewidth]{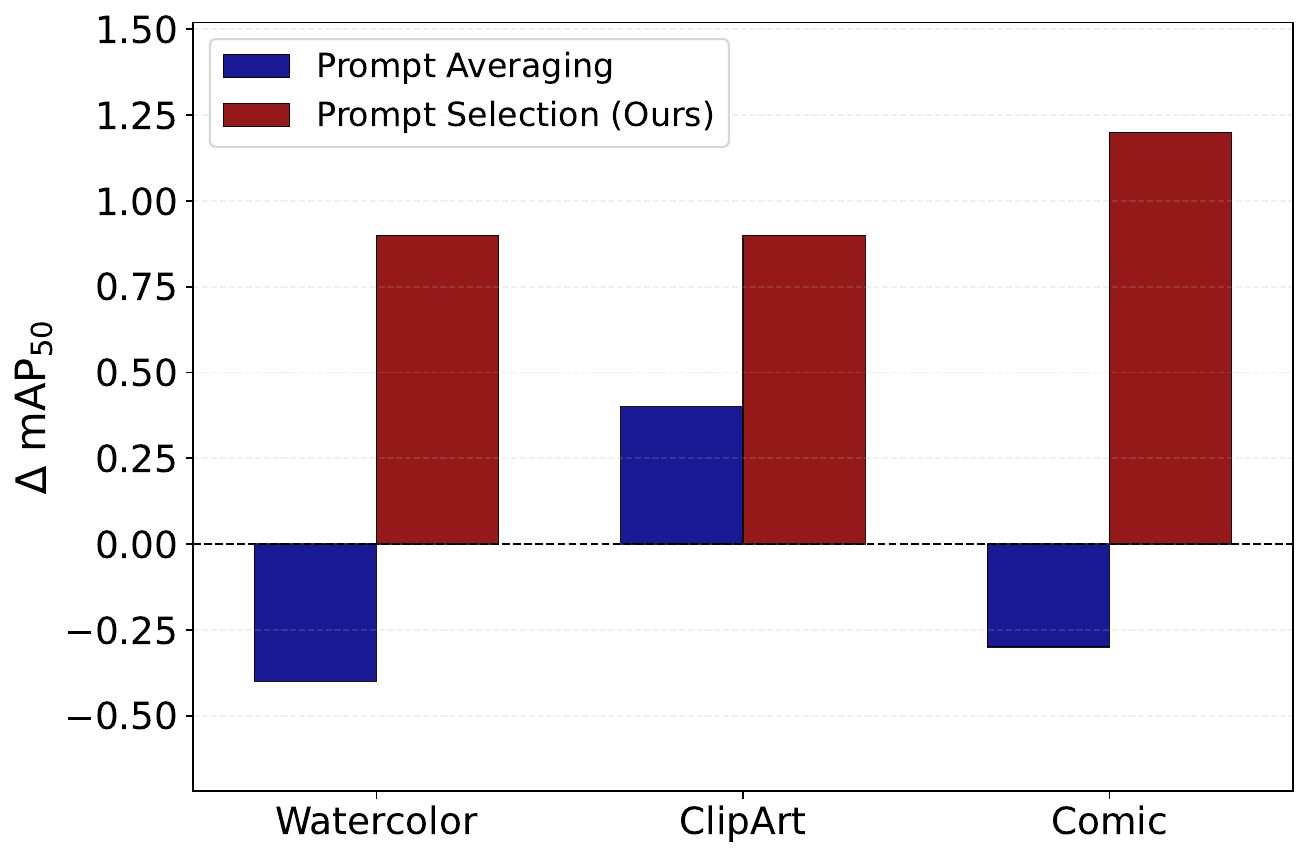}
    \vspace{-0.4cm}
    \subcaption{Prompt averaging vs.\ prompt selection.}
    
    \label{fig:pevsps}
\end{minipage}
\hfill
\begin{minipage}[t]{0.32\textwidth}
    \centering
    \vspace{0pt}
    \includegraphics[width=\linewidth]{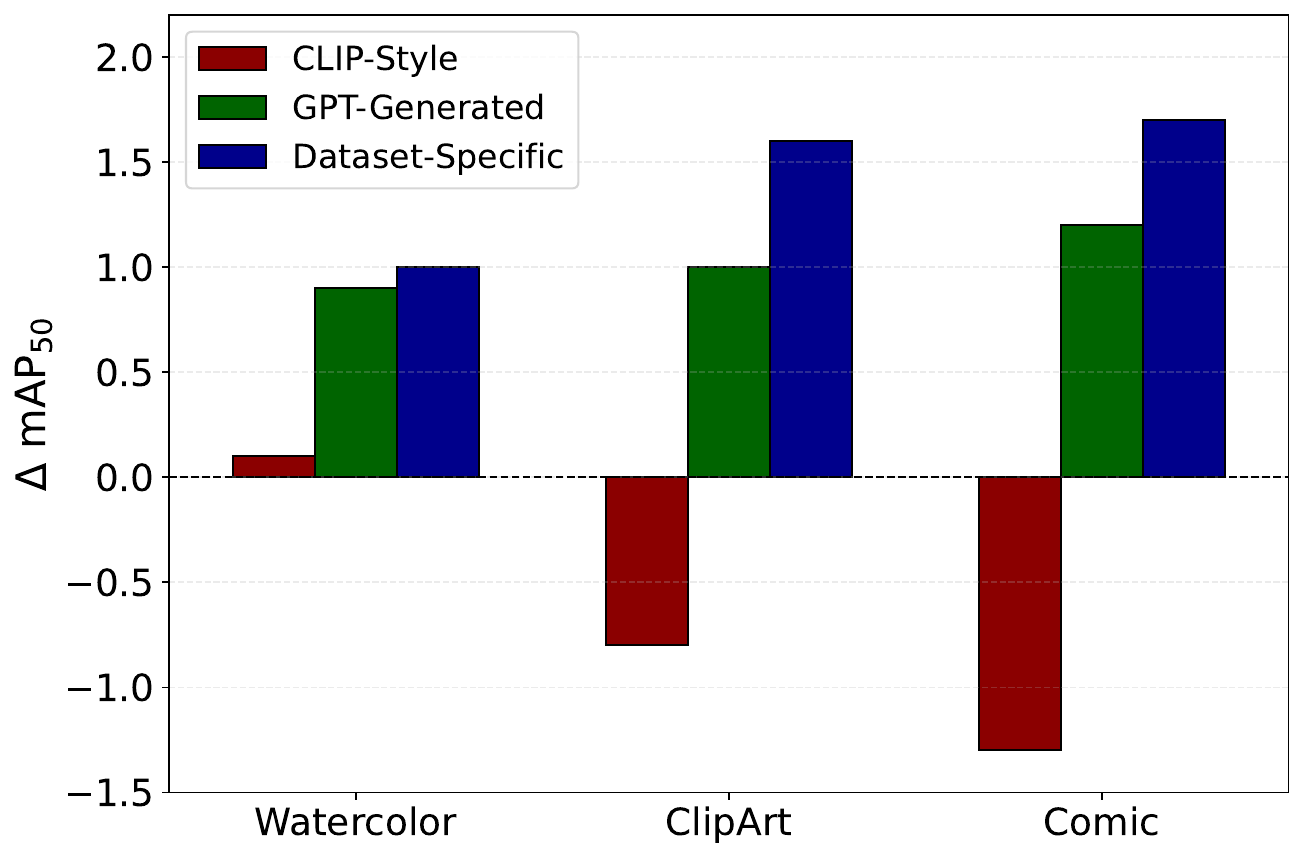}
    \vspace{-0.4cm}
    \subcaption{Prompt generation strategies.}
    \label{fig:prompt_strategy}
\end{minipage}
\hfill
\begin{minipage}[t]{0.32\textwidth}
    \centering
    \vspace{0pt}
    \includegraphics[width=\linewidth]{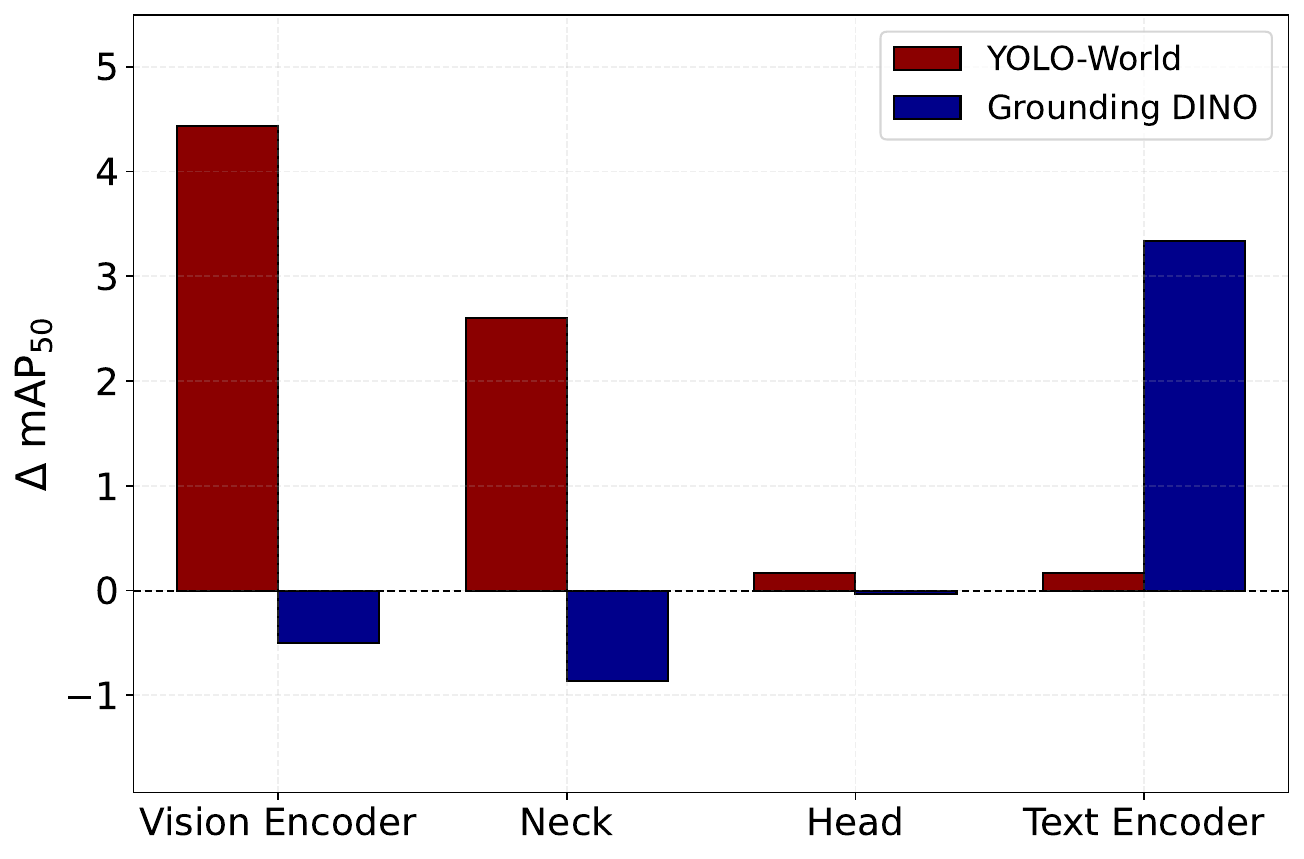}
    \vspace{-0.4cm}
    \subcaption{Adapter placement across modules.}
    \label{fig:adapter_optimized}
\end{minipage}

\vspace{-0.1cm}
\caption{\textbf{Ablation studies on prompt design and adapter placement.}
(a) Comparison between prompt averaging and prompt selection.
(b) Comparison of prompt generation strategies.
(c) Effect of adapter placement across modules.
All values report $\Delta$mAP$_{50}$ over three style-shift datasets relative to zero-shot (ZS).}
\label{fig:ablation_prompt_adapter}
\vspace{-0.5cm}
\end{figure*}

\noindent \textbf{Prompt averaging vs.\ prompt selection.  } In \Cref{fig:pevsps}, we compare two ways of using language prompts in YW: (i) prompt averaging, which averages multiple templates per class into a single text embedding as in CLIP~\cite{clip}, and (ii) prompt selection, which retains only the most informative prompts for each image. Prompt averaging provides only a limited benefit, slightly improving ClipArt while reducing AP$_{50}$ on Watercolor ($-0.35$) and Comic ($-0.30$) relative to ZS. In contrast, prompt selection improves AP$_{50}$ on all three datasets, with an average gain of $+1.0$ over ZS. These results show that uniform prompt aggregation offers limited benefit for VLODs, whereas prompt selection provides a more reliable mechanism for adapting text representations at test time.\\

\noindent\textbf{Variation with different prompt-generation strategies.  } Our default setting uses dataset-agnostic GPT prompts. We compare this setting with dataset-specific GPT prompts and CLIP-style prompts. The AP$_{50}$ improvement over ZS on Watercolor, ClipArt, and Comic is summarized in \Cref{fig:prompt_strategy}. CLIP-style prompts provide a small gain on Watercolor but reduce performance on ClipArt ($-0.8$) and Comic ($-1.3$), which we attribute to a pretraining bias toward label-only text prompts. Dataset-agnostic GPT prompts improve over ZS on all three datasets, with gains of $+0.9$ on Watercolor, $+1.0$ on ClipArt, and $+1.2$ on Comic. Dataset-specific prompts yield the largest gains, reaching $+1.0$ on Watercolor, $+1.6$ on ClipArt, and $+1.7$ on Comic, which suggests that domain-specific cues can further improve adaptation. Since such information is unrealistic in the TTA setting, we do not use dataset-specific prompts in the main experiments.\\

\noindent\textbf{Effect of adapter placement across modules. } In \Cref{fig:adapter_optimized}, we insert adapters into one module at a time and report the change in AP$_{50}$ relative to ZS, averaged over Watercolor, ClipArt, and Comic. For YW, adapting the vision backbone yields the largest gain (+4.4 AP$_{50}$), followed by the neck (+2.6), whereas adapting the head or text encoder provides little benefit. This behavior reflects YW's architecture, where the detector and text encoder are largely decoupled and interact only at the final scoring stage. As a result, updating the text encoder mainly affects classification scores and has a limited impact on detection performance. This observation is also consistent with YW pretraining results, where fine-tuning the text encoder can even degrade performance~\cite{yoloworld}. We therefore omit head adapters in the main experiments to reduce computation.

For GD, the trend is reversed. Adapting the text encoder yields the largest gain (+3.3 AP$_{50}$), whereas adapting the vision encoder or neck slightly degrades performance. This difference is consistent with GD's early cross-modal fusion design, in which text features influence both localization and classification throughout the detector. Consequently, adapting the text encoder is more effective in GD than in YW. This finding also aligns with GD pretraining, where jointly optimizing the text and vision encoders improves performance~\cite{groundingdino}. We do not adapt GD's vision encoder in the main experiments because it tends to overfit to a single test image, likely due to its much larger capacity (172M parameters for GD vs.\ 13M for YW).\\

\noindent\textbf{IoU-weighted pseudo-labeling.  } To test whether IoU-based weighting is useful beyond entropy minimization, we integrate it into a standard pseudo-label-based TTA scheme for ODs. Specifically, teacher predictions are used as supervision for student updates, and overlapping teacher boxes are clustered using IoU, with each pseudo-label weighted by its normalized cluster size. As shown in \Cref{tab:iou_weighted_pseudo_label}, IoU-weighted pseudo-labeling consistently improves both mAP and AP$_{50}$ over the standard pseudo-label baseline across Watercolor, ClipArt, and Comic. This suggests that IoU-based proposal weighting is a general principle that can also strengthen pseudo-label-driven adaptation.\\

\begin{figure}[t]
\centering

\begin{minipage}[t]{0.52\textwidth}
\vspace{-.05cm}
\centering
\captionof{table}{\textbf{IoU-weighted pseudo-labeling (IWPL).} Integrating IoU-weighting into a mean-teacher pseudo-labeling TTA scheme improves AP on three style-shift datasets.}
\label{tab:iou_weighted_pseudo_label}

\scriptsize
\setlength{\tabcolsep}{2.4pt}
\renewcommand{\arraystretch}{0.95}

\begin{adjustbox}{max width=\linewidth}
\begin{tabular}{
  l
  *{6}{S[table-format=2.1]}
}
\toprule
& \multicolumn{2}{c}{\textbf{Watercolor}}
& \multicolumn{2}{c}{\textbf{ClipArt}}
& \multicolumn{2}{c}{\textbf{Comic}} \\
\cmidrule(lr){2-3}\cmidrule(lr){4-5}\cmidrule(lr){6-7}
\textbf{Method}
& \textbf{mAP} & \textbf{AP$_{50}$}
& \textbf{mAP} & \textbf{AP$_{50}$}
& \textbf{mAP} & \textbf{AP$_{50}$} \\
\midrule
ZS     & 26.9 & 47.9 & 24.4 & 40.1 & 17.8 & 29.4 \\
Pseudo-label  & 28.3 & 50.1 & 25.9 & 42.3 & 19.2 & 31.9 \\
\midrule
\rowcolor{gray!12}
\textbf{IWPL} & \textbf{29.1} & \textbf{51.6}
              & \textbf{27.3} & \textbf{43.7}
              & \textbf{20.5} & \textbf{33.2} \\
\bottomrule
\end{tabular}
\end{adjustbox}
\end{minipage}
\hfill
\begin{minipage}[t]{0.44\textwidth}
\vspace{.08cm}
\centering
\captionof{table}{\textbf{Efficiency and performance on YW.} We report FPS, trainable parameters (M), and AP$_{50}$.}
\label{tab:fps-trainparams}
\scriptsize
\setlength{\tabcolsep}{4.5pt}
\renewcommand{\arraystretch}{0.95}

\begin{adjustbox}{max width=\linewidth}
\begin{tabular}{l S[table-format=2.0] S[table-format=1.2] S[table-format=2.1]}
\toprule
\textbf{Method} & \textbf{FPS} $\uparrow$ & \textbf{Params. (M)} $\downarrow$ & \textbf{AP$_{50}$} $\uparrow$ \\
\midrule
ZS      & 89 & 0.00 & 47.9 \\
TPT     & 9  & 1.12 & 48.5 \\
VPT     & 18 & 3.93 & 49.1 \\
DPE     & 15 & \textbf{0.31} & 48.9 \\
Adapter & \textbf{22} & 1.52 & 51.5 \\
\midrule
\rowcolor{gray!12}
\textbf{\textsc{VLOD-TTA}} & 20 & 1.61 & \textbf{53.1} \\
\bottomrule
\end{tabular}
\end{adjustbox}
\end{minipage}
\vspace{-.5cm}
\end{figure}

\noindent \textbf{Runtime and parameter cost. } TTA introduces additional computation beyond ZS inference, so practical deployment requires a favorable accuracy--efficiency trade-off. \Cref{tab:fps-trainparams} compares throughput, trainable parameters, and AP$_{50}$ on YW. \textsc{VLOD-TTA} is faster than TPT, VPT, and DPE, and only slightly slower than Adapter because of IoU-graph construction. Despite this small overhead, \textsc{VLOD-TTA} uses only 1.61M trainable parameters, achieves the best AP$_{50}$, and provides a favorable efficiency--performance trade-off among the TTA baselines.

\subsection{Qualitative Analysis}
\Cref{fig:detection_viz} compares detections from ZS, the Adapter baseline, and \textsc{VLOD-TTA}. The Adapter baseline often sharpens incorrect predictions, for example, the person in the bottom row, which reflects proposal-level confirmation bias under standard entropy minimization.  In contrast, \textsc{VLOD-TTA} suppresses isolated proposals and yields more consistent detections with fewer false positives by emphasizing spatially coherent clusters of overlapping proposals. We also observe cleaner localization, including fewer duplicate boxes. Moreover, \textsc{VLOD-TTA} can detect plausible objects that are missing from the ground truth or refine loosely annotated instances. Although such cases are counted as errors under standard evaluation, they qualitatively suggest improved localization and semantic grounding.

\newcommand{\quadrowwithlabels}[2][0.95\linewidth]{%
\begin{tikzpicture}
  \node[anchor=north west, inner sep=0] (im) at (0,0)
    {\includegraphics[width=#1]{#2}};
  \foreach \txt/\t in {(a) GT/0.125, (b) ZS/0.375, (c) Adapter/0.625, (d) \textsc{VLOD-TTA}/0.875}{
    \node[anchor=south, yshift=2pt,
          fill=white, opacity=0.85, text opacity=1,
          inner sep=1pt, rounded corners=1pt]
      at ($(im.north west)!\t!(im.north east)$) {\scriptsize\textbf{\txt}};
  }
\end{tikzpicture}%
}
\newcommand{\quadrow}[2][0.95\linewidth]{%
\begin{tikzpicture}
  \node[anchor=north west, inner sep=0] (im) at (0,0)
    {\includegraphics[width=#1]{#2}};
\end{tikzpicture}%
}

\begin{figure*}[t]
\vspace{-.1cm}
\begingroup
\setlength{\parskip}{0pt}
\centering

\quadrowwithlabels{\detokenize{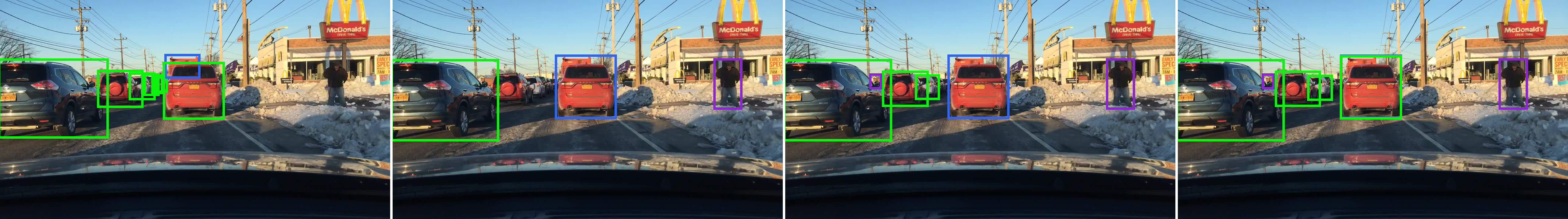}}\vspace{-0.01em}
\quadrow{\detokenize{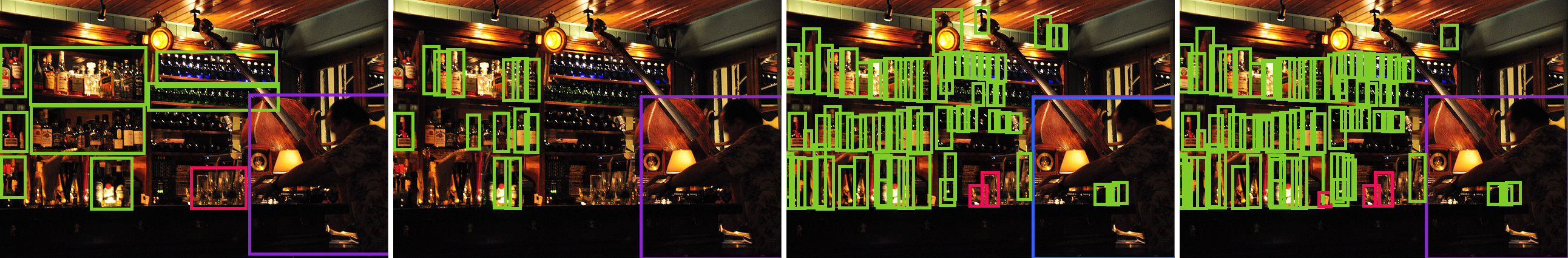}}\vspace{-0.01em}
\caption{\textbf{YW detections across different approaches}: Each column corresponds to a different method: (a) GT (Ground Truth), (b) ZS (Zero-Shot), (c) Adapter, and (d) \textsc{VLOD-TTA}. Each color represents a different object category.}
\label{fig:detection_viz}
\endgroup
\vspace{-.4cm}
\end{figure*}

%% file: 5-conclusion.tex
\section{Conclusion}
TTA has emerged as a practical strategy for improving robustness under domain shift, yet it remains largely unexplored for VLODs. In this paper, we introduce \textsc{VLOD-TTA}, a TTA framework for VLODs that combines IoU-weighted entropy minimization with image-conditioned prompt selection while updating only lightweight adapter parameters.
We validate \textsc{VLOD-TTA} across diverse distribution shifts on two representative VLODs, YOLO-World and Grounding DINO. Across these settings, \textsc{VLOD-TTA} consistently outperforms standard TTA baselines and the prior state-of-the-art VLOD TTA method while maintaining low adaptation overhead.

Despite these strong and consistent gains, IoU-weighted entropy can be less effective in scenes dominated by many small objects with limited proposal overlap, such as Cityscapes. 
Although \textsc{VLOD-TTA} substantially reduces adaptation latency relative to the prior method, it remains slower than zero-shot due to backpropagation at test time. Inspired by recent TTA strategies explored for VLMs, future work will investigate gradient-free adaptation for VLODs to further reduce latency.\\
\noindent\textbf{Supplementary material. } 
It provides additional theoretical analysis, implementation details, Cityscapes failure case analysis, hyperparameter and design ablations, robustness studies, extended COCO-C and PASCAL-C results, and additional qualitative visualizations.\\
\noindent\textbf{Acknowledgments. }
This work was supported in part by Distech Controls Inc., the Natural Sciences and Engineering
Research Council of Canada, the Digital Research Alliance of Canada, and MITACS.

%% file: 6-supp.tex
\section{Supplementary Material}

The supplementary material provides additional method details, ablations, empirical analyses, extended benchmark results, and qualitative visualizations for \textsc{VLOD-TTA}. \Cref{sec:method_details} presents Cosine--Euclidean equivalence, baseline details, adapter configuration, and examples of prompts. \Cref{sec:ablation} reports hyperparameter sensitivity analysis, results on batch size, augmentation, and pre-adaptation fine-tuning. \Cref{sec:empirical} analyzes failure cases on Cityscapes, robustness across detector backbones, and evaluates performance on a specialized underwater domain. \Cref{sec:benchmark} provides a detailed comparison with TTAOD-F and comprehensive results on COCO-C and PASCAL-C. Finally, \Cref{sec:qualitative} compares standard entropy with IoU-weighted entropy and presents additional detection visualizations.

\subsection{Method Details}
\label{sec:method_details}

\noindent\textbf{Cosine--Euclidean Equivalence. }
Let $\hat{\mathbf{v}}_i,\hat{\mathbf{e}}_{k,t}\in\mathbb{R}^d$ denote $\ell_2$-normalized region features and prompt embeddings, respectively, such that $\|\hat{\mathbf{v}}_i\|_2=\|\hat{\mathbf{e}}_{k,t}\|_2=1$.
Define the per-proposal cosine similarity as
\[
z_{i,k,t}=\hat{\mathbf{v}}_i^\top \hat{\mathbf{e}}_{k,t}\in[-1,1],
\]
and its image-level average as 
\[
r_{k,t}=\frac{1}{N}\sum_{i=1}^N z_{i,k,t}.
\]

\paragraph{Proposition.}
The mean squared Euclidean distance between the normalized region features and the prompt embeddings satisfies
\[
\frac{1}{N}\sum_{i=1}^N \bigl\|\hat{\mathbf{v}}_i-\hat{\mathbf{e}}_{k,t}\bigr\|_2^2
= 2 - 2\,r_{k,t}.
\]
Hence, maximizing $r_{k,t}$ is equivalent to minimizing the mean squared Euclidean distance.

\paragraph{Proof.}
For each proposal $i$, since $\|\hat{\mathbf{v}}_i\|_2=\|\hat{\mathbf{e}}_{k,t}\|_2=1$, we have
\[
\bigl\|\hat{\mathbf{v}}_i-\hat{\mathbf{e}}_{k,t}\bigr\|_2^2
= \|\hat{\mathbf{v}}_i\|_2^2+\|\hat{\mathbf{e}}_{k,t}\|_2^2 - 2\,\hat{\mathbf{v}}_i^\top\hat{\mathbf{e}}_{k,t}
= 2 - 2\,z_{i,k,t}.
\]
Averaging over all proposals gives
\[
\frac{1}{N}\sum_{i=1}^N \bigl\|\hat{\mathbf{v}}_i-\hat{\mathbf{e}}_{k,t}\bigr\|_2^2
= \frac{1}{N}\sum_{i=1}^N (2-2\,z_{i,k,t})
= 2 - 2\,r_{k,t}.
\]
This proves the claim.\\

\noindent\textbf{Baseline implementation details.} We adapt all classification-based baselines to operate on proposal-level predictions. For TPT, proposals from the corresponding prediction sets are matched using IoU overlap, and the marginal-entropy objective is computed over the matched proposal-level class distributions. We do not use image augmentations for TPT, allowing us to isolate improvements due to prompt tuning from those obtained through test-time augmentation. VPT follows the same proposal-selection and optimization pipeline, but updates visual prompt parameters instead of textual prompts. For DPE, the text and visual caches are constructed from high-confidence proposals. Cache-based logits are computed independently for each proposal and subsequently fused with the corresponding detector logits. For the Adapter baseline, we update the same lightweight adapter parameters used by our method using the standard proposal-level entropy objective. All baselines use a batch size of one and a single adaptation step under the same evaluation protocol as \textsc{VLOD-TTA}.\\

\noindent\textbf{Adapter Placement and Configuration. } For YOLO-World, we insert adapters after every ConvModule in the backbone and neck~\cite{convadapter}. Concretely, for each convolutional block with feature map $x\in\mathbb{R}^{C\times H\times W}$, we append a lightweight residual path consisting of a $1{\times}1$ down-projection to $C/r$ channels, a depthwise $k{\times}k$ convolution, and a $1{\times}1$ up-projection back to $C$, whose output is added to $x$. The final $1{\times}1$ layer is zero-initialized so that the adapter branch outputs zero at initialization. During adaptation, all pre-trained detector weights are frozen, and only the adapter parameters are updated. In our experiments, we use $r{=}4$ and $k{=}3$.

For Grounding DINO, we insert adapters after the output sublayer of every Transformer block in the BERT encoder~\cite{adapter}. Each adapter is a two-layer bottleneck MLP with GELU, added residually to the layer output, with the up-projection zero-initialized to preserve the pre-trained function at the start of adaptation. We use a bottleneck reduction ratio $r$, so for hidden size $d$, the adapter hidden width is $d/r$. In our experiments, we use $r{=}16$. We disable dropout in the language backbone, freeze all BERT weights, and update only the adapter parameters. Text features are computed exactly as in the baseline, using the average of the last $K$ hidden layers, with $K{=}1$ in our experiments.

The corresponding adapter parameter overhead is $\tfrac{2C^{2}}{r}+\tfrac{C}{r}k^{2}$ for a convolution with $C$ output channels, and $\tfrac{2d^{2}}{r}$ for a Transformer layer with hidden size $d$.\\

\noindent\textbf{Prompt Examples. }
Our prompt selection module selects relevant prompts from a pool of candidate prompts. In CLIP~\cite{clip}, prompts often follow generic templates such as ``a photo of \texttt{<class>}'' or ``an origami of \texttt{<class>}''. For VLODs, however, we observe that such templates are often ineffective. Prompts based on synonyms or verb-centric phrases perform slightly better, so we use a GPT model to generate such candidates. The prompts used in our main experiments do not include any dataset-specific cues. In one ablation study, however, we also evaluate dataset-specific prompts. To generate them, we provide a set of training images to the GPT model together with the dataset name. \Cref{tab:prompt_example} shows example prompts for three classes from the ClipArt dataset under the three prompt-generation strategies.

\begin{table}[h]
\centering
\caption{\textbf{Example prompts under different prompt-generation strategies.} Prompt examples for three ClipArt classes under CLIP-style, GPT-generated, and dataset-specific prompt generation.}
\label{tab:prompt_example}

\scriptsize
\setlength{\tabcolsep}{6pt}
\renewcommand{\arraystretch}{0.95}
\begin{adjustbox}{max width=\columnwidth}

\begin{tabular}{c p{0.28\linewidth} p{0.28\linewidth} p{0.28\linewidth}}
\toprule
\textbf{Prompt Strategy} & \multicolumn{1}{c}{\textbf{Aeroplane}} & \multicolumn{1}{c}{\textbf{Bicycle}} & \multicolumn{1}{c}{\textbf{Bird}} \\
\midrule

\multirow{1}{*}[-8em]{CLIP-Style}
& "aeroplane", "a photo of an aeroplane", "a photograph of an aeroplane",
  "an image of an aeroplane", "a picture of an aeroplane", "a close-up photo of an aeroplane",
  "a cropped photo of an aeroplane", "a low-angle photo of an aeroplane", "a high-angle photo of an aeroplane",
  "a side view of an aeroplane", "a front view of an aeroplane", "a rear view of an aeroplane",
  "a black and white photo of an aeroplane", "a blurry photo of an aeroplane", "a bright photo of an aeroplane",
  "a dark photo of an aeroplane"
& "bicycle", "a photo of a bicycle", "a photograph of a bicycle",
  "an image of a bicycle", "a picture of a bicycle", "a close-up photo of a bicycle",
  "a cropped photo of a bicycle", "a low-angle photo of a bicycle", "a high-angle photo of a bicycle",
  "a side view of a bicycle", "a front view of a bicycle", "a rear view of a bicycle",
  "a black and white photo of a bicycle", "a blurry photo of a bicycle", "a bright photo of a bicycle",
  "a dark photo of a bicycle"
& "bird", "a photo of a bird", "a photograph of a bird",
  "an image of a bird", "a picture of a bird", "a close-up photo of a bird",
  "a cropped photo of a bird", "a low-angle photo of a bird", "a high-angle photo of a bird",
  "a side view of a bird", "a front view of a bird", "a rear view of a bird",
  "a black and white photo of a bird", "a blurry photo of a bird", "a bright photo of a bird",
  "a dark photo of a bird" \\

\midrule

\multirow{1}{*}[-3.5em]{GPT-Generated}
& "aeroplane", "an airplane", "a passenger jet", "a commercial airliner", "a propeller plane",
  "a small aircraft", "a jet aircraft", "an aircraft taking off", "an aircraft landing",
  "a plane in flight", "a plane on the runway", "a twin-engine plane", "a private jet",
  "a cargo plane", "a jetliner", "an air transport aircraft"
& "bicycle", "a pedal bicycle", "a road bike", "a mountain bike", "a commuter bicycle",
  "a racing bike", "a city bicycle", "a bike with basket", "a kids bike", "a fixed-gear bike",
  "a folding bicycle", "an electric bicycle", "a touring bike", "a parked bicycle",
  "a BMX bike", "a two-wheeled cycle"
& "bird", "a flying bird", "a small bird", "a songbird", "a seabird", "a waterfowl",
  "a raptor", "a perching bird", "a wading bird", "a wild bird", "a bird in flight",
  "a perched bird", "a migratory bird", "a backyard bird", "a shorebird", "an avian animal"\\

\midrule

\multirow{1}{*}[-4em]{Dataset-specific}
& "aeroplane", "cartoon airplane", "vector airplane", "flat-color airplane", "outlined airplane",
  "clip-art airplane", "airplane icon", "airplane silhouette", "bold-outline airplane",
  "comic-style airplane", "line-art airplane", "solid-fill airplane", "two-tone airplane",
  "SVG-style airplane", "white-background airplane", "no-texture airplane"
& "bicycle", "cartoon bicycle", "vector bicycle", "flat-color bicycle", "outlined bicycle",
  "clip-art bicycle", "bicycle icon", "bicycle silhouette", "bold-outline bicycle",
  "comic-style bicycle", "line-art bicycle", "solid-fill bicycle", "two-tone bicycle",
  "SVG-style bicycle", "white-background bicycle", "no-texture bicycle"
& "bird", "cartoon bird", "vector bird", "flat-color bird", "outlined bird", "clip-art bird",
  "bird icon", "bird silhouette", "bold-outline bird", "comic-style bird", "line-art bird",
  "solid-fill bird", "two-tone bird", "SVG-style bird", "white-background bird", "no-texture bird"\\
\bottomrule
\end{tabular}
\end{adjustbox}

\end{table}

\subsection{Ablation and Sensitivity Studies}
\label{sec:sup_ablation}
\begin{figure*}[t]
  \centering
  
  \begin{subfigure}[t]{0.31\linewidth}
    \centering
    \includegraphics[width=\linewidth]{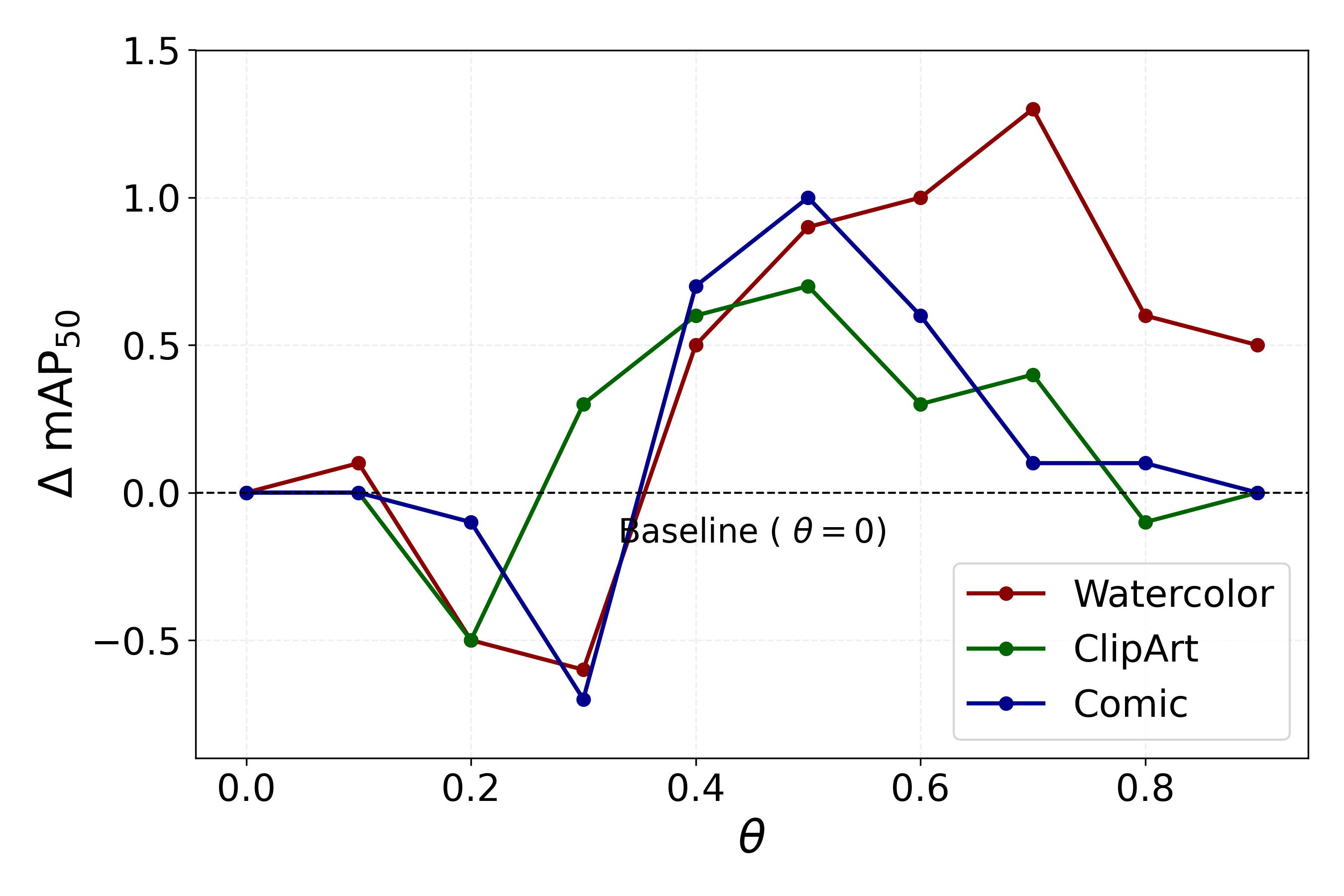}
    \caption{IoU threshold}
  \end{subfigure}\hfill
  \begin{subfigure}[t]{0.31\linewidth}
    \centering
    \includegraphics[width=\linewidth]{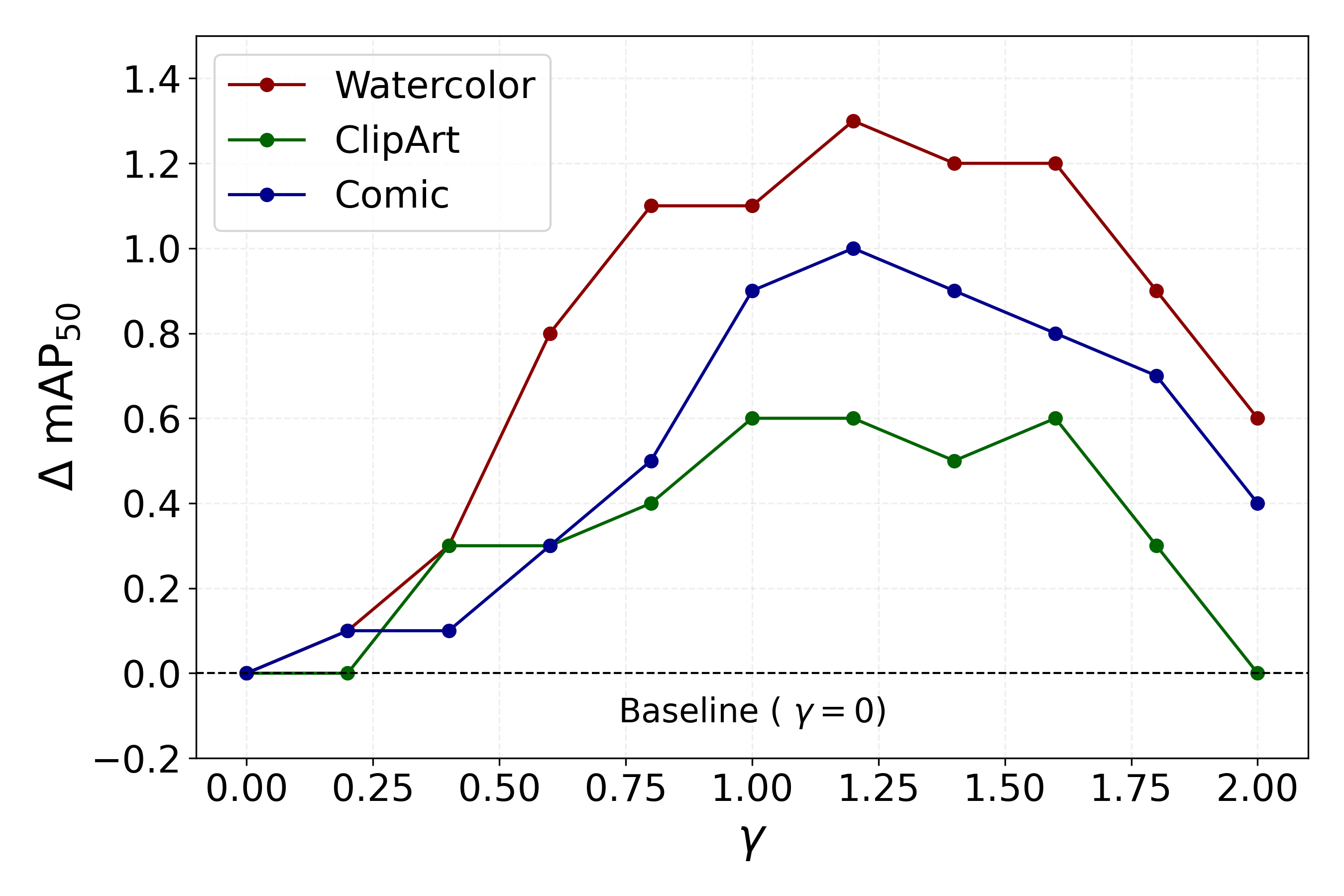}
    \caption{Cluster size exponent}
  \end{subfigure}\hfill
  \begin{subfigure}[t]{0.31\linewidth}
    \centering
    \includegraphics[width=\linewidth]{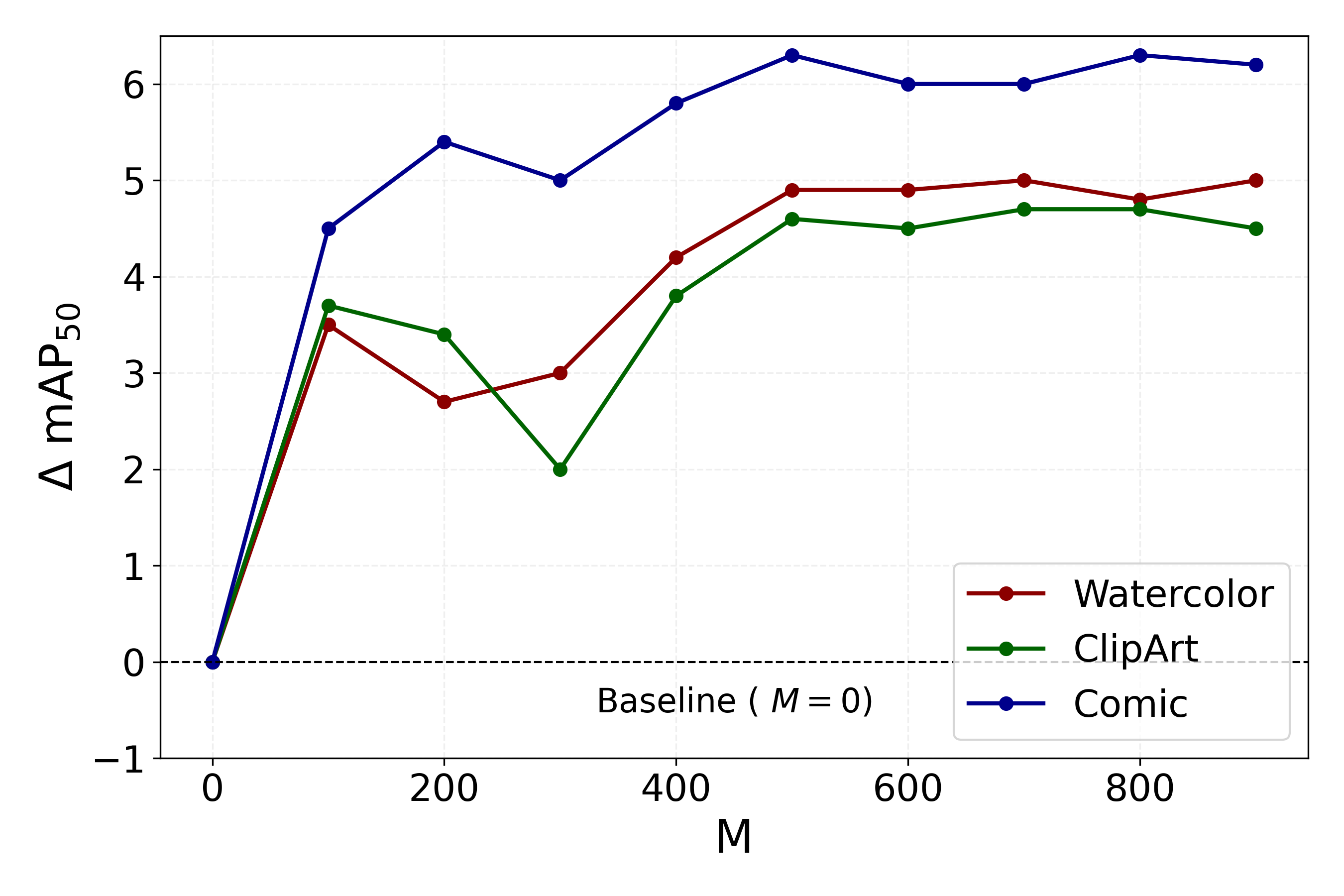}
    \caption{Number of selected proposals}
  \end{subfigure}

  \vspace{0.4em}

  \makebox[\linewidth][c]{%
    \begin{subfigure}[t]{0.31\linewidth}
      \centering
      \includegraphics[width=\linewidth]{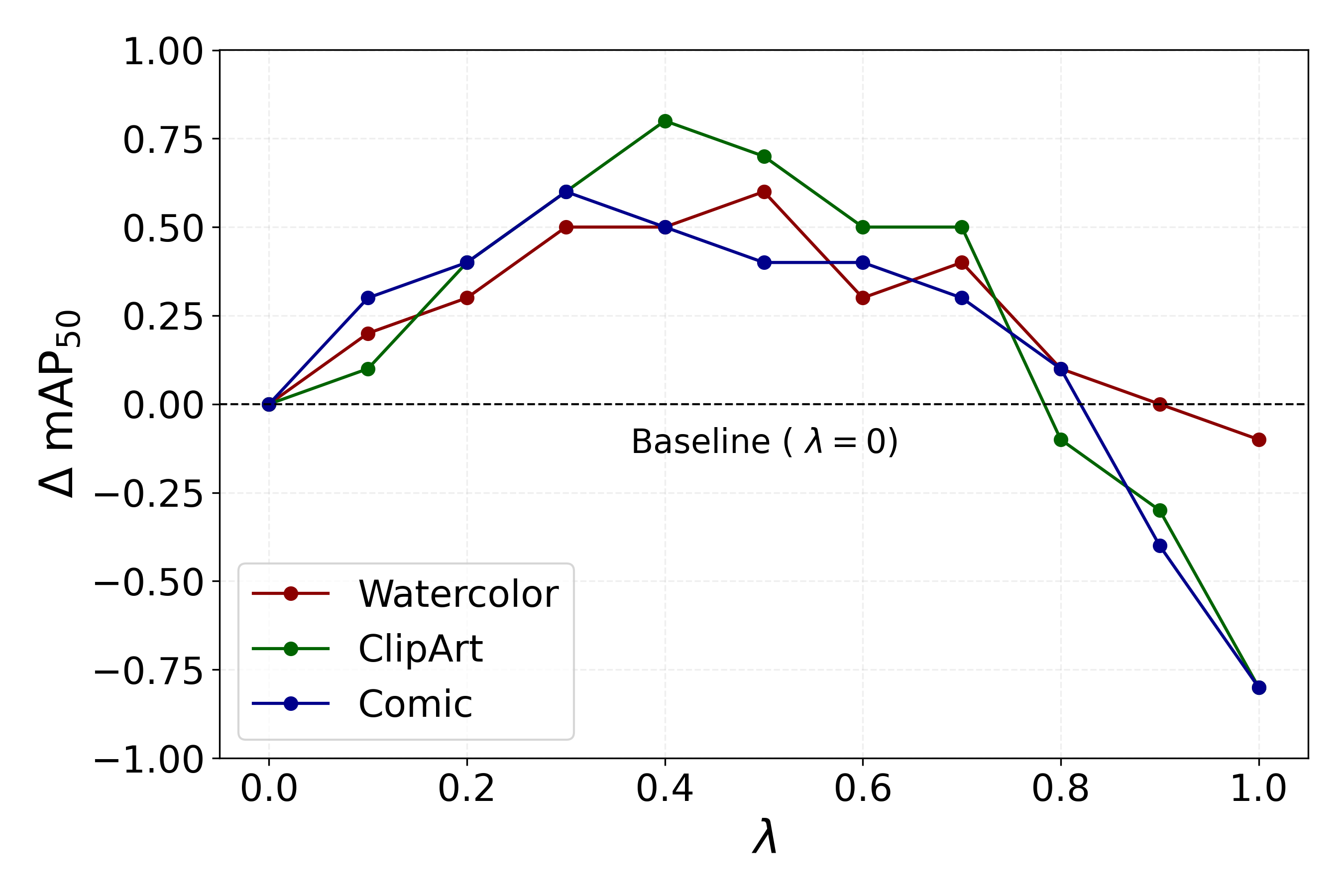}
      \caption{Fusion coefficient}
    \end{subfigure}\hspace{0.04\linewidth}
    \begin{subfigure}[t]{0.31\linewidth}
      \centering
      \includegraphics[width=\linewidth]{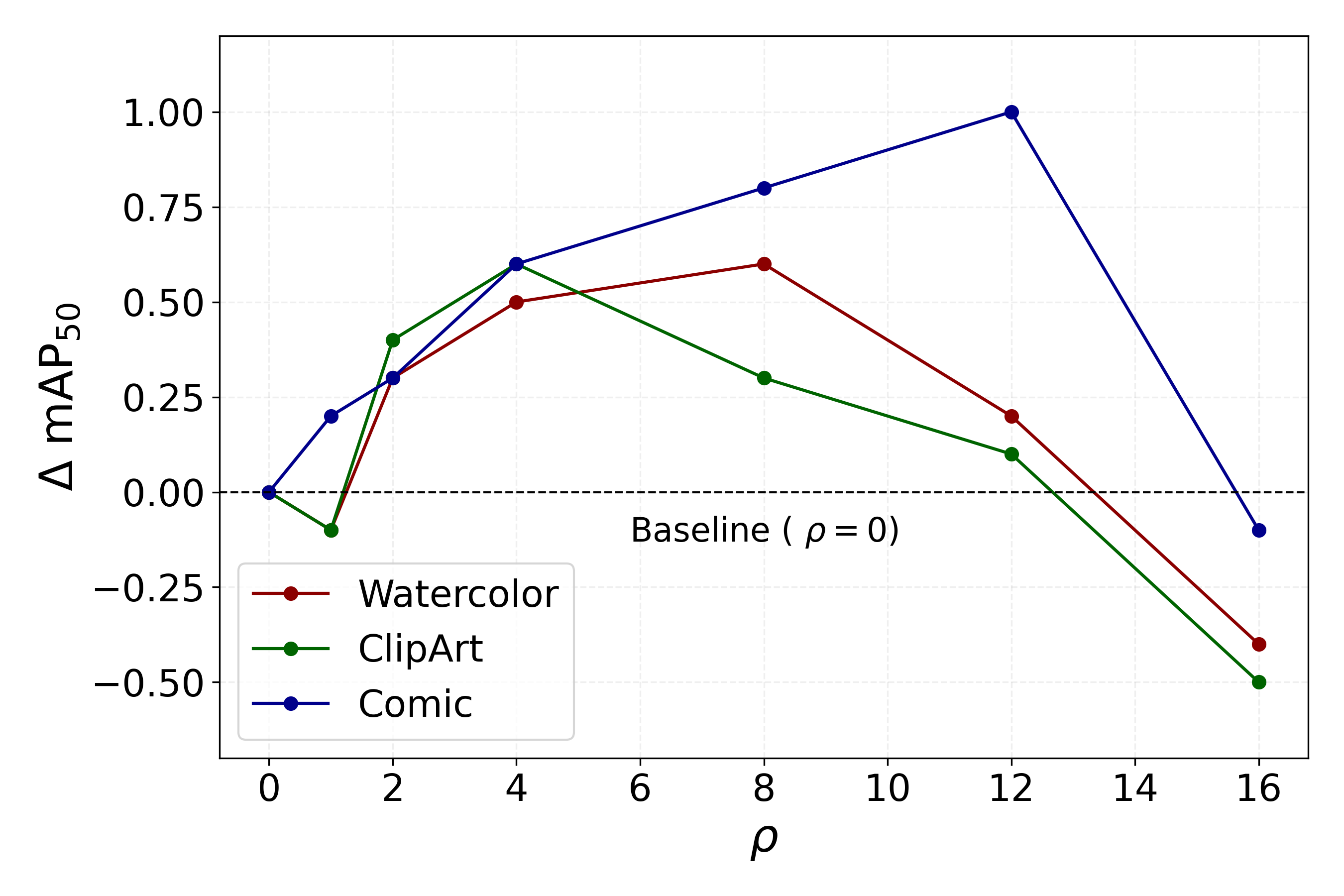}
      \caption{Prompt-selection fraction}
    \end{subfigure}
  }
  
  \caption{\textbf{Sensitivity to hyperparameters on three style-shift datasets.} The IoU threshold ($\theta$), Cluster size exponent ($\gamma$), and number of selected proposals ($M$) affect IWE, while the fusion coefficient ($\lambda$) and prompt-selection fraction ($\rho$) control IPS.}
  \label{fig:hyperparameter}
  \vspace{-.2cm}
\end{figure*}

\noindent\textbf{Hyperparameter Sensitivity Analysis. } We conduct sensitivity analyses on five key hyperparameters that influence the performance of \textsc{VLOD-TTA}. Results on the three style-shift datasets are shown in \Cref{fig:hyperparameter}.

\noindent\textbf{(a) Effect of $\theta$ for graph construction.} The IoU threshold $\theta$ determines how proposals are clustered within each class. As shown in \Cref{fig:hyperparameter}(a), as $\theta\!\to\!0$, the proposals of a class collapse into a single cluster, so the objective behaves similarly to standard entropy. Interestingly, performance drops when $\theta$ is around $0.2$--$0.3$, likely because meaningful clusters are not yet formed at these thresholds. As $\theta\!\to\!1$, proposals are rarely grouped, which reduces the advantage of our method. Empirically, Watercolor, which contains slightly larger objects, achieves its best performance around $\theta \approx 0.7$, whereas ClipArt and Comic, which contain smaller objects on average, peak near $\theta \approx 0.5$. Overall, although the optimal value shows mild dataset dependence, performance is generally stable for $0.5 \le \theta \le 0.7$.

\noindent\textbf{(b) Effect of $\gamma$ for graph construction.} The exponent $\gamma$ controls how strongly component size influences the IoU-weighted entropy. When $\gamma{=}0$, all $w_i$ are equal and the objective reduces to standard entropy. As shown in \Cref{fig:hyperparameter}(b), performance improves from $\gamma{=}0$ and peaks around $\gamma{\approx}1.0$--$1.2$ across datasets. For very large $\gamma$, performance drops because large clusters can suppress small but correct objects. Overall, $\gamma\in[0.6,1.6]$ is stable across datasets, with $\gamma{\approx}1.0$--$1.2$ performing best in our experiments.

\noindent\textbf{(c) Effect of top-$M$ proposal selection.} The number of selected proposals, $M$, controls how many high-confidence boxes are used to construct the IoU graph. As shown in \Cref{fig:hyperparameter}(c), when $M$ is too small, improvements over ZS are limited because the resulting graph fails to capture the overall proposal structure. Increasing $M$ improves performance by retaining more informative proposals while still filtering out extremely noisy boxes. Performance peaks around $M=600$, after which further increases yield little additional benefit.

\noindent\textbf{(d) Effect of $\lambda$ in prompt selection.} The fusion coefficient $\lambda$ balances the selected-prompt score $\tilde{z}_{i,k}$ and the original detector score $s_{i,k}$. As shown in \Cref{fig:hyperparameter}(d), performance increases from $\lambda=0$ and peaks between $\lambda \approx 0.3$ and $0.5$, depending on the dataset. Beyond this range, performance decreases steadily and drops sharply as $\lambda\!\to\!1$. This decline arises because early visual--text fusion in VLODs makes the region features partly dependent on the text embeddings, so relying only on selected prompts discards useful information carried by the original detector prompts. For GD, where fusion occurs at multiple stages, the original detector score is even more important, and the best performance is obtained at $\lambda=0.1$.

\noindent\textbf{(e) Effect of $\rho$ in prompt selection.} For each class, we retain the top-$\rho$ fraction of prompts according to their similarity scores. As shown in \Cref{fig:hyperparameter}(e), increasing $\rho$ from $0$ initially improves performance by incorporating more informative templates, after which the gains saturate and eventually decline as weaker templates are included. In general, very small $\rho$ underuses the prompt pool, very large $\rho$ introduces noise, and $\rho \approx 0.25$--$0.5$ works best.\\

\begin{figure*}[!h]
  \centering
  \begin{subfigure}[t]{0.48\linewidth}
    \centering
    \includegraphics[width=\linewidth]{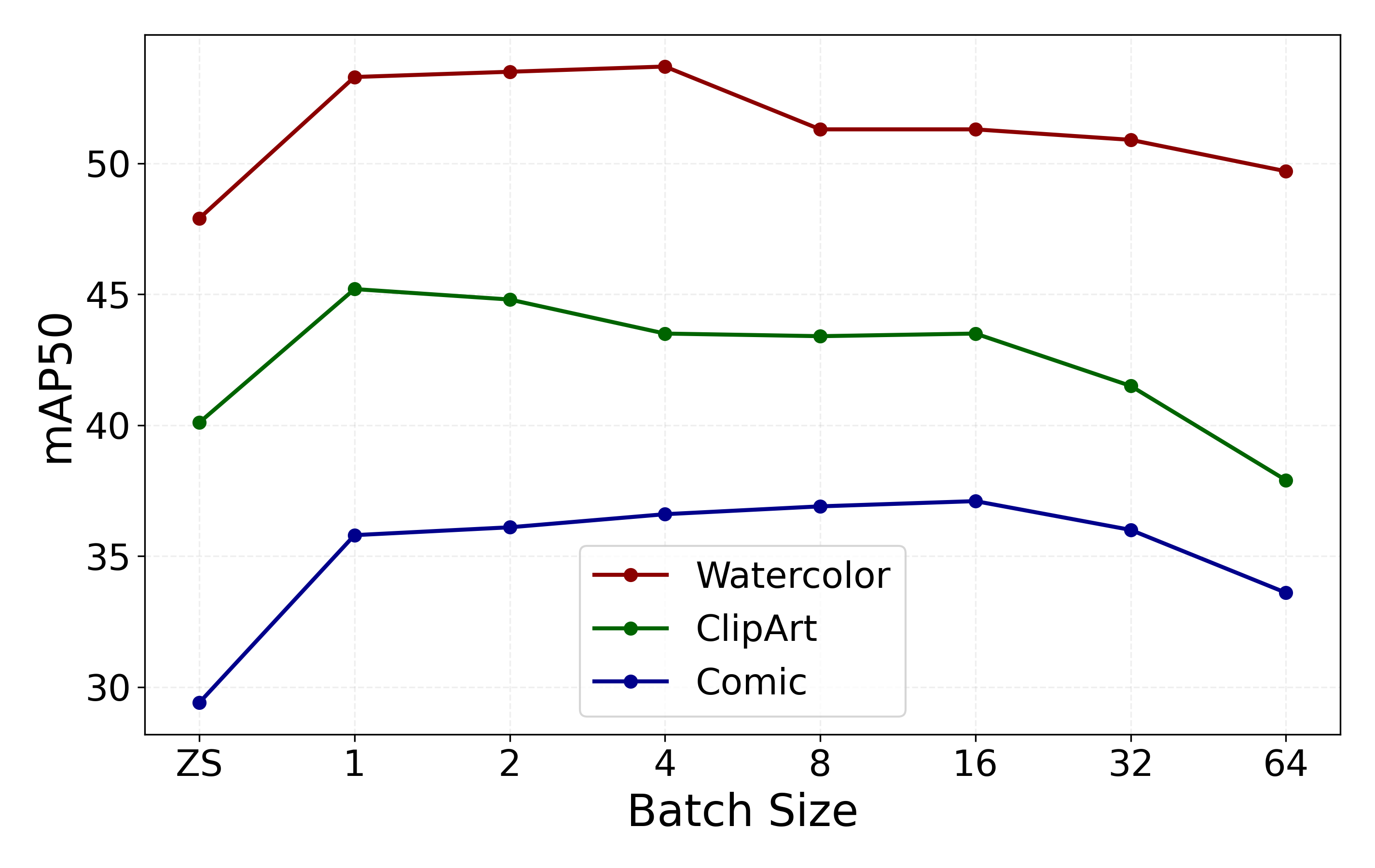}
    \caption{Effect of batch size.}
    \label{fig:batch-size}
  \end{subfigure}
  \hfill
  \begin{subfigure}[t]{0.48\linewidth}
    \centering
    \includegraphics[width=\linewidth]{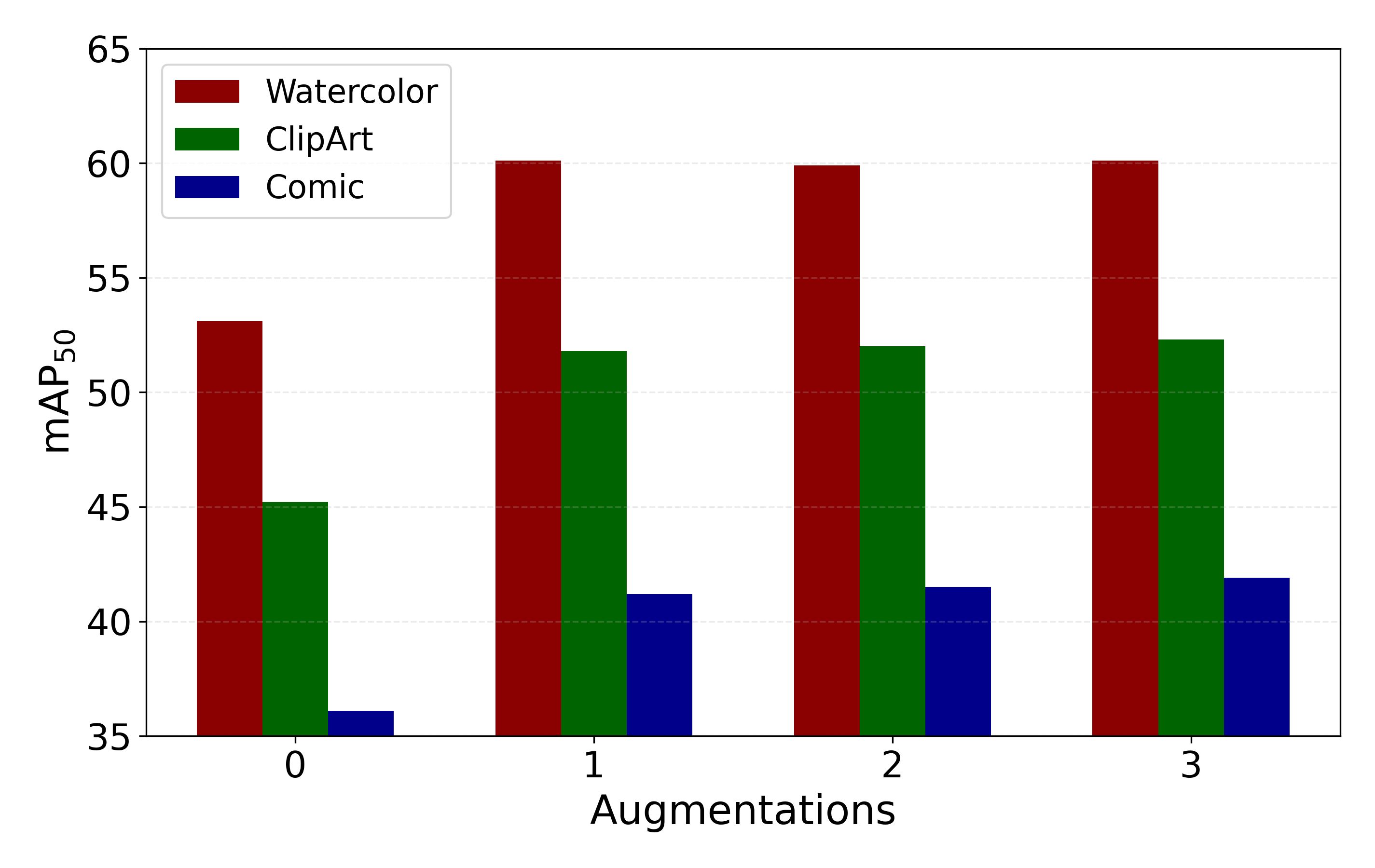}
    \caption{Effect of the number of augmentations.}
    \label{fig:augmentation}
  \end{subfigure}
  \caption{\textbf{Effect of batch size and the number of augmentations on performance.} We report AP$_{50}$ on three style-shift datasets using YOLO-World.}
  \label{fig:batch-vs-aug}
\end{figure*}

\noindent\textbf{Effect of Batch Size. } We use a batch size of $1$ in our main experiments because it reflects a practical TTA setting. In \Cref{fig:batch-size}, we ablate over batch size. Across all three datasets, we observe a similar trend. Performance rises slightly as the batch size increases to about $4$–$8$, which indicates that our approach is not restricted to a batch size of $1$. Beyond a batch size of $16$, performance drops slightly. A likely reason is that the growing number of proposals makes entropy minimization less selective, so the optimization struggles to focus on the correct classes. 

\noindent\textbf{Effect of Augmentations. } In this section, we study how augmentations affect the performance of our method. In these experiments, we use only scale augmentations, which we found to be the most effective in preliminary tests. Augmentations are added in the order that performed best in those preliminary tests. As shown in \Cref{fig:augmentation}, adding a single augmentation improves AP$_{50}$ by $+6.0$, $+6.6$, and $+5.1$ on Watercolor, ClipArt, and Comic, respectively. This indicates that our approach benefits from modest augmentation. Adding more than one augmentation does not yield further gains in our setting, although tailoring the augmentation type and magnitude to each dataset may lead to larger improvements.\\

\noindent\textbf{Effect of Fine-tuning VLODs before Adaptation. } Prior TTA methods for ODs~\cite{stfar,fullytesttimeOD} first fine-tune the detector on a source domain that is closer to the target domain before adaptation. For VLODs, however, this step is not necessarily required, since they already exhibit strong ZS performance on most datasets. We therefore study the effect of fine-tuning before adaptation and assess whether it is beneficial for VLODs. We consider two settings. First, the source domain is PASCAL VOC~\cite{pascal_voc}, and the target domains are Watercolor, Comic, and ClipArt~\cite{inoue2018crossdomain}. This setting is common in OD domain adaptation and introduces a substantial domain shift that challenges adaptation. Second, the source domain is COCO~\cite{lin2014microsoft}, and the target domain is COCO-C. This setting is less challenging because adaptation remains within the same dataset, and only the synthetic corruptions differ.

The results for the two settings are reported in \Cref{tab:ft-pascal,tab:ft-coco}. We observe two distinct patterns. Fine-tuning on PASCAL VOC reduces the model's generalization ability, leading to lower performance on the three style-shift domains. Although our method improves over ZS in both cases by a similar margin, the absolute AP after fine-tuning remains lower than without fine-tuning. In contrast, fine-tuning on COCO improves ZS performance on COCO-C. This is likely because the domain shift between source and target is small and the training set is large. Our method further improves over ZS in this setting, which confirms its effectiveness. Overall, fine-tuning VLODs is beneficial when training data are abundant and the target domain is close to the source, but it introduces an additional training step and extra computational cost.

\begin{table*}[h]
\centering
\caption{\textbf{Effect of fine-tuning before adaptation on YOLO-World.} In the No Fine-Tune setting, the pretrained YOLO-World model is adapted directly. In the Fine-Tune setting, the pretrained model is first fine-tuned on PASCAL VOC and then adapted. We report ZS and \textsc{VLOD-TTA} performance on three style-shift datasets. AP$_{50}$ on PASCAL VOC for both settings is reported in the top row.}
\label{tab:ft-pascal}
\scriptsize
\setlength{\tabcolsep}{2.4pt}
\renewcommand{\arraystretch}{0.95}
\begin{adjustbox}{max width=\textwidth}
\begin{tabular}{l *{9}{S[table-format=2.1]} | l *{3}{S[table-format=2.1]}}
\toprule

\multicolumn{13}{c}{\textbf{No Fine-Tune} \; {\tiny (PASCAL VOC AP$_{50}$ = 78.6)}}\\
\addlinespace[0.25em]
\cmidrule(lr){1-13}

& \multicolumn{3}{c}{\textbf{Watercolor}}
& \multicolumn{3}{c}{\textbf{ClipArt}}
& \multicolumn{3}{c}{\textbf{Comic}}
& \multicolumn{3}{c}{\textbf{Avg}}\\
\cmidrule(lr){2-4}\cmidrule(lr){5-7}\cmidrule(lr){8-10}\cmidrule(lr){11-13}
\textbf{Method}
& \textbf{mAP} & \textbf{AP$_{50}$} & \textbf{AP$_{75}$}
& \textbf{mAP} & \textbf{AP$_{50}$} & \textbf{AP$_{75}$}
& \textbf{mAP} & \textbf{AP$_{50}$} & \textbf{AP$_{75}$}
& \textbf{mAP} & \textbf{AP$_{50}$} & \textbf{AP$_{75}$} \\
\midrule
ZS
& 26.9 & 47.9 & 25.9
& 24.4 & 40.1 & 26.2
& 17.8 & 29.4 & 18.8
& 23.0 & 39.1 & 23.6 \\
\rowcolor{gray!12}
\textbf{\textsc{VLOD-TTA}}
& \textbf{29.6} & \textbf{53.1} & \textbf{28.7}
& \textbf{28.1} & \textbf{45.2} & \textbf{29.9}
& \textbf{21.4} & \textbf{36.1} & \textbf{22.1}
& \textbf{26.4} & \textbf{44.8} & \textbf{26.9} \\

\toprule
\multicolumn{13}{c}{\textbf{Fine-Tune} \; {\tiny (PASCAL VOC AP$_{50}$ = 82.3)}}\\
\addlinespace[0.25em]
\cmidrule(lr){1-13}

& \multicolumn{3}{c}{\textbf{Watercolor}}
& \multicolumn{3}{c}{\textbf{ClipArt}}
& \multicolumn{3}{c}{\textbf{Comic}}
& \multicolumn{3}{c}{\textbf{Avg}}\\
\cmidrule(lr){2-4}\cmidrule(lr){5-7}\cmidrule(lr){8-10}\cmidrule(lr){11-13}
\textbf{Method}
& \textbf{mAP} & \textbf{AP$_{50}$} & \textbf{AP$_{75}$}
& \textbf{mAP} & \textbf{AP$_{50}$} & \textbf{AP$_{75}$}
& \textbf{mAP} & \textbf{AP$_{50}$} & \textbf{AP$_{75}$}
& \textbf{mAP} & \textbf{AP$_{50}$} & \textbf{AP$_{75}$} \\
\midrule
ZS
& 25.3 & 44.3 & 25.8
& 23.9 & 39.3 & 25.2
& 15.3 & 24.8 & 16.3
& 21.5 & 36.1 & 22.4 \\
\rowcolor{gray!12}
\textbf{\textsc{VLOD-TTA}}
& \textbf{27.8} & \textbf{49.8} & \textbf{27.8}
& \textbf{25.6} & \textbf{44.2} & \textbf{28.6}
& \textbf{19.1} & \textbf{31.1} & \textbf{19.2}
& \textbf{24.2} & \textbf{41.7} & \textbf{25.2} \\
\bottomrule
\end{tabular}
\end{adjustbox}
\end{table*}

\begin{table*}[h]
\centering
\caption{\textbf{Effect of fine-tuning before adaptation on YOLO-World.} In the No Fine-Tune setting, the pretrained YOLO-World model is adapted directly. In the Fine-Tune setting, the pretrained model is first fine-tuned on COCO and then adapted. We report ZS and \textsc{VLOD-TTA} performance on 15 different corruptions. AP$_{50}$ on COCO for both settings is reported in the top row.}
\label{tab:ft-coco}

\scriptsize
\setlength{\tabcolsep}{2.4pt}
\renewcommand{\arraystretch}{0.95}
\begin{adjustbox}{max width=\textwidth}
\begin{tabular}{
  l
  *{15}{S[table-format=2.1]} | S[table-format=2.1]
}
\toprule

\multicolumn{17}{c}{\textbf{No Fine-Tune}\; {\tiny (COCO
 AP$_{50}$ = 51.9)}}\\
\addlinespace[0.25em]
\cmidrule(lr){1-17}

& \multicolumn{3}{c}{\textbf{Noise}}
& \multicolumn{4}{c}{\textbf{Blur}}
& \multicolumn{3}{c}{\textbf{Weather}}
& \multicolumn{5}{c}{\textbf{Digital}}
& {\textbf{Avg}}\\
\cmidrule(lr){2-4}\cmidrule(lr){5-8}\cmidrule(lr){9-11}\cmidrule(lr){12-16}\cmidrule(lr){17-17}
\textbf{Method}
& \textbf{Gauss} & \textbf{Shot} & \textbf{Impul}
& \textbf{Defoc} & \textbf{Glass} & \textbf{Motn} & \textbf{Zoom}
& \textbf{Snow} & \textbf{Frost} & \textbf{Fog}
& \textbf{Brit} & \textbf{Contr} & \textbf{Elast} & \textbf{Pixel} & \textbf{JPEG}
& \textbf{Avg}
\\
\midrule
ZS
& 7.8 & 7.4 & 6.7
& 22.6 & 6.1 & 13.4 & 10.1
& 23.5 & 32.0 & 45.9
& 47.3 & 27.5 & 30.2 & 6.2 & 14.0
& 20.0
\\
\rowcolor{gray!12}\textbf{\textsc{VLOD-TTA}}
& \textbf{9.3} & \textbf{10.2} & \textbf{8.9}
& \textbf{25.2} & \textbf{8.9} & \textbf{14.8} & \textbf{11.8}
& \textbf{25.6} & \textbf{36.2} & \textbf{48.1}
& \textbf{49.1} & \textbf{30.7} & \textbf{34.1} & \textbf{17.5} & \textbf{19.6}
& \textbf{23.3}
\\

\toprule
\multicolumn{17}{c}{\textbf{Fine-Tune}\; {\tiny (COCO
 AP$_{50}$ = 57.8)}}\\
\addlinespace[0.25em]
\cmidrule(lr){1-17}

& \multicolumn{3}{c}{\textbf{Noise}}
& \multicolumn{4}{c}{\textbf{Blur}}
& \multicolumn{3}{c}{\textbf{Weather}}
& \multicolumn{5}{c}{\textbf{Digital}}
& {\textbf{Avg}}\\
\cmidrule(lr){2-4}\cmidrule(lr){5-8}\cmidrule(lr){9-11}\cmidrule(lr){12-16}\cmidrule(lr){17-17}
\textbf{Method}
& \textbf{Gauss} & \textbf{Shot} & \textbf{Impul}
& \textbf{Defoc} & \textbf{Glass} & \textbf{Motn} & \textbf{Zoom}
& \textbf{Snow} & \textbf{Frost} & \textbf{Fog}
& \textbf{Brit} & \textbf{Contr} & \textbf{Elast} & \textbf{Pixel} & \textbf{JPEG}
& \textbf{Avg}
\\
\midrule
ZS
& 13.7 & 13.3 & 12.6
& 26.2 & 9.6 & 18.0 & 11.9
& 27.9 & 37.2 & 52.1
& 52.9 & 30.3 & 34.5 & 12.7 & 19.4
& 24.8
\\
\rowcolor{gray!12}\textbf{\textsc{VLOD-TTA}}
& \textbf{15.5} & \textbf{14.6} & \textbf{14.5}
& \textbf{27.8} & \textbf{{12.6}} & \textbf{19.6} & \textbf{12.9}
& \textbf{29.2} & \textbf{38.8} & \textbf{53.5}
& \textbf{53.8} & \textbf{33.2} & \textbf{36.7} & \textbf{19.3} & \textbf{24.1}
& \textbf{27.1}
\\
\bottomrule
\end{tabular}
\end{adjustbox}
\end{table*}

\subsection{Additional Empirical Analysis}
\label{sec:empirical}

\noindent\textbf{Cityscapes Failure-Case Analysis. } Our method underperforms on the Cityscapes dataset, so we conduct a detailed ablation study to identify the main causes. We find that two factors are primarily responsible: label overlap between \textit{rider} and \textit{person}, and the large proportion of small objects.

\begin{figure}[h]
\centering

\begin{minipage}{\linewidth}
  \centering
  {\small
\makebox[\linewidth]{%
    \hspace{0.2\linewidth}\textbf{(a) GT}\hfill\textbf{(b) ZS}\hspace{0.2\linewidth}%
}\\[0.3em]
}
  \includegraphics[width=\linewidth]{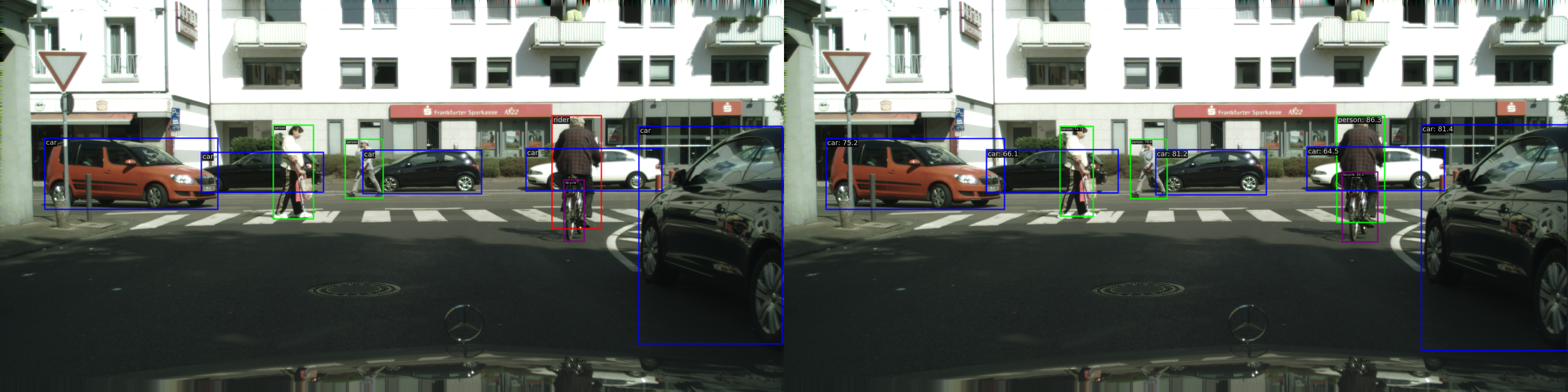}
\end{minipage}

\vspace{0.1em}

\begin{minipage}{\linewidth}
  \centering
  \includegraphics[width=\linewidth]{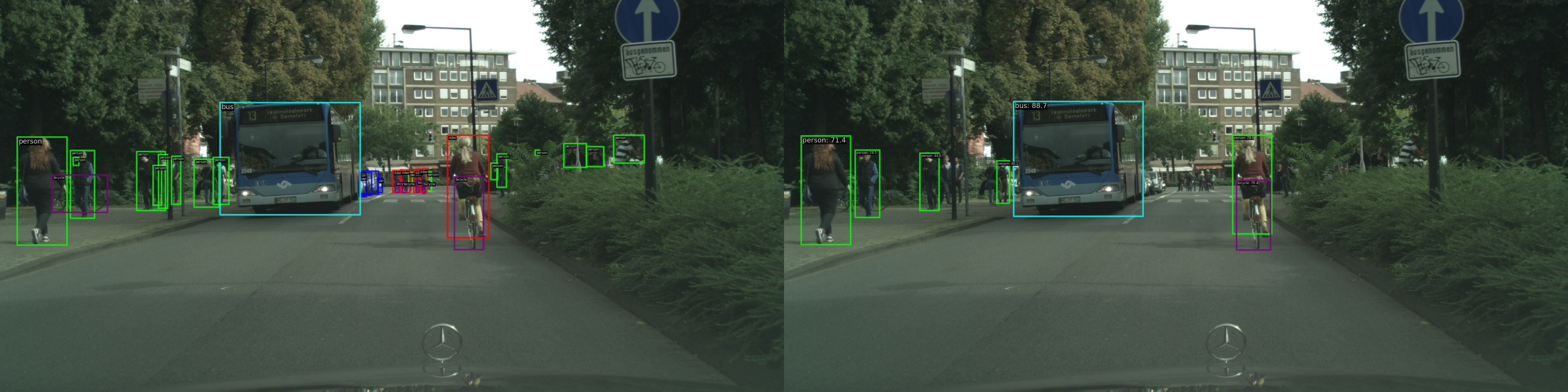}
\end{minipage}

\caption{\textbf{Cityscapes detections with ground-truth (GT) and zero-shot (ZS).} The ZS model detects \textit{rider} (red) as \textit{person} (green).}
\label{fig:cityscapes_rider_vis}
\end{figure}

\noindent\textbf{1. Label overlap between \textit{rider} and \textit{person}.}
Cityscapes contains eight classes, including \textit{rider} and \textit{person}. In practice, riders are visually very similar to persons, and the ZS model is biased toward the \textit{person} label. As a result, a rider is often localized correctly but classified as \textit{person}, as shown in \Cref{fig:cityscapes_rider_vis}. Our IoU-weighted entropy (IWE) can further reinforce this behavior by sharpening high-IoU proposal clusters, which pushes ambiguous rider--person cases toward the \textit{person} label. Under the original annotation protocol, these predictions are still counted as errors even when localization is correct. This effect is reflected in \Cref{tab:cityscapes_rider_person}, where we report class-wise AP for \textit{person} and \textit{rider}. Compared with ZS, \textsc{VLOD-TTA} reduces AP for both classes.

\begin{figure}[t]
\centering

\begin{minipage}[t]{0.48\textwidth}
\vspace{0pt}
\centering
\captionof{table}{\textbf{Class-wise AP on Cityscapes.} AP for person and rider before and after merging rider (MR) into person.}
\label{tab:cityscapes_rider_person}

\scriptsize
\setlength{\tabcolsep}{3pt}
\renewcommand{\arraystretch}{0.95}

\begin{adjustbox}{max width=\linewidth}
\begin{tabular}{lcc}
\toprule
\textbf{Method} & \textbf{Person} & \textbf{Rider} \\
\midrule
ZS                          & 17.8 & 6.3 \\
VLOD-TTA                    & 16.5 & 1.9 \\
\midrule
ZS (MR)            & 23.2 & -- \\
\rowcolor{gray!15}
\textbf{VLOD-TTA (MR)} & \textbf{24.4} & -- \\
\bottomrule
\end{tabular}
\end{adjustbox}
\end{minipage}
\hfill
\begin{minipage}[t]{0.48\textwidth}
\vspace{0pt}
\centering
\captionof{table}{\textbf{Effect of merging rider (MR) into person on Cityscapes.} Detection performance with the 8-class labels and with MR into person.}
\label{tab:cityscapes_merged}

\scriptsize
\setlength{\tabcolsep}{3pt}
\renewcommand{\arraystretch}{0.95}

\begin{adjustbox}{max width=\linewidth}
\begin{tabular}{lccc}
\toprule
\textbf{Method} & \textbf{mAP} & \textbf{AP$_{50}$} & \textbf{AP$_{75}$} \\
\midrule
ZS              & 18.8 & 31.0 & 17.9 \\
VLOD-TTA        & 19.4 & 31.8 & 18.6 \\
\midrule
ZS (MR)         & 21.3 & 34.7 & 20.5 \\
\rowcolor{gray!15}
\textbf{VLOD-TTA (MR)} & \textbf{22.5} & \textbf{35.9} & \textbf{21.1} \\
\bottomrule
\end{tabular}
\end{adjustbox}
\end{minipage}
\end{figure}

\noindent\textbf{Ablation. }
We merge the \textit{rider} class into \textit{person}, producing a 7-class annotation set. As shown in \Cref{tab:cityscapes_merged}, the ZS baseline improves after this relabeling, and the class-wise AP in \Cref{tab:cityscapes_rider_person} confirms that this improvement corresponds to a higher AP for the merged \textit{person} class. \textsc{VLOD-TTA} improves further in this setting because IWE no longer needs to distribute probability mass between \textit{rider} and \textit{person}. Instead, entropy minimization concentrates on a unified \textit{person} cluster and yields larger gains.\\

\begin{figure}[t]
\centering

\begin{minipage}{\linewidth}
  \centering
  {\small
  \makebox[\linewidth]{%
    \makebox[0.333\linewidth][c]{\textbf{(a) ZS}}%
    \makebox[0.333\linewidth][c]{\textbf{(b) VLOD-TTA (640)}}%
    \makebox[0.333\linewidth][c]{\textbf{(c) VLOD-TTA (1024)}}%
  }\\[0.3em]
  }
  \includegraphics[width=\linewidth]{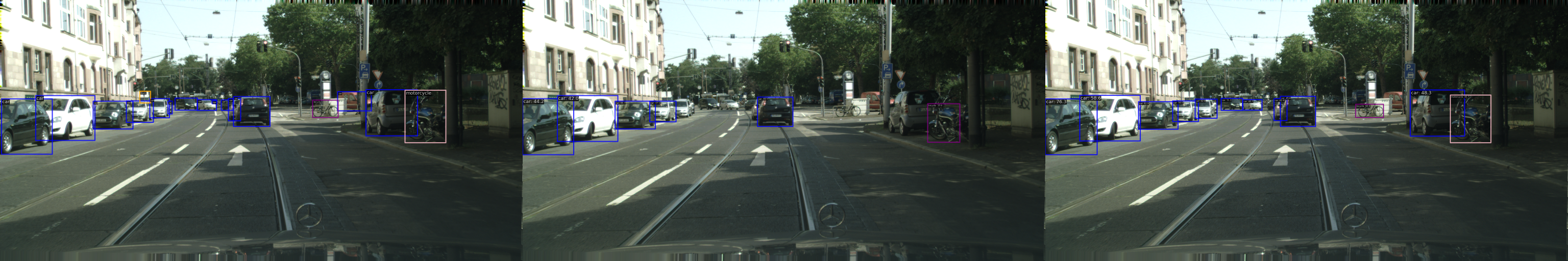}
\end{minipage}

\vspace{0.1em}

\begin{minipage}{\linewidth}
  \centering
  \includegraphics[width=\linewidth]{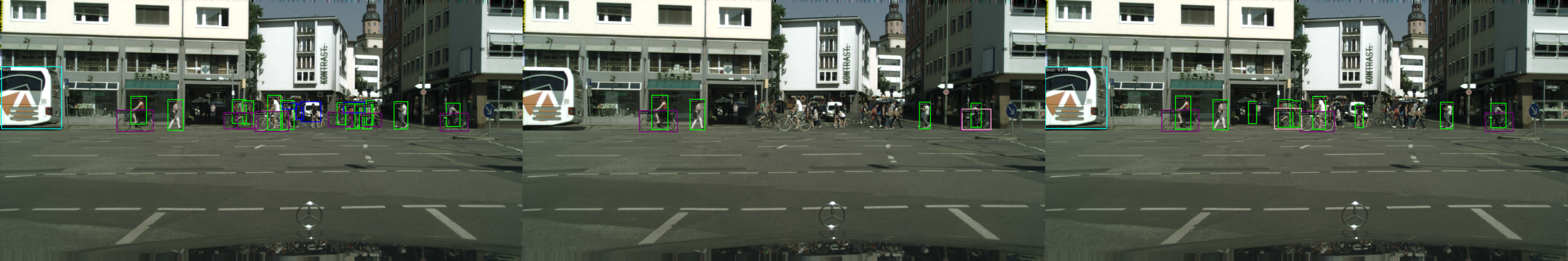}
\end{minipage}

\caption{\textbf{Cityscapes detections with zero-shot (ZS) and \textsc{VLOD-TTA} at image scales 640$\times$640 and 1024$\times$1024.} \textsc{VLOD-TTA} at 1024$\times$1024 detects more small objects than at 640$\times$640.}
\label{fig:cityscapes_scale_vis}
\end{figure}

\noindent\textbf{2. Small objects and input resolution.}
Cityscapes contains many small objects, and its original image resolution is 2048$\times$1024. YOLO-World resizes inputs to 640$\times$640 by default, which further shrinks distant cars and pedestrians and leads to missed detections. At this resolution, many small objects do not generate sufficiently stable overlapping proposals, which limits the effectiveness of IWE. Qualitative examples are shown in \Cref{fig:cityscapes_scale_vis}.

\noindent\textbf{Ablation. }
We evaluate the model at an input resolution of 1024$\times$1024. As shown in \Cref{tab:cityscapes_resolution}, the zero-shot baseline improves substantially at higher resolution. \textsc{VLOD-TTA} then yields additional gains, with a modest improvement at 640$\times$640 (+0.6 mAP) and a much larger improvement at 1024$\times$1024 (+2.7 mAP). At higher resolution, more small objects produce stable overlapping proposals that IWE can effectively weight and sharpen during adaptation. This confirms that small objects are a key factor behind the original underperformance, as illustrated in \Cref{fig:cityscapes_scale_vis}.

\begin{table}[t]
\centering
\caption{\textbf{Effect of input resolution on Cityscapes.} Detection performance at input resolutions of 640$\times$640 and 1024$\times$1024.}
\label{tab:cityscapes_resolution}
\scriptsize
\setlength{\tabcolsep}{4pt}
\renewcommand{\arraystretch}{0.95}
\begin{tabular}{lccc}
\toprule
\textbf{Method} & \textbf{mAP} & \textbf{AP$_{50}$} & \textbf{AP$_{75}$} \\
\midrule
ZS (640)                  & 18.8 & 31.0 & 17.9 \\
VLOD-TTA (640)            & 19.4 & 31.8 & 18.6 \\
\midrule
ZS (1024)                 & 28.5 & 43.9 & 28.3 \\
\rowcolor{gray!15}
\textbf{VLOD-TTA (1024)}  & \textbf{31.2} & \textbf{46.6} & \textbf{30.5} \\
\bottomrule
\end{tabular}
\end{table}

\noindent\textbf{Conclusion.} The relatively lower performance on Cityscapes stems from label overlap between \textit{rider} and \textit{person}, as well as the prevalence of small objects at low input resolution. After merging \textit{rider} into \textit{person} and increasing the input resolution, \textsc{VLOD-TTA} yields substantially larger gains on Cityscapes.

\noindent\textbf{Variation in Performance across Backbones. } In \Cref{tab:backbone}, we evaluate the effect of the detector backbone by applying our method to YOLO-World-Large (YW-L) and Grounding DINO-Big (GD-B). Across Watercolor, ClipArt, and Comic, both models show consistent improvements over their ZS baselines in mAP, AP$_{50}$, and AP$_{75}$. Although GD-B achieves higher absolute scores, the relative gains from adaptation are similar for both backbones, indicating that our method is not tied to a specific architecture. Overall, these results demonstrate that IoU-weighted entropy and image-conditioned prompt selection generalize well across backbones.

\begin{table*}[h]
\centering
\captionof{table}{\textbf{Detection performance with YOLO-World-L and Grounding DINO-B.}
We report performance compared with ZS on three style-shift datasets.}
\label{tab:backbone}
\scriptsize
\setlength{\tabcolsep}{2.4pt}

\begin{adjustbox}{max width=\linewidth}
\begin{tabular}{l *{9}{S[table-format=2.1]}}
\toprule

\multicolumn{10}{c}{\textbf{YOLO-World-L}}\\
\addlinespace[0.25em]
\cmidrule(lr){1-10}

& \multicolumn{3}{c}{\textbf{Watercolor}}
& \multicolumn{3}{c}{\textbf{ClipArt}}
& \multicolumn{3}{c}{\textbf{Comic}}\\
\cmidrule(lr){2-4}\cmidrule(lr){5-7}\cmidrule(lr){8-10}
\textbf{Method}
& \textbf{mAP} & \textbf{AP$_{50}$} & \textbf{AP$_{75}$}
& \textbf{mAP} & \textbf{AP$_{50}$} & \textbf{AP$_{75}$}
& \textbf{mAP} & \textbf{AP$_{50}$} & \textbf{AP$_{75}$} \\
\midrule
ZS    & 32.8 & 55.3 & 33.0 & 31.1 & 50.6 & 32.6 & 23.3 & 37.9 & 23.6 \\
\rowcolor{gray!12}
\textbf{\textsc{VLOD-TTA}} & \textbf{34.1} & \textbf{58.3} & \textbf{34.2} & \textbf{33.2} & \textbf{53.9} & \textbf{34.3} & \textbf{26.5} & \textbf{42.8} & \textbf{27.2} \\
\midrule

\multicolumn{10}{c}{\textbf{Grounding DINO-B}}\\
\addlinespace[0.25em]
\cmidrule(lr){1-10}

& \multicolumn{3}{c}{\textbf{Watercolor}}
& \multicolumn{3}{c}{\textbf{ClipArt}}
& \multicolumn{3}{c}{\textbf{Comic}}\\
\cmidrule(lr){2-4}\cmidrule(lr){5-7}\cmidrule(lr){8-10}
\textbf{Method}
& \textbf{mAP} & \textbf{AP$_{50}$} & \textbf{AP$_{75}$}
& \textbf{mAP} & \textbf{AP$_{50}$} & \textbf{AP$_{75}$}
& \textbf{mAP} & \textbf{AP$_{50}$} & \textbf{AP$_{75}$} \\
\midrule
ZS    & 42.6 & 70.5 & 44.6 & 53.0 & 77.9 & 58.8 & 38.7 & 64.7 & 39.2 \\
\rowcolor{gray!12}
\textbf{\textsc{VLOD-TTA}} & \textbf{44.7} & \textbf{72.8} & \textbf{46.9} & \textbf{55.7} & \textbf{81.2} & \textbf{60.9} & \textbf{39.9} & \textbf{67.1} & \textbf{41.1} \\
\bottomrule
\end{tabular}
\end{adjustbox}
\end{table*}

\noindent\textbf{Effectiveness of \textsc{VLOD-TTA} on Specialized Domains. } To assess whether \textsc{VLOD-TTA} remains effective beyond generic benchmarks, we evaluate it on the Aquarium Object Detection dataset~\cite{aquarium_dataset}, which contains underwater animals such as fish, jellyfish, penguin, puffin, shark, starfish, and stingray. These categories are rare in standard benchmarks and define a specialized domain that is challenging for prompt selection. As shown in \Cref{tab:aquarium_ips}, CLIP-style prompts degrade ZS performance, demonstrating that they are ineffective for ODs. In contrast, IPS improves performance over the ZS baseline across all metrics, and combining IPS with IWE in \textsc{VLOD-TTA} yields further gains. These results indicate that \textsc{VLOD-TTA} remains effective even in specialized domains.

\begin{table}[h]
\centering
\caption{\textbf{Effectiveness of \textsc{VLOD-TTA} on a specialized domain.} Detection performance on the Aquarium Object Detection dataset, which represents a specialized underwater domain.}
\scriptsize
\setlength{\tabcolsep}{4pt}
\renewcommand{\arraystretch}{0.95}
\begin{tabular}{lccc}
\toprule
\textbf{Aquarium}   & \textbf{mAP} & \textbf{AP$_{50}$} & \textbf{AP$_{75}$} \\
\midrule
Zero-shot           & 11.9 & 20.3 & 11.6 \\
CLIP-Style Prompts & 11.6 & 19.7 & 11.2 \\
IPS                 & 12.4 & 21.5 & 12.1 \\
\midrule
\rowcolor{gray!15}
\textbf{VLOD-TTA}   & \textbf{14.5} & \textbf{25.1} & \textbf{14.8} \\
\bottomrule
\end{tabular}

\label{tab:aquarium_ips}
\end{table}

\subsection{Extended Benchmark Results}
\label{sec:benchmark}

\noindent\textbf{Detailed Comparison with TTAOD-F. } \Cref{tab:ttaod-f} reports the full corruption-wise mAP comparison on COCO-C for Grounding DINO. TTAOD-F~\cite{ttaod-f} is the only prior TTA method specifically designed for VLODs, so it provides the most relevant baseline for comparison. Consistent with the summary results reported in the main paper, both \textsc{VLOD-TTA} and \textsc{VLOD-TTA}$^*$ improve substantially over the ZS baseline across most corruption types.

Compared with TTAOD-F, \textsc{VLOD-TTA} achieves a slightly higher average AP (26.2 vs.\ 26.0). The gains are especially clear on several blur and weather corruptions, including Defocus Blur (20.6 vs.\ 17.8), Motion Blur (18.4 vs.\ 16.9), Zoom Blur (10.1 vs.\ 7.9), Frost (36.6 vs.\ 34.7), and Fog (48.2 vs.\ 45.1). It also improves on Brightness (46.2 vs.\ 44.9). TTAOD-F performs better on some digital corruptions, particularly Contrast, Elastic Transform, Pixelate, and JPEG, and also slightly outperforms \textsc{VLOD-TTA} on Impulse Noise and Glass Blur. Overall, however, \textsc{VLOD-TTA} provides the best balance between robustness and efficiency while maintaining a small adaptation budget.

The warm-start variant, \textsc{VLOD-TTA}$^*$, further strengthens this comparison. Using the same test-time initialization strategy as TTAOD-F, it improves the average AP to 27.3 and outperforms both TTAOD-F and the default \textsc{VLOD-TTA} on most corruption types. In particular, \textsc{VLOD-TTA}$^*$ achieves the strongest results on Gaussian Noise, Shot Noise, Defocus Blur, Motion Blur, Snow, Frost, Brightness, Contrast, Elastic Transform, and Pixelate, while matching the best JPEG result. These results show that the gains of \textsc{VLOD-TTA} are not tied to a specific initialization scheme and can be further amplified by warm-start initialization.

Taken together, the detailed results in \Cref{tab:ttaod-f} support the conclusion from the main paper. \textsc{VLOD-TTA} consistently improves robustness over ZS, slightly surpasses TTAOD-F in average AP, and does so with a much better efficiency profile. When combined with warm-start initialization, \textsc{VLOD-TTA}$^*$ establishes the strongest overall performance on COCO-C.\\

\begin{table*}[h]
\centering
\caption{\textbf{Detection performance of TTA methods on COCO-C.}  mAP is reported for the \textbf{Grounding-DINO} detector on 15 different data corruptions. 
Best results are highlighted in bold.}
\scriptsize
\setlength{\tabcolsep}{2.4pt}
\renewcommand{\arraystretch}{0.95}

\begin{adjustbox}{max width=\textwidth}
\begin{tabular}{
  l
  *{15}{S[table-format=2.1]}
  | S[table-format=2.1]
}
\toprule

& \multicolumn{3}{c}{\textbf{Noise}}
& \multicolumn{4}{c}{\textbf{Blur}}
& \multicolumn{3}{c}{\textbf{Weather}}
& \multicolumn{5}{c}{\textbf{Digital}}
& \multicolumn{1}{c}{\textbf{}} \\
\cmidrule(lr){2-4}\cmidrule(lr){5-8}\cmidrule(lr){9-11}\cmidrule(lr){12-16}\cmidrule(lr){17-17}

\textbf{Method}
& \textbf{Gaussi.}
& \textbf{Shot}
& \textbf{Impulse}
& \textbf{Defocus}
& \textbf{Glass}
& \textbf{Motion}
& \textbf{Zoom}
& \textbf{Snow}
& \textbf{Frost}
& \textbf{Fog}
& \textbf{Bright.}
& \textbf{Contrast}
& \textbf{Elastic}
& \textbf{Pixel}
& \textbf{JPEG}
& \textbf{Avg}
\\
\midrule
ZS\cite{groundingdino}
& 13.7 & 16 & 15 & 16.8 & 7.5 & 13.6 & 6.7 & 27.5 & 32.5 & 44.2 & 44.1 & 21.9 & 22.5 & 5.3 & 21.1 & 20.6 \\
TTAOD-F\cite{ttaod-f}
& 20.2 & 22 & \textbf{21.4} & 17.8 & \textbf{14.5} & 16.9 & 7.9 & 31.1 & 34.7 & 45.1 & 44.9 & 30.6 & 29.9 & 23.6 & \textbf{29.2} & 26 \\
\rowcolor{gray!12}
\bottomrule
\textbf{VLOD-TTA}
& 20.5 & 21.8 & 20.9 & 20.6 & 13.1 &
  18.4 & 10.1 & 30.8 & 36.6 & \textbf{48.2} &
  46.2 & 28.8 & 28.7 & 19.7 & 28.6 & 26.2 \\
  \rowcolor{gray!12}
\textbf{VLOD-TTA$^*$}
& \textbf{21.1} & \textbf{22.2} & 21.2 & \textbf{21.4} & 12.9 &
  \textbf{19.7} & \textbf{9.9} & \textbf{32.3} & \textbf{36.9} & 47.8 & \textbf{47} &
  \textbf{31.6} & \textbf{30.1} & \textbf{25.9} & \textbf{29.2} & \textbf{27.3} \\
\bottomrule
\vspace{-1cm}

\end{tabular}
\end{adjustbox}
\label{tab:ttaod-f}
\end{table*}

\noindent\textbf{Variation in performance across corruption types and severity levels on COCO-C. }
We evaluate our approach on the COCO-C~\cite{michaelis2019benchmarking} benchmark across five corruption severities and fifteen corruption types. Full results are reported in \Cref{tab:coco_c_map,tab:coco_c_map50,tab:coco_c_map75}. Compared with PASCAL-C~\cite{michaelis2019benchmarking}, COCO-C contains 80 categories, which makes test-time adaptation (TTA) more challenging. Overall, the zero-shot (ZS) YOLO-World baseline degrades consistently as corruption severity increases for nearly all corruption types.

For certain corruptions, such as Gaussian noise, shot noise, impulse noise, and pixelation, the baseline mAP approaches zero at high severity, underscoring the need for TTA in vision--language object detection (VLOD). The TPT and VPT baselines also overfit on this benchmark and sometimes perform worse than ZS, highlighting the difficulty of adapting to COCO-C. The DPE baseline likewise struggles, yielding only marginal gains over ZS. In several cases, standard entropy minimization is the strongest baseline and can marginally outperform our method on specific corruptions. Nevertheless, our method improves over the ZS baseline in every setting, including mAP, AP$_{50}$, and AP$_{75}$, across all corruption types and severities.

The most noticeable gains are observed for the digital corruptions. Across severity levels, gains are largest at severities 2--4. At severity 1, the improvements are smaller because the test distribution remains close to the training distribution. At severity 5, performance is severely degraded for all methods, making most predictions unreliable and limiting the potential for improvement. Even so, our approach still yields consistent positive gains over ZS. These results demonstrate that the proposed TTA strategy substantially enhances robustness on COCO-C, particularly under moderate corruption, while still providing benefits under severe distribution shifts.\\

\begin{table*}[h]
\centering
\caption{\textbf{Detection performance of our method and zero-shot across all severity levels on COCO-C.} We report \textbf{AP} for \textbf{YOLO-World} on 15 corruption types and five severity levels. For each corruption, the best result is highlighted in bold.}
\scriptsize
\setlength{\tabcolsep}{2.4pt}
\renewcommand{\arraystretch}{0.95}
\begin{adjustbox}{max width=\textwidth}
\begin{tabular}{
  c l
  *{15}{S[table-format=2.1]}
  | S[table-format=2.1]
}
\toprule
& & \multicolumn{3}{c}{\textbf{Noise}}
& \multicolumn{4}{c}{\textbf{Blur}}
& \multicolumn{3}{c}{\textbf{Weather}}
& \multicolumn{5}{c}{\textbf{Digital}}
& {\textbf{}}\\
\cmidrule(lr){3-5}\cmidrule(lr){6-9}\cmidrule(lr){10-12}\cmidrule(lr){13-17}\cmidrule(lr){18-18}
\textbf{Sev.} & \textbf{Method}
& \textbf{Gauss} & \textbf{Shot} & \textbf{Impul}
& \textbf{Defoc} & \textbf{Glass} & \textbf{Motn} & \textbf{Zoom}
& \textbf{Snow} & \textbf{Frost} & \textbf{Fog}
& \textbf{Brit} & \textbf{Contr} & \textbf{Elast} & \textbf{Pixel} & \textbf{JPEG}
& \textbf{Avg}\\

\midrule

\multirow{6}{*}{1} & ZS
 & 29.0 & 29.3 & 25.2 & 32.6 & 28.8 & 30.2 & 13.4 & 27.3 & 32.2 & 34.8 & 36.7 & 34.9 & 32.2 & 24.8 & 26.8
& { 29.2 } \\
& TPT  & 27.2 & 27.5 & 23.3 & 30.8 & 28.8 & 29.8 & 12.5 & 26.7 & 31.2 & 33.9 & 36.0 & 34.0 & 31.5 & 24.7 & 26.6
& { 28.3 } \\
& VPT  & 28.3 & 28.6 & 24.7 & 31.1 & 28.5 & 29.3 & 12.9 & 26.8 & 31.1 & 33.5 & 35.5 & 33.5 & 31.5 & 24.9 & 26.6
& { 28.5 } \\
& DPE  & 28.8 & 29.0 & 25.1 & 32.8 & 29.3 & 30.5 & 13.7 & 27.5 & 32.3 & 34.7 & 36.5 & 35.1 & 32.4 & 25.0 & 27.2
& { 29.3 } \\
& Adapter  & 28.7 & 29.1 & 25.8 & 32.3 & \textbf{29.5} & 30.6 & 13.7 & 28.0 & 32.4 & 34.8 & 36.2 & 34.7 & 32.8 & 28.4 & 28.7
& { 29.7 } \\
 \rowcolor{gray!12}\cellcolor{white} & \textbf{Our}
 & \textbf{29.6} & \textbf{30.1} & \textbf{26.1} & \textbf{33.8} & 29.4 & \textbf{31.1} & \textbf{14.5} & \textbf{28.6} & \textbf{33.1} & \textbf{36.1} & \textbf{38.2} & \textbf{35.7} & \textbf{33.0} & \textbf{29.7} & \textbf{29.5}
& {\textbf{30.6}} \\
\cmidrule(lr){1-18}

\multirow{6}{*}{2} & ZS
 & 22.9 & 22.8 & 18.8 & 29.3 & 21.7 & 23.1 & 8.3 & 19.6 & 26.6 & 33.9 & 35.9 & 33.2 & 28.8 & 17.1 & 20.4
& { 24.2 } \\
& TPT  & 21.3 & 21.4 & 17.2 & 28.8 & 21.7 & 23.0 & 8.7 & 19.4 & 26.0 & 33.0 & 35.2 & 32.4 & 28.1 & 17.3 & 20.5
& { 23.6 } \\
& VPT  & 22.8 & 22.8 & 18.9 & 27.8 & 21.8 & 22.4 & 8.0 & 19.5 & 25.7 & 32.5 & 34.7 & 31.9 & 28.0 & 17.7 & 20.6
& { 23.7 } \\
& DPE  & 23.0 & 23.2 & 19.3 & 29.4 & 22.2 & 23.4 & 8.5 & 19.7 & 26.4 & 34.0 & 35.7 & 33.3 & 29.4 & 17.6 & 20.5
& { 24.4 } \\
& Adapter  & 22.8 & 23.3 & 20.1 & 29.1 & 24.5 & 23.8 & 8.5 & 21.0 & 27.8 & 34.1 & 35.3 & 33.2 & 30.1 & 21.0 & 23.6
& { 25.2 } \\
 \rowcolor{gray!12}\cellcolor{white} & \textbf{Our}
 & \textbf{23.7} & \textbf{23.5} & \textbf{20.4} & \textbf{31.0} & \textbf{24.7} & \textbf{24.6} & \textbf{ 9.6} & \textbf{21.4} & \textbf{27.9} & \textbf{35.4} & \textbf{37.5} & \textbf{35.4} & \textbf{30.2} & \textbf{24.1} & \textbf{24.0}
& {\textbf{26.2}} \\
\cmidrule(lr){1-18}

\multirow{6}{*}{3} & ZS
 & 13.7 & 14.9 & 13.5 & 22.0 & 6.3 & 14.6 & 6.3 & 20.0 & 23.0 & 32.9 & 35.0 & 29.7 & 23.4 & 7.7 & 16.8
& { 18.7 } \\
& TPT  & 12.6 & 13.6 & 13.3 & 21.8 & 5.5 & 14.6 & 6.5 & 19.6 & 23.3 & 32.0 & 34.3 & 29.1 & 22.9 & 8.5 & 17.1
& { 18.3 } \\
& VPT  & 14.2 & 15.3 & 14.1 & 20.9 & 6.9 & 14.5 & 6.3 & 20.0 & 22.5 & 31.6 & 33.8 & 29.1 & 23.4 & 8.7 & 17.6
& { 18.6 } \\
& DPE  & 14.1 & 15.4 & 13.9 & 22.5 & 6.6 & 14.8 & 6.4 & 20.3 & 23.2 & 33.1 & 34.6 & 29.9 & 23.8 & 8.5 & 17.4
& { 19.0 } \\
& Adapter  & \textbf{14.3} & 15.7 & 15.0 & 22.4 & 9.2 & 15.6 & 6.7 & 20.9 & 24.0 & 33.0 & 34.3 & 30.7 & 25.3 & 10.2 & 20.2
& { 19.8 } \\
 \rowcolor{gray!12}\cellcolor{white} & \textbf{Our}
 & 14.1 & \textbf{16.1} & \textbf{15.4} & \textbf{23.1} & \textbf{  9.7} & \textbf{15.8} & \textbf{  7.7} & \textbf{21.6} & \textbf{24.7} & \textbf{33.1} & \textbf{36.4} & \textbf{30.9} & \textbf{25.6} & \textbf{11.8} & \textbf{20.9}
& {\textbf{20.5}} \\
\cmidrule(lr){1-18}

\multirow{6}{*}{4} & ZS
 & 5.1 & 4.9 & 4.4 & 14.6 & 4.0 & 8.1 & 4.4 & 15.7 & 22.1 & 32.7 & 33.6 & 19.0 & 19.7 & 4.3 & 9.1
& { 13.4 } \\
& TPT  & 4.5 & 4.2 & 3.8 & 13.8 & 3.5 & 7.8 & 4.0 & 15.6 & 21.4 & 32.0 & 32.9 & 18.9 & 19.5 & 4.0 & 8.7
& { 13.0 } \\
& VPT  & 5.7 & 5.6 & 5.0 & 13.9 & 4.6 & 8.0 & 4.5 & 16.0 & 21.8 & 31.8 & 32.6 & 19.0 & 19.5 & 4.8 & 9.8
& { 13.5 } \\
& DPE  & 5.6 & 5.4 & 5.1 & 15.1 & 4.2 & 8.5 & 4.6 & 16.2 & 22.2 & 32.6 & 33.9 & 19.4 & 20.2 & 4.9 & 10.3
& { 13.9 } \\
& Adapter  & 6.1 & 6.4 & 5.5 & 15.7 & 5.6 & 8.7 & 4.6 & 16.9 & 23.7 & 33.0 & 33.3 & 21.0 & \textbf{21.9} & 6.3 & 12.2
& { 14.7 } \\
 \rowcolor{gray!12}\cellcolor{white} & \textbf{Our}
 & \textbf{  6.5} & \textbf{  6.7} & \textbf{  5.9} & \textbf{16.3} & \textbf{  5.9} & \textbf{  8.9} & \textbf{  5.3} & \textbf{17.2} & \textbf{24.4} & \textbf{34.9} & \textbf{34.8} & \textbf{21.3} & 21.8 & \textbf{11.9} & \textbf{13.2}
& {\textbf{15.7}} \\
\cmidrule(lr){1-18}

\multirow{6}{*}{5} & ZS
 & 1.0 & 1.6 & 0.1 & 8.8 & 2.8 & 5.2 & 3.7 & 15.4 & 20.0 & 31.4 & 31.8 & 5.4 & 15.0 & 2.8 & 4.2
& { 9.9 } \\
& TPT  & 0.8 & 1.4 & 0.7 & 8.2 & 2.3 & 4.7 & 3.3 & 14.9 & 19.2 & 30.8 & 31.1 & 5.9 & 15.2 & 2.6 & 4.0
& { 9.7 } \\
& VPT  & 1.0 & 1.8 & 1.0 & 8.6 & 2.9 & 5.4 & 3.9 & 15.6 & 19.9 & 30.6 & 31.1 & 6.2 & 14.8 & 3.3 & 4.6
& { 10.0 } \\
& DPE  & 1.1 & 1.9 & 0.8 & 9.1 & 2.9 & 5.6 & 4.2 & 15.7 & 20.3 & 31.1 & 32.3 & 5.9 & 15.8 & 3.2 & 5.2
& { 10.3 } \\
& Adapter  & 1.1 & 2.1 & 1.0 & 9.8 & 3.2 & 6.0 & 4.0 & 16.8 & 21.4 & 32.1 & 31.7 & 7.6 & 16.7 & 4.1 & 6.2
& { 10.9 } \\
 \rowcolor{gray!12}\cellcolor{white} & \textbf{Our}
 & \textbf{  1.9} & \textbf{  2.3} & \textbf{  1.7} & \textbf{10.4} & \textbf{  3.4} & \textbf{  6.9} & \textbf{  5.1} & \textbf{16.9} & \textbf{21.7} & \textbf{33.2} & \textbf{32.1} & \textbf{  7.9} & \textbf{17.1} & \textbf{  4.7} & \textbf{  6.7}
& {\textbf{11.5}} \\
\bottomrule
\end{tabular}
\end{adjustbox}
\label{tab:coco_c_map}
\end{table*}

\begin{table*}[h]
\centering
\caption{\textbf{Detection performance of our method and zero-shot across all severity levels on COCO-C.} We report \textbf{AP$_{50}$} for \textbf{YOLO-World} on 15 corruption types and five severity levels. For each corruption, the best result is highlighted in bold.}
\scriptsize
\setlength{\tabcolsep}{2.4pt}
\renewcommand{\arraystretch}{0.95}
\begin{adjustbox}{max width=\textwidth}
\begin{tabular}{
  c l
  *{15}{S[table-format=2.1]}
  | S[table-format=2.1]
}
\toprule
& & \multicolumn{3}{c}{\textbf{Noise}}
& \multicolumn{4}{c}{\textbf{Blur}}
& \multicolumn{3}{c}{\textbf{Weather}}
& \multicolumn{5}{c}{\textbf{Digital}}
& {\textbf{}}\\
\cmidrule(lr){3-5}\cmidrule(lr){6-9}\cmidrule(lr){10-12}\cmidrule(lr){13-17}\cmidrule(lr){18-18}
\textbf{Sev.} & \textbf{Method}
& \textbf{Gauss} & \textbf{Shot} & \textbf{Impul}
& \textbf{Defoc} & \textbf{Glass} & \textbf{Motn} & \textbf{Zoom}
& \textbf{Snow} & \textbf{Frost} & \textbf{Fog}
& \textbf{Brit} & \textbf{Contr} & \textbf{Elast} & \textbf{Pixel} & \textbf{JPEG}
& \textbf{Avg}\\
\midrule

\multirow{6}{*}{1} & ZS
 & 41.5 & 42.0 & 36.1 & 46.0 & 40.8 & 44.0 & 24.3 & 39.0 & 45.4 & 48.7 & 51.3 & 48.8 & 46.6 & 35.1 & 38.7
& { 41.9 } \\
& TPT  & 38.9 & 39.5 & 33.5 & 43.5 & 41.8 & 43.3 & 22.9 & 28.5 & 44.4 & 48.0 & 50.7 & 48.1 & 45.7 & 34.5 & 38.0
& { 40.1 } \\
& VPT  & 41.0 & 41.4 & 36.2 & 44.9 & 42.3 & 43.2 & 23.9 & 38.7 & 44.0 & 47.6 & 50.2 & 47.7 & 45.6 & 35.6 & 39.0
& { 41.4 } \\
& DPE  & 41.5 & 41.7 & 36.6 & 46.2 & 41.3 & 44.5 & 24.9 & 39.6 & 45.8 & 48.8 & 51.1 & 48.9 & 46.9 & 35.4 & 39.3
& { 42.2 } \\
& Adapter  & 41.4 & 42.1 & 37.0 & 45.9 & 42.3 & 44.7 & 25.2 & 40.1 & 45.9 & 48.8 & 50.7 & 48.8 & 47.2 & 40.2 & 41.7
& { 42.8 } \\
 \rowcolor{gray!12}\cellcolor{white} & \textbf{Our}
 & \textbf{42.1} & \textbf{43.1} & \textbf{37.1} & \textbf{47.6} & \textbf{42.5} & \textbf{44.9} & \textbf{25.5} & \textbf{40.7} & \textbf{46.6} & \textbf{49.7} & \textbf{52.4} & \textbf{49.5} & \textbf{47.5} & \textbf{41.9} & \textbf{42.5}
& {\textbf{43.6}} \\
\cmidrule(lr){1-18}

\multirow{6}{*}{2} & ZS
 & 33.4 & 33.2 & 27.6 & 41.9 & 31.3 & 35.1 & 16.7 & 28.7 & 38.1 & 47.6 & 50.2 & 46.6 & 42.2 & 24.1 & 30.3
& { 35.1 } \\
& TPT  & 30.9 & 31.2 & 25.2 & 40.8 & 30.6 & 34.4 & 16.9 & 27.5 & 37.6 & 45.8 & 48.6 & 45.0 & 40.4 & 24.4 & 30.8
& { 34.0 } \\
& VPT  & 33.6 & 33.6 & 28.1 & 40.7 & 32.0 & 34.6 & 16.5 & 28.7 & 37.7 & 46.4 & 49.1 & 45.5 & 41.4 & 25.5 & 30.8
& { 34.9 } \\
& DPE  & 33.9 & 34.3 & 28.6 & 42.2 & 32.3 & 35.6 & 17.0 & 28.9 & 37.8 & 47.8 & 49.8 & 46.8 & 42.9 & 25.2 & 30.4
& { 35.6 } \\
& Adapter  & 33.5 & 34.2 & 29.8 & 42.0 & \textbf{35.5} & 36.3 & 17.3 & 30.8 & \textbf{40.0} & 47.9 & 49.5 & 46.9 & \textbf{44.0} & 29.9 & 35.0
& { 36.8 } \\
 \rowcolor{gray!12}\cellcolor{white} & \textbf{Our}
 & \textbf{34.8} & \textbf{34.8} & \textbf{29.9} & \textbf{43.1} & \textbf{35.5} & \textbf{36.9} & \textbf{17.9} & \textbf{30.9} & \textbf{40.0} & \textbf{48.5} & \textbf{51.1} & \textbf{49.5} & 43.7 & \textbf{34.3} & \textbf{35.5}
& {\textbf{37.8}} \\
\cmidrule(lr){1-18}

\multirow{6}{*}{3} & ZS
 & 20.5 & 22.2 & 20.1 & 32.9 & 9.5 & 23.1 & 13.7 & 29.4 & 33.1 & 46.1 & 49.0 & 41.9 & 35.2 & 11.1 & 25.1
& { 27.5 } \\
& TPT  & 19.7 & 21.3 & 19.1 & 32.0 & 8.3 & 22.5 & 13.9 & 28.2 & 32.6 & 45.6 & 48.4 & 41.7 & 34.8 & 11.7 & 25.6
& { 27.0 } \\
& VPT  & 21.6 & 23.1 & 21.3 & 32.2 & 10.8 & 23.4 & 13.7 & 29.7 & 33.1 & 45.1 & 47.9 & 41.8 & 35.5 & 12.8 & 27.0
& { 27.9 } \\
& DPE  & 20.8 & 22.9 & 21.1 & 33.4 & 10.3 & 23.8 & 14.0 & 29.9 & 33.3 & 46.3 & 48.6 & 42.3 & 35.9 & 12.5 & 26.4
& { 28.1 } \\
& Adapter  & 22.2 & 23.8 & 22.6 & 33.6 & 14.5 & 25.0 & 14.5 & 30.9 & 34.8 & 46.4 & 48.3 & 43.4 & 38.4 & 14.7 & 30.6
& { 29.6 } \\
 \rowcolor{gray!12}\cellcolor{white} & \textbf{Our}
 & \textbf{22.8} & \textbf{24.3} & \textbf{22.9} & \textbf{34.1} & \textbf{14.9} & \textbf{25.1} & \textbf{14.8} & \textbf{31.2} & \textbf{35.6} & \textbf{47.4} & \textbf{50.5} & \textbf{43.5} & \textbf{38.5} & \textbf{17.0} & \textbf{31.3}
& {\textbf{30.3}} \\
\cmidrule(lr){1-18}

\multirow{6}{*}{4} & ZS
 & 7.8 & 7.4 & 6.7 & 22.6 & 6.1 & 13.4 & 10.1 & 23.5 & 32.0 & 45.9 & 47.3 & 27.5 & 30.2 & 6.2 & 14.0
& { 20.0 } \\
& TPT  & 6.8 & 6.4 & 5.8 & 21.2 & 5.2 & 12.9 & 9.3 & 23.6 & 31.4 & 45.4 & 46.8 & 27.7 & 29.0 & 5.8 & 13.1
& { 19.4 } \\
& VPT  & 8.7 & 8.6 & 7.7 & 22.2 & 7.1 & 13.5 & 10.4 & 24.1 & 31.9 & 45.2 & 46.4 & 28.0 & 30.1 & 7.1 & 15.1
& { 20.4 } \\
& DPE  & 8.5 & 8.4 & 7.8 & 23.2 & 6.5 & 13.9 & 10.6 & 24.5 & 32.3 & 46.0 & 47.6 & 28.2 & 30.9 & 7.4 & 15.5
& { 20.8 } \\
& Adapter  & \textbf{  9.6} & 10.0 & 8.6 & 24.5 & 8.6 & 14.7 & 10.7 & \textbf{25.6} & 34.3 & 46.6 & 47.1 & \textbf{30.7} & 33.8 & 9.2 & 19.0
& { 22.2 } \\
 \rowcolor{gray!12}\cellcolor{white} & \textbf{Our}
 & 9.3 & \textbf{10.2} & \textbf{  8.9} & \textbf{25.2} & \textbf{  8.9} & \textbf{14.8} & \textbf{11.8} & \textbf{25.6} & \textbf{36.2} & \textbf{48.1} & \textbf{49.1} & \textbf{30.7} & \textbf{34.1} & \textbf{17.5} & \textbf{19.6}
& {\textbf{23.3}} \\
\cmidrule(lr){1-18}

\multirow{6}{*}{5} & ZS
 & 1.4 & 2.5 & 0.2 & 13.8 & 4.3 & 9.1 & 8.9 & 22.9 & 28.9 & 44.2 & 45.0 & 8.0 & 23.6 & 3.9 & 6.5
& { 14.9 } \\
& TPT  & 1.2 & 2.1 & 1.1 & 12.8 & 3.5 & 8.1 & 8.1 & 21.6 & 37.2 & 43.8 & 44.4 & 8.2 & 24.0 & 3.6 & 6.6
& { 15.1 } \\
& VPT  & 1.5 & 2.8 & 1.6 & 13.8 & 4.6 & 9.3 & 9.1 & 23.6 & 29.2 & 43.8 & 44.3 & 9.2 & 23.7 & 4.7 & 7.3
& { 15.2 } \\
& DPE  & 1.6 & 2.9 & 1.4 & 14.2 & 4.5 & 9.5 & 9.3 & 23.4 & 29.6 & 44.1 & 45.6 & 8.9 & 24.8 & 4.4 & 7.9
& { 15.5 } \\
& Adapter  & 1.6 & 3.3 & 1.6 & 15.9 & 5.1 & 10.5 & 9.4 & 25.2 & 31.3 & 45.2 & 45.0 & 11.4 & 26.5 & 5.9 & 9.8
& { 16.5 } \\
 \rowcolor{gray!12}\cellcolor{white} & \textbf{Our}
 & \textbf{  1.9} & \textbf{  3.4} & \textbf{  1.9} & \textbf{16.4} & \textbf{  5.5} & \textbf{10.9} & \textbf{  9.8} & \textbf{25.2} & \textbf{31.4} & \textbf{45.9} & \textbf{46.1} & \textbf{11.7} & \textbf{26.7} & \textbf{  6.3} & \textbf{  9.9}
& {\textbf{16.9}} \\
\bottomrule
\end{tabular}
\end{adjustbox}
\vspace{-.3cm}
\label{tab:coco_c_map50}
\end{table*}

\begin{table*}[h]
\centering
\caption{\textbf{Detection performance of our method and zero-shot across all severity levels on COCO-C.} We report \textbf{AP$_{75}$} for \textbf{YOLO-World} on 15 corruption types and five severity levels. For each corruption, the best result is highlighted in bold.}
\scriptsize
\setlength{\tabcolsep}{2.4pt}
\renewcommand{\arraystretch}{0.95}
\begin{adjustbox}{max width=\textwidth}
\begin{tabular}{
  c l
  *{15}{S[table-format=2.1]}
  | S[table-format=2.1]
}
\toprule
& & \multicolumn{3}{c}{\textbf{Noise}}
& \multicolumn{4}{c}{\textbf{Blur}}
& \multicolumn{3}{c}{\textbf{Weather}}
& \multicolumn{5}{c}{\textbf{Digital}}
& {\textbf{}}\\
\cmidrule(lr){3-5}\cmidrule(lr){6-9}\cmidrule(lr){10-12}\cmidrule(lr){13-17}\cmidrule(lr){18-18}
\textbf{Sev.} & \textbf{Method}
& \textbf{Gauss} & \textbf{Shot} & \textbf{Impul}
& \textbf{Defoc} & \textbf{Glass} & \textbf{Motn} & \textbf{Zoom}
& \textbf{Snow} & \textbf{Frost} & \textbf{Fog}
& \textbf{Brit} & \textbf{Contr} & \textbf{Elast} & \textbf{Pixel} & \textbf{JPEG}
& \textbf{Avg}\\
\midrule

\multirow{6}{*}{1} & ZS
 & 31.3 & 31.5 & 27.1 & 35.3 & 30.5 & 32.5 & 13.3 & 29.4 & 34.9 & 37.8 & 39.9 & 37.9 & 34.8 & 26.8 & 28.6
& { 31.4 } \\
& TPT  & 29.1 & 29.6 & 25.0 & 33.5 & 31.1 & 31.9 & 12.4 & 28.9 & 33.8 & 36.5 & 38.9 & 37.1 & 34.0 & 26.5 & 28.3
& { 30.4 } \\
& VPT  & 30.2 & 30.5 & 26.0 & 33.2 & 30.2 & 31.1 & 12.7 & 28.7 & 33.2 & 35.7 & 38.1 & 35.7 & 33.9 & 26.6 & 28.5
& { 30.3 } \\
& DPE  & 30.9 & 31.3 & 27.2 & 35.4 & 30.9 & 32.7 & 13.6 & 29.7 & 35.1 & 37.7 & 39.6 & 38.0 & 34.9 & 27.1 & 29.1
& { 31.5 } \\
& Adapter  & 30.6 & 31.1 & 27.1 & 34.9 & 31.4 & 32.7 & 13.7 & 29.9 & 34.8 & 37.4 & 38.9 & 37.5 & 35.2 & 30.6 & 30.5
& { 31.8 } \\
 \rowcolor{gray!12}\cellcolor{white} & \textbf{Our}
 & \textbf{32.0} & \textbf{31.9} & \textbf{27.3} & \textbf{36.9} & \textbf{31.5} & \textbf{32.9} & \textbf{14.8} & \textbf{30.6} & \textbf{35.6} & \textbf{38.5} & \textbf{41.1} & \textbf{39.1} & \textbf{35.9} & \textbf{31.8} & \textbf{30.8}
& {\textbf{32.7}} \\
\cmidrule(lr){1-18}

\multirow{6}{*}{2} & ZS
 & 24.6 & 24.5 & 20.2 & 31.6 & 23.3 & 24.5 & 7.5 & 21.0 & 28.5 & 36.7 & 39.0 & 35.8 & 30.8 & 18.3 & 21.9
& { 25.9 } \\
& TPT  & 22.9 & 23.1 & 18.4 & 31.1 & 23.1 & 24.3 & 7.9 & 20.7 & 28.0 & 35.6 & 38.1 & 35.1 & 29.9 & 18.6 & 21.9
& { 25.2 } \\
& VPT  & 24.1 & 24.1 & 20.0 & 29.6 & 23.3 & 23.5 & 7.0 & 20.6 & 27.2 & 34.6 & 37.1 & 34.3 & 29.9 & 19.0 & 21.8
& { 25.1 } \\
& DPE  & 24.5 & 24.6 & 20.7 & 31.7 & 23.8 & 24.8 & 7.9 & 21.1 & 28.3 & 36.8 & 38.6 & 35.6 & 31.4 & 18.8 & 21.9
& { 26.0 } \\
& Adapter  & 24.0 & 24.7 & 21.2 & 31.2 & 26.2 & 24.7 & 7.4 & 22.2 & \textbf{29.7} & 36.5 & 37.9 & 35.7 & \textbf{32.1} & 22.5 & 25.0
& { 26.7 } \\
 \rowcolor{gray!12}\cellcolor{white} & \textbf{Our}
 & \textbf{25.1} & \textbf{25.1} & \textbf{23.4} & \textbf{33.1} & \textbf{26.4} & \textbf{25.9} & \textbf{  8.8} & \textbf{22.4} & \textbf{29.7} & \textbf{37.5} & \textbf{40.3} & \textbf{37.7} & 31.9 & \textbf{25.6} & \textbf{25.3}
& {\textbf{27.9}} \\
\cmidrule(lr){1-18}

\multirow{6}{*}{3} & ZS
 & 14.6 & 15.9 & 14.3 & 23.5 & 6.5 & 15.1 & 4.8 & 21.4 & 24.6 & 35.6 & 37.6 & 31.9 & 24.9 & 8.1 & 17.7
& { 19.8 } \\
& TPT  & 14.1 & 14.5 & 14.0 & 23.1 & 5.6 & 15.1 & 5.1 & 20.9 & 23.8 & 34.4 & 36.8 & 31.2 & 24.1 & 8.6 & 18.0
& { 19.3 } \\
& VPT  & 15.0 & 16.2 & 14.8 & 22.0 & 7.1 & 14.9 & 4.8 & 21.1 & 23.8 & 33.7 & 36.1 & 31.2 & 24.8 & 9.3 & 18.2
& { 19.5 } \\
& DPE  & 14.9 & 16.2 & 14.7 & 24.1 & 7.4 & 15.4 & 5.0 & 21.9 & 24.7 & 35.9 & 37.5 & 31.9 & 25.3 & 9.2 & 18.3
& { 20.2 } \\
& Adapter  & 15.5 & 16.4 & 15.7 & 23.8 & 10.0 & 16.1 & 5.2 & 22.2 & 25.7 & 35.4 & 36.7 & 32.8 & 26.3 & 10.8 & 21.0
& { 20.9 } \\
 \rowcolor{gray!12}\cellcolor{white} & \textbf{Our}
 & \textbf{15.8} & \textbf{16.9} & \textbf{16.3} & \textbf{24.4} & \textbf{10.2} & \textbf{16.5} & \textbf{  6.0} & \textbf{22.6} & \textbf{26.1} & \textbf{37.4} & \textbf{38.9} & \textbf{32.9} & \textbf{26.4} & \textbf{12.6} & \textbf{21.8}
& {\textbf{21.7}} \\
\cmidrule(lr){1-18}

\multirow{6}{*}{4} & ZS
 & 5.4 & 5.2 & 4.6 & 15.5 & 4.2 & 8.4 & 3.1 & 16.6 & 23.7 & 35.2 & 36.1 & 20.5 & 20.6 & 4.6 & 9.6
& { 14.2 } \\
& TPT  & 4.7 & 4.4 & 4.0 & 14.7 & 3.7 & 8.1 & 2.8 & 16.6 & 22.9 & 34.6 & 35.3 & 20.2 & 20.5 & 4.3 & 9.2
& { 13.7 } \\
& VPT  & 5.9 & 5.8 & 5.1 & 14.5 & 4.7 & 8.1 & 3.1 & 16.8 & 23.3 & 34.2 & 34.7 & 20.3 & 20.5 & 5.1 & 10.3
& { 14.2 } \\
& DPE  & 5.7 & 5.7 & 5.1 & 15.2 & 4.3 & 8.9 & 3.3 & 16.9 & 23.7 & 35.4 & 36.4 & 20.8 & 21.1 & 5.3 & 10.5
& { 14.6 } \\
& Adapter  & 6.3 & 6.7 & 5.8 & 16.5 & 5.8 & 8.8 & 3.1 & 17.9 & 25.2 & 35.2 & 35.6 & 22.5 & \textbf{22.9} & 6.7 & 12.8
& { 15.5 } \\
 \rowcolor{gray!12}\cellcolor{white} & \textbf{Our}
 & \textbf{  6.9} & \textbf{  6.8} & \textbf{  6.2} & \textbf{16.8} & \textbf{  6.2} & \textbf{  9.8} & \textbf{  5.8} & \textbf{18.7} & \textbf{25.6} & \textbf{37.1} & \textbf{37.3} & \textbf{22.2} & 22.7 & \textbf{12.7} & \textbf{13.1}
& {\textbf{16.5}} \\
\cmidrule(lr){1-18}

\multirow{6}{*}{5} & ZS
 & 0.9 & 1.6 & 0.1 & 9.3 & 2.9 & 5.3 & 2.6 & 16.4 & 21.3 & 33.8 & 34.1 & 5.8 & 15.4 & 3.0 & 4.4
& { 10.5 } \\
& TPT  & 0.8 & 1.4 & 0.7 & 8.7 & 2.5 & 4.8 & 2.3 & 15.8 & 20.5 & 33.2 & 33.4 & 6.2 & 15.8 & 2.8 & 4.5
& { 10.2 } \\
& VPT  & 1.0 & 1.8 & 1.1 & 9.0 & 3.0 & 5.3 & 2.7 & 16.4 & 21.0 & 32.6 & 32.9 & 6.5 & 15.2 & 3.5 & 4.8
& { 10.5 } \\
& DPE  & 1.0 & 1.8 & 0.8 & 9.4 & 3.2 & 5.5 & 3.1 & 16.8 & 21.8 & 33.6 & 34.3 & 6.4 & 16.0 & 3.3 & 5.1
& { 10.8 } \\
& Adapter  & 1.1 & \textbf{  2.2} & 1.0 & 10.1 & 3.3 & 5.8 & 2.8 & 17.6 & 22.6 & 34.5 & 33.8 & \textbf{  7.9} & 17.1 & 4.3 & 6.5
& { 11.4 } \\
 \rowcolor{gray!12}\cellcolor{white} & \textbf{Our}
 & \textbf{  1.3} & \textbf{  2.2} & \textbf{  1.6} & \textbf{10.6} & \textbf{  3.5} & \textbf{  6.4} & \textbf{  4.2} & \textbf{17.8} & \textbf{23.0} & \textbf{35.1} & \textbf{35.4} & 7.4 & \textbf{17.3} & \textbf{  4.9} & \textbf{  7.3}
& {\textbf{11.9}} \\
\bottomrule
\end{tabular}
\end{adjustbox}
\label{tab:coco_c_map75}
\end{table*}

\noindent\textbf{Additional results on PASCAL-C. } \Cref{tab:pascal-c-map,tab:pascal-c-75} report mAP and AP$_{75}$ for our method and the baselines on PASCAL-C. Across all 15 corruptions, our approach consistently outperforms the ZS baseline, mirroring the trend observed for AP$_{50}$ on the same dataset. Gains are evident across the Noise, Blur, Weather, and Digital families, with particularly strong improvements on challenging digital corruptions such as pixelate, JPEG, and contrast, as well as solid gains on classical noise corruptions. Overall, our method yields average improvements of $2.6$ mAP and $2.6$ AP$_{75}$ over ZS, indicating improved robustness under both metrics.

\begin{table*}[h]
\centering
\caption{\textbf{Detection performance of different test-time adaptation strategies on PASCAL-C.} We report \textbf{mAP} for \textbf{YOLO-World} on 15 corruption types. For each corruption, the best result is highlighted in bold.}
\scriptsize
\setlength{\tabcolsep}{2.4pt}
\renewcommand{\arraystretch}{0.95}

\begin{adjustbox}{max width=\textwidth}
\begin{tabular}{
  l
  *{15}{S[table-format=2.1]}
  | S[table-format=2.1]
}
\toprule

& \multicolumn{3}{c}{\textbf{Noise}}
& \multicolumn{4}{c}{\textbf{Blur}}
& \multicolumn{3}{c}{\textbf{Weather}}
& \multicolumn{5}{c}{\textbf{Digital}}
& {\textbf{}} \\
\cmidrule(lr){2-4}\cmidrule(lr){5-8}\cmidrule(lr){9-11}\cmidrule(lr){12-16}\cmidrule(lr){17-17}

\textbf{Method}
& \textbf{Gauss} & \textbf{Shot} & \textbf{Impul}
& \textbf{Defoc} & \textbf{Glass} & \textbf{Motn} & \textbf{Zoom}
& \textbf{Snow} & \textbf{Frost} & \textbf{Fog}
& \textbf{Brit} & \textbf{Contr} & \textbf{Elast} & \textbf{Pixel} & \textbf{JPEG}
& \textbf{Avg}\\

\midrule
ZS
& 7.2 & 7.1 & 6.8 & 35.8 & 10.3 & 22.4 & 12.6 & 24.3 & 36.1 & 54.9 & 57.4 & 31.5 & 36.9 & 6.8 & 14.0 & 24.3 \\
TPT
& 7.5 & 7.9 & 7.2 & 36.1 & 10.4 & 22.1 & 12.8 & 25.1 & 36.8 & 54.7 & 57.5 & 32.2 & 37.2 & 7.2 & 15.1 & 24.7 \\
VPT
& 7.9 & 7.6 & 7.1 & 35.6 & 10.8 & 22.1 & 12.6 & 24.8 & 36.4 & 54.9 & 56.8 & 32.7 & 37.3 & 7.4 & 15.3 & 24.6 \\
DPE
& 7.9 & 8.2 & 7.6 & 36.3 & 10.9 & 22.6 & 12.8 & 25.3 & 36.9 & 55.1 & 57.9 & 33.6 & 37.4 & 8.2 & 16.4 & 25.1 \\
Adapter
& 9.0 & 9.5 & 8.1 & 36.4 & 12.9 & 21.5 & 11.9 & 26.7 & 37.5 & 54.3 & 55.5 & 33.4 & 40.1 & 10.1 & 19.2 & 25.7 \\
\bottomrule
\rowcolor{gray!12}
\textbf{Our}
& \textbf{10.3} & \textbf{9.9} & \textbf{9.2} & \textbf{38.1} & \textbf{14.9} & \textbf{22.7} & \textbf{12.9} & \textbf{27.2} & \textbf{38.4} & \textbf{55.6} & \textbf{58.1} & \textbf{35.4} & \textbf{40.7} & \textbf{10.6} & \textbf{19.5} & \textbf{26.9} \\
\bottomrule
\end{tabular}
\end{adjustbox}
\label{tab:pascal-c-map}
\end{table*}

\begin{table*}[h]
\centering
\caption{\textbf{Detection performance of different test-time adaptation strategies on PASCAL-C.} We report \textbf{AP$_{75}$} for \textbf{YOLO-World} on 15 corruption types. For each corruption, the best result is highlighted in bold.}
\scriptsize
\setlength{\tabcolsep}{2.4pt}
\renewcommand{\arraystretch}{0.95}

\begin{adjustbox}{max width=\textwidth}
\begin{tabular}{
  l
  *{15}{S[table-format=2.1]}
  | S[table-format=2.1]
}
\toprule

& \multicolumn{3}{c}{\textbf{Noise}}
& \multicolumn{4}{c}{\textbf{Blur}}
& \multicolumn{3}{c}{\textbf{Weather}}
& \multicolumn{5}{c}{\textbf{Digital}}
& {\textbf{}} \\
\cmidrule(lr){2-4}\cmidrule(lr){5-8}\cmidrule(lr){9-11}\cmidrule(lr){12-16}\cmidrule(lr){17-17}

\textbf{Method}
& \textbf{Gauss} & \textbf{Shot} & \textbf{Impul}
& \textbf{Defoc} & \textbf{Glass} & \textbf{Motn} & \textbf{Zoom}
& \textbf{Snow} & \textbf{Frost} & \textbf{Fog}
& \textbf{Brit} & \textbf{Contr} & \textbf{Elast} & \textbf{Pixel} & \textbf{JPEG}
& \textbf{Avg}\\

\midrule
ZS
& 7.4 & 7.1 & 6.9 & 39.3 & 10.8 & 24.1 & 8.7 & 26.2 & 39.1 & 59.8 & 62.8 & 34.1 & 40.2 & 7.1 & 15.0 & 25.9 \\
TPT
& 7.9 & 7.8 & 7.2 & 39.4 & 11.2 & 24.0 & 9.0 & 26.9 & 39.3 & 59.7 & 62.9 & 35.2 & 41.1 & 7.9 & 16.2 & 26.4 \\
VPT
& 7.9 & 7.7 & 7.2 & 38.5 & 11.4 & 23.5 & 8.7 & 26.2 & 39.3 & 59.8 & 61.8 & 35.3 & 41.2 & 7.8 & 16.4 & 26.2 \\
DPE
& 8.3 & 8.3 & 7.9 & 39.5 & 11.3 & 24.3 & 9.1 & 27.2 & 39.8 & 59.9 & 63.1 & 35.9 & 41.5 & 8.3 & 17.6 & 26.8 \\
Adapter
& 9.1 & 9.7 & 8.3 & 39.2 & 13.6 & 22.6 & 8.3 & 28.0 & 40.4 & 58.9 & 60.2 & 35.7 & 43.3 & 10.7 & 20.5 & 27.2 \\
\bottomrule
\rowcolor{gray!12}
\textbf{Our}
& \textbf{10.3} & \textbf{10.2} & \textbf{9.1} & \textbf{40.9} & \textbf{15.4} & \textbf{24.4} & \textbf{8.9} & \textbf{28.8} & \textbf{41.5} & \textbf{60.1} & \textbf{63.3} & \textbf{38.1} & \textbf{44.3} & \textbf{11.1} & \textbf{20.6} & \textbf{28.5} \\
\bottomrule
\end{tabular}
\end{adjustbox}
\label{tab:pascal-c-75}
\end{table*}

\subsection{Qualitative Results}
\label{sec:qualitative}

\begin{figure}[!t]
	\centering
		\includegraphics[width=0.99\linewidth]{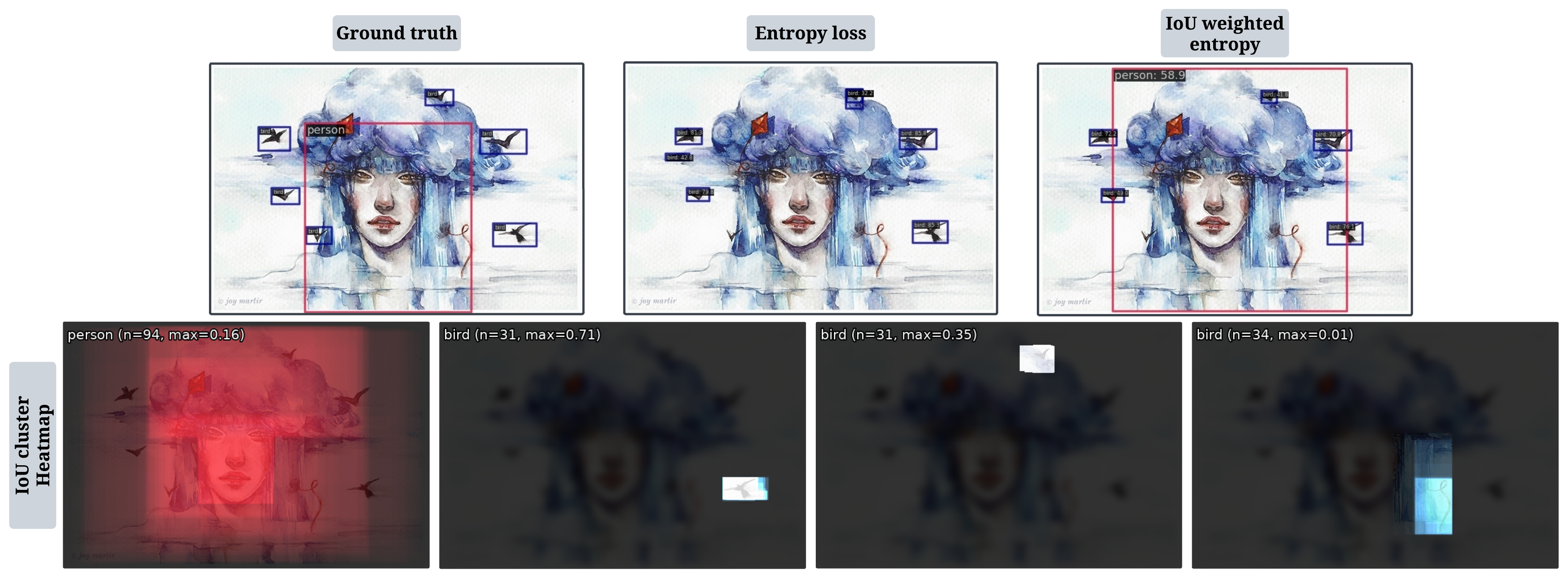}
    \caption{\textbf{Qualitative comparison of standard entropy and IoU-weighted entropy.} Top row: ground truth and predictions obtained with standard entropy and IoU-weighted entropy. Bottom row: heatmaps of the dominant IoU-graph clusters computed from the ZS proposals. For each cluster, we report the predicted category, cluster size, and maximum score. Only the top four clusters are shown.}

	\label{fig:ent_vs_iou}
\end{figure}

\noindent\textbf{Qualitative Analysis of Entropy and IoU-Weighted Entropy. } \Cref{fig:ent_vs_iou} provides a qualitative comparison between standard entropy minimization and our IoU-weighted entropy. The bottom-row heatmap of the ZS proposals shows one large but low-confidence \textit{person} cluster and several smaller \textit{bird} clusters. Although the largest \textit{person} cluster has a low maximum score ($0.16$), it contains $94$ mutually overlapping proposals and therefore represents the most spatially coherent object hypothesis in the image. In contrast, the \textit{bird} predictions are distributed across smaller clusters, whose maximum scores are higher locally but whose spatial support is much weaker.

Standard entropy minimization does not account for this overlap structure. It sharpens predictions based mainly on confidence, which favors the visually salient \textit{bird} proposals and suppresses the low-confidence \textit{person} prediction. As a result, the model increases the \textit{bird} scores while missing the actual \textit{person} object, leading to a false negative for the \textit{person} in the scene.

IoU-weighted entropy instead incorporates cluster structure into the adaptation objective. By assigning larger weights to proposals that belong to large, coherent overlap clusters, it amplifies the contribution of the dominant \textit{person} cluster while reducing the influence of smaller \textit{bird} clusters. This shifts adaptation toward the spatially consistent object hypothesis and recovers the correct \textit{person} detection. This example highlights the key advantage of IoU-weighted entropy: it prioritizes spatial consensus rather than relying only on per-proposal confidence.\\

\begin{figure*}[t]
\begingroup
\setlength{\parskip}{0pt}
\centering

\quadrowwithlabels{\detokenize{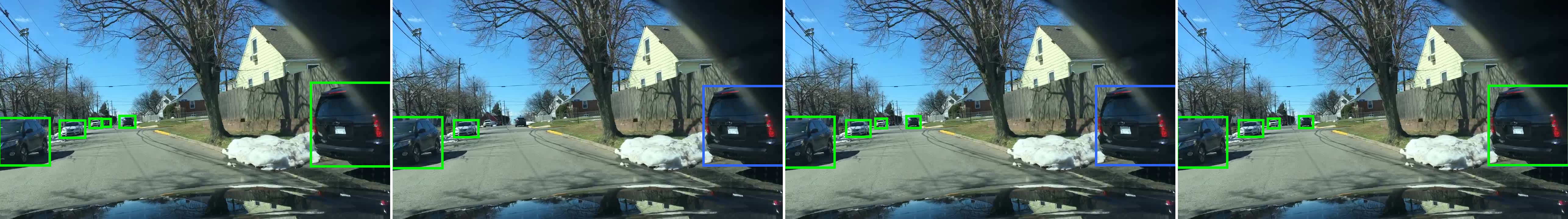}}\vspace{-0.01em}
\quadrow{\detokenize{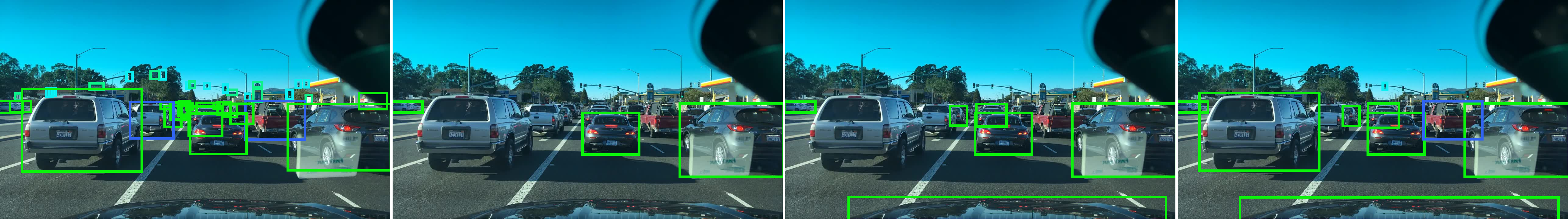}}\vspace{-0.01em}
\quadrow{\detokenize{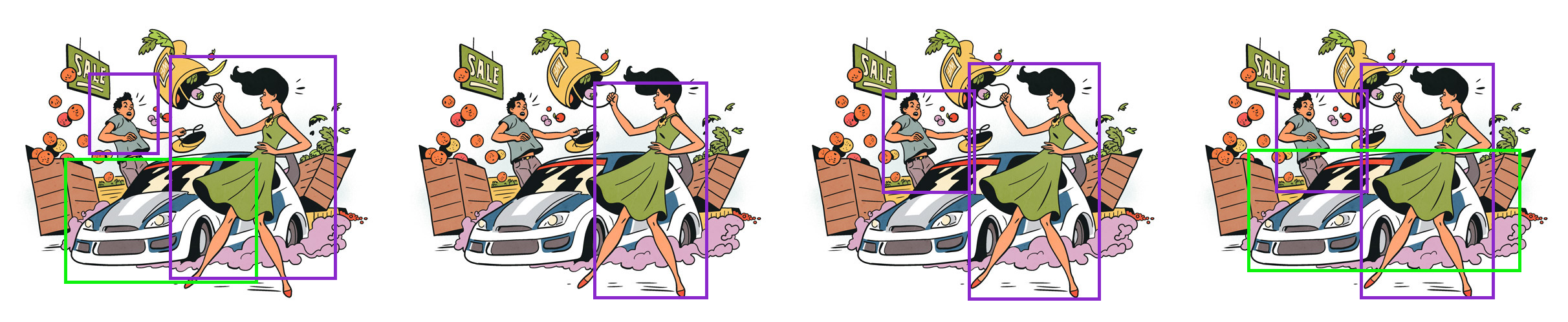}}
\quadrow{\detokenize{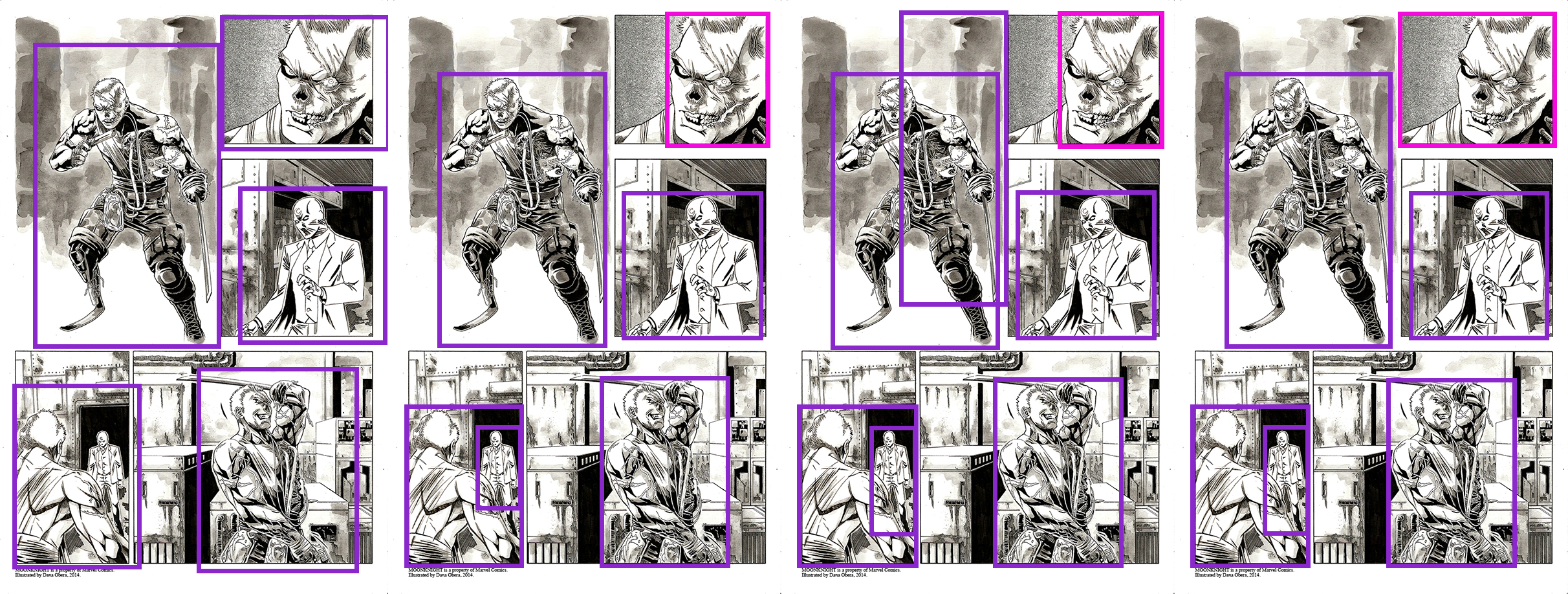}}
\quadrow{\detokenize{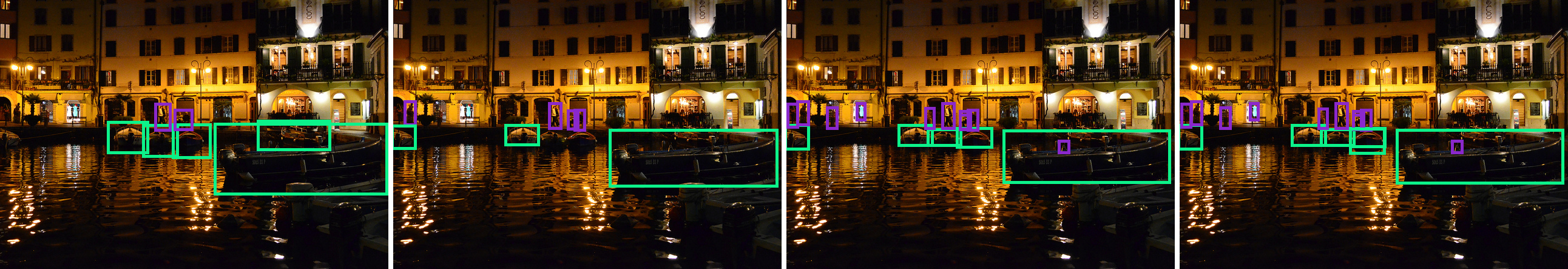}}
\quadrow{\detokenize{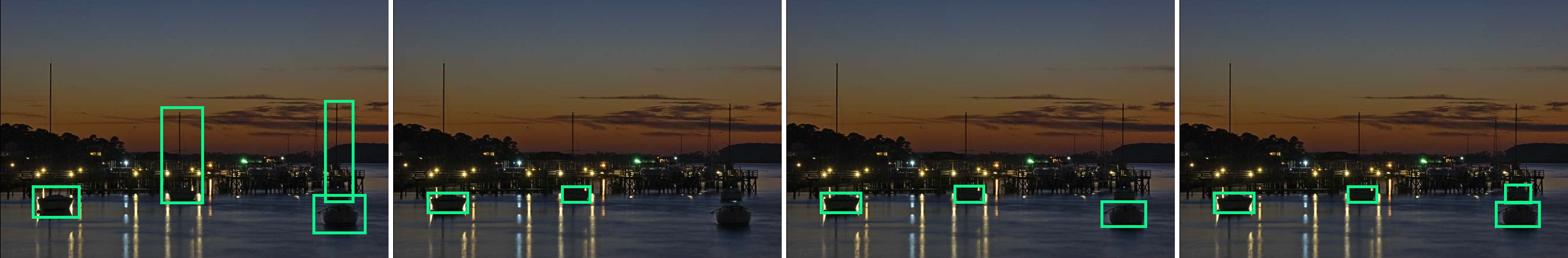}}
\quadrow{\detokenize{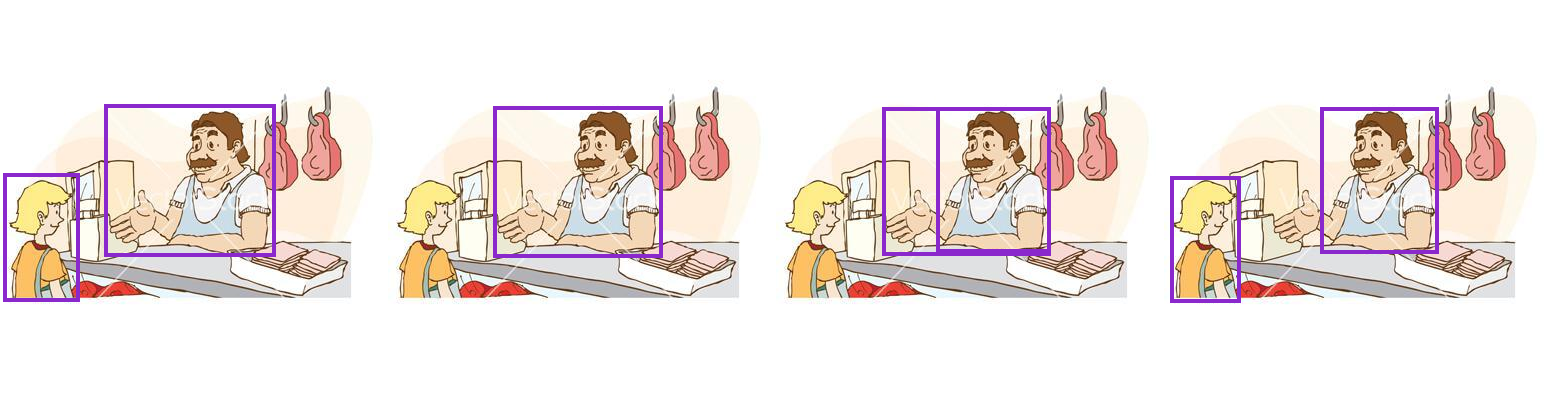}}
\caption{\textbf{YOLO-World detections across different approaches}: Each column corresponds to a different approach: (a) GT (Ground Truth), (b) ZS (Zero-Shot), (c) Adapter, and (d) \textsc{VLOD-TTA}.}
\label{fig:detection_viz2}
\endgroup
\end{figure*}

\noindent\textbf{Additional Detection Visualizations. } \Cref{fig:detection_viz2} provides additional qualitative examples from BDD, ExDark, Comic, and ClipArt. For each image, we show GT, ZS, Adapter, and \textsc{VLOD-TTA}. Compared with ZS and Adapter, our method typically (i) removes clear false positives, (ii) recovers missed objects under low-light and style-shift conditions, and (iii) produces tighter boxes with fewer duplicates. These qualitative trends are consistent with the quantitative gains reported in the main text.

